\documentclass[twocolumn,pagebackref,breaklinks,colorlinks,allcolors=iccvblue]{arxiv/fairmeta}

\definecolor{iccvblue}{rgb}{0.21,0.49,0.74}
\definecolor{citecolor}{HTML}{0071bc}
\hypersetup{
    colorlinks=true,%
    citecolor=citecolor,%
    filecolor=citecolor,%
    linkcolor=citecolor,%
    urlcolor=citecolor
}

\makeatletter
\DeclareFontFamily{T1}{optimistic}{%
  \@tempdima=0.88\dimexpr\f@size pt\relax
  \fontdimen2\font=0.28\@tempdima   %
  \fontdimen3\font=0.14\@tempdima   %
  \fontdimen4\font=0.07\@tempdima   %
  \fontdimen5\font=0.52\@tempdima   %
  \fontdimen6\font=\@tempdima       %
  \fontdimen7\font=0.07\@tempdima}  %
\makeatother

\usepackage{amsmath}
\usepackage{amssymb}
\usepackage{xspace}
\usepackage{array}
\usepackage{multirow}
\usepackage{colortbl}           %
\usepackage{placeins}
\usepackage{microtype}
\usepackage[symbol]{footmisc}
\usepackage{enumitem}

\providecommand{\eg}{\emph{e.g.}\xspace}

\newcommand{\themodel}{\textsc{AgenTAP}\xspace}
\newcommand{\thebench}{\textsc{TrackingBench}\xspace}
\newcommand{\thedataset}{\textsc{EgoTAP-4D}\xspace}

\newcommand{\papertitle}{Tracking Any Point with Agentic Visual Reasoning}

\definecolor{metablue}{HTML}{0081FB}  %

\newcommand{\toolname}[1]{\mbox{%
  \raisebox{-0.12em}{\includegraphics[height=0.95em]{figures/graphics/tool_icons/#1.png}}%
  \hspace{0.18em}\texttt{#1}}}

\definecolor{tabfirst}{HTML}{FFCCCC}
\definecolor{tabsecond}{HTML}{FFE6B3}
\definecolor{tabthird}{HTML}{FFFFCC}
\definecolor{tabband}{HTML}{F1F4F7}                %

\newcommand{\best}[1]{\cellcolor{tabfirst}{#1}}
\newcommand{\secondbest}[1]{\cellcolor{tabsecond}{#1}}
\newcommand{\thirdbest}[1]{\cellcolor{tabthird}{#1}}

\definecolor{pos}{RGB}{0,128,0}

\newcommand{\arxivbuild}{}


\setcitestyle{numbers,square,comma}

\renewcommand{\textfraction}{0}
\renewcommand{\topfraction}{1.0}
\makeatletter
\let\arxivincludegraphics\includegraphics
\renewcommand{\maketitle}{\twocolumn[\mymaketitle\vskip 0.05cm]}    %
\makeatother

\title{\papertitle}

\definecolor{uvaslate}{HTML}{3D4B74}
\definecolor{princetonorange}{HTML}{D16100}
\definecolor{umgold}{HTML}{AB7900}
\definecolor{umblue}{HTML}{00274C}
\definecolor{ummaize}{HTML}{FFCB05}
\newcommand{\equalfirstmark}{\textcolor{black}{*}}
\newcommand{\equalseniormark}{\textcolor{black}{\dagger}}
\newcommand{\leadmark}{\textcolor{black}{\ddagger}}
\DeclareRobustCommand{\metamark}{\raisebox{0.05em}{\includegraphics[height=0.45em,trim=0 0 2590 0,clip]{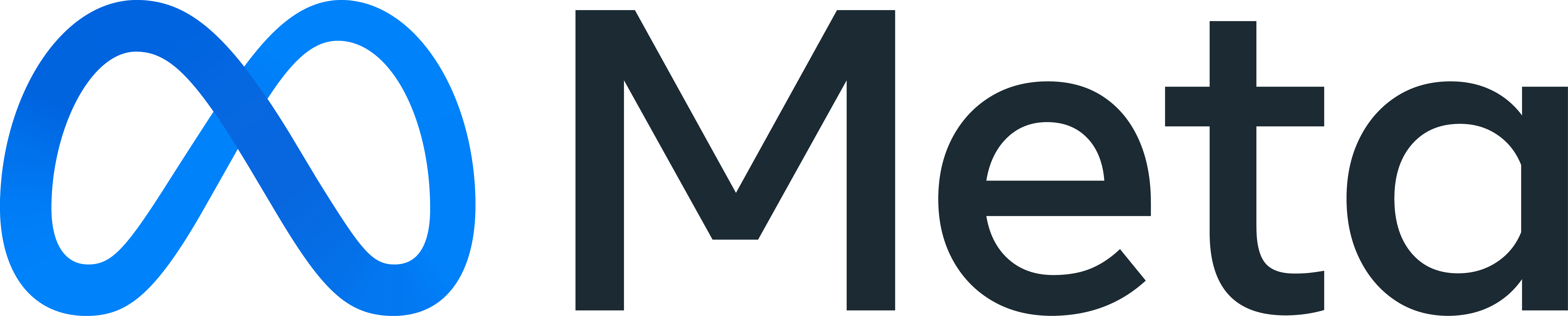}}}
\DeclareRobustCommand{\uvamark}{\raisebox{-0.125em}{\includegraphics[height=0.8em]{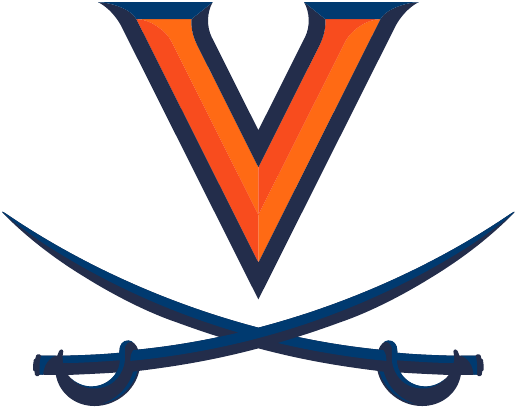}}}
\DeclareRobustCommand{\princetonmark}{\raisebox{0em}{\includegraphics[height=0.6em]{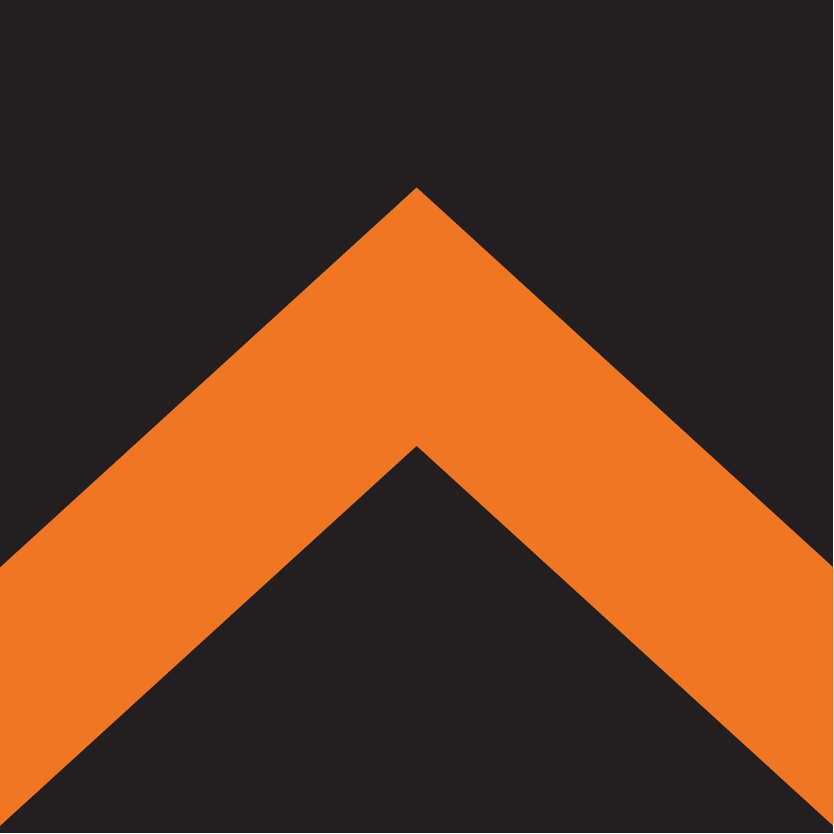}}}
\DeclareRobustCommand{\umichmark}{\text{\tikz[baseline=(m.base)]{\node[inner sep=.07em,fill=umblue,rounded corners=.07em,text=ummaize,font=\sffamily\bfseries] (m) {M};}}}
\newcommand{\marksep}{\,}

\author[\equalfirstmark\marksep\leadmark\marksep\uvamark]{Xuweiyi~Chen}
\author[\equalfirstmark\marksep\uvamark]{Jin~Yao}
\author[\equalfirstmark\marksep\princetonmark]{Linzhan~Mou}
\author[\equalfirstmark\marksep\uvamark]{Lei~Gong}
\author[\metamark]{Di~Liu}
\author[\metamark]{Cheng~Zhang}
\author[\umichmark]{Ziqiao~Ma}
\author[\umichmark]{Sihan~Xu}
\author[\metamark]{Adam~W.~Harley}
\author[\equalseniormark\marksep\metamark]{Shengyi~Qian}
\author[\equalseniormark\marksep\uvamark]{Zezhou~Cheng}

\affiliation[\uvamark]{\textcolor{uvaslate}{University of Virginia}}
\affiliation[\metamark]{\textcolor{metablue}{Meta}}
\affiliation[\princetonmark]{\textcolor{princetonorange}{Princeton University}}
\affiliation[\umichmark]{\textcolor{umgold}{University of Michigan}}

\contribution[\equalfirstmark]{equal contribution}
\contribution[\leadmark]{project lead}
\contribution[\equalseniormark]{equal advising}

\usepackage{fontawesome5}
\pdfmapfile{+fontawesome5.map}   %
\definecolor{projectlinkcolor}{HTML}{112D4E}
\definecolor{codelinkcolor}{HTML}{3F72AF}
\definecolor{benchmarklinkcolor}{HTML}{846491}
\definecolor{datalinkcolor}{HTML}{2E7D6B}
\newcommand{\projectlinks}{{\small\sffamily
  \href{https://agentap.cs.virginia.edu}{\textcolor{projectlinkcolor}{\faGlobe\enspace Project page}}\quad
  \href{https://github.com/UVA-Computer-Vision-Lab/agentap}{\textcolor{codelinkcolor}{\faGithub\enspace Code}}\quad
  \href{https://huggingface.co/datasets/uva-cv-lab/TrackingBench}{\textcolor{benchmarklinkcolor}{\faTrophy\enspace TrackingBench}}\quad
  \href{https://huggingface.co/datasets/uva-cv-lab/EgoTAP-4D}{\textcolor{datalinkcolor}{\faDatabase\enspace EgoTAP-4D}}%
}}
\makeatletter
\g@addto@macro\contributionlist{\par\vskip 0.2cm\projectlinks}
\makeatother

\abstract{
A general-purpose Vision-Language Model (VLM) has surpassed expert vision models in many domains such as object detection and video recognition. However, point correspondence is known to be challenging for VLMs since it goes beyond object-level recognition: it requires pixel-level localization of the same physical surface point across observations, even when appearance changes or the point disappears and later reappears.
In this paper, we introduce \themodel, a VLM-based agent that directly predicts point coordinates and visibility and actively reasons over videos.
Starting from its own estimate, the agent autonomously chooses what to inspect and uses the resulting evidence to verify point identity and refine its predictions.
With GPT-6 Astra, it surpasses expert vision models in both point localization and visibility prediction, without correspondence-specific fine-tuning.
We further use \themodel to re-identify physical points after disappearance and reconnect fragmented tracks, constructing \thedataset, a dataset of 3D point trajectories in long real-world egocentric videos.
On sampled reappearing points, its predictions approach human agreement, and adding \thedataset to D4RT training improves \mbox{average 3D tracking performance}.

}

\newlength{\arxivtitlegap}
\makeatletter
\patchcmd{\mymaketitle}{\setlength{\parskip}{0.5cm}}{\setlength{\parskip}{\arxivtitlegap}}{}{\PackageWarning{main_arxiv}{title-box gap patch 1 failed}}
\patchcmd{\mymaketitle}{\vskip 0.5cm}{\vskip\arxivtitlegap}{}{\PackageWarning{main_arxiv}{title-box gap patch 2 failed}}
\makeatother
\date{\today}

\begin{document}
\maketitle
\arrayrulecolor{black}

\begingroup
\setlength{\parskip}{3pt plus 1pt minus 1pt}
\titlespacing*{\section}{0pt}{10pt plus 2pt minus 2pt}{6pt}
\titlespacing*{\subsection}{0pt}{7pt plus 2pt minus 1pt}{3pt}
\widowpenalty=10000
\clubpenalty=10000

\let\arxivinput\input
\makeatletter
\renewcommand{\input}[1]{\ifstrequal{#1}{figures/tex/teaser}%
  {\begingroup\let\@addtotoporbot\@addtobot
   \renewcommand{\includegraphics}[2][]{\arxivincludegraphics[width=0.92\linewidth]{##2}}%
   \arxivinput{#1}\par\endgroup}{\arxivinput{#1}}}
\makeatother
\section{Introduction}
\label{sec:introduction}

Visual correspondence is a fundamental capability in computer vision~\citep{sift,tapvid,romav2}.
This task is traditionally tackled in two lines of research.
Image matching~\citep{superglue,loftr,roma,romav2} establishes correspondences across viewpoints, while point tracking~\citep{pips,tapir,cotracker3,alltracker} follows moving surfaces through time. Both rely on expert vision models trained to exploit appearance and temporal cues.
As a result, preserving point identity remains challenging when appearance changes dramatically or a point disappears and later reappears.

\begin{figure}[t]
    \centering
    \includegraphics[width=\linewidth]{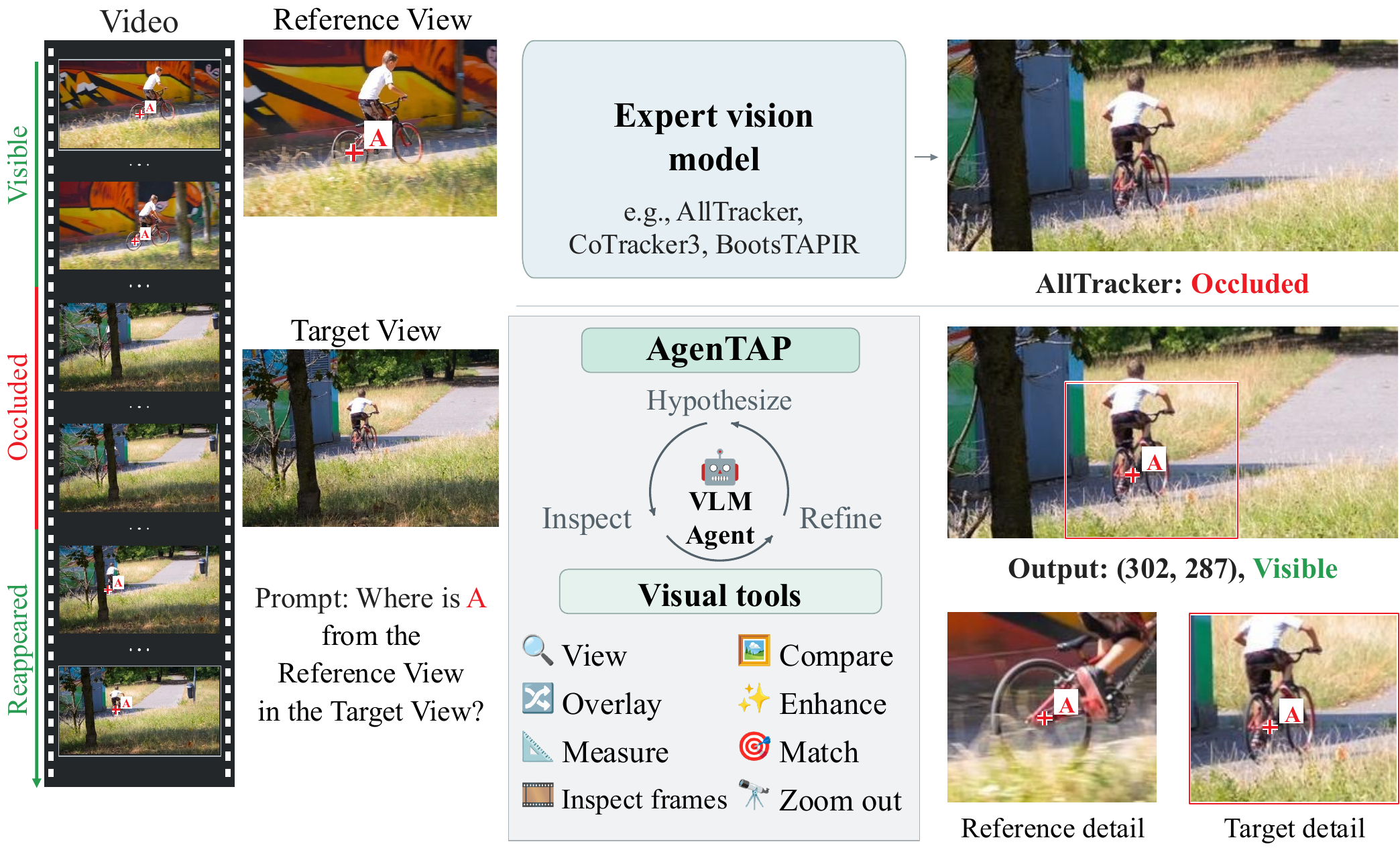}\caption{\textbf{Tracking Any Point with Agentic Visual Reasoning.} \themodel takes a top-down approach: a VLM agent hypothesizes point identity and visibility, selects visual tools to test its hypotheses, and refines predictions from the returned evidence. Here, \themodel locates a reappearing point that an expert vision model (\eg, AllTracker~\citep{alltracker}) incorrectly predicts as occluded.}
    \label{fig:teaser}
\end{figure}

Vision-language models (VLMs) have emerged as powerful general-purpose systems for visual understanding, reasoning, and interaction, demonstrating broad capabilities in interpreting complex scenes~\citep{blip2,llava}, solving visual reasoning problems~\citep{visualsketchpad}, grounding instructions~\citep{groundhog,glamm}, and autonomously operating computers~\citep{uitars2,agents3}.
However, visual correspondence remains an underexplored frontier for VLMs.
Their understanding of objects and spatial relationships could help preserve point identity across appearance changes, occlusion, and reappearance.
Our findings reveal a turning point: with agentic inference, GPT-6 Astra~\citep{gpt6-astra} becomes the first VLM in our evaluation to outperform expert vision models on the real-video benchmark.
Fable 5.2~\cite{claude-fable5.2} performs on par, while other VLMs have yet to cross this threshold.
We view this result as an early sign of a broader shift and expect more VLMs to follow as their capabilities advance.
This crossover in turn motivates a broader question:
\par\vspace{3pt}
{\centering\emph{How far have general-purpose VLMs progressed toward\\*
reliable pixel-level physical correspondence?}\par}
\vspace{3pt}

We systematically investigate this question using \thebench (Sec.~\ref{sec:benchmark}).
We curate 2,000 queries from 272 real-video clips across six domains, selecting visually diverse scenes and balancing visibility, motion, and frame gap.
An additional synthetic evaluation and independently annotated held-out videos probe distribution shifts and generalization.
Without tools, substantial gaps remain: VLMs fall far behind expert vision models on synthetic videos, and even GPT-6 Astra often identifies the right region without locating the exact point.
It places 87.1\% of visible points within 16 pixels but only 23.8\% within one pixel.
Other errors are semantically plausible but physically wrong, such as selecting the same part of a similar-looking instance, echoing prior findings on VLM hallucination~\citep{pope,chen2024multiobject,hallusionbench}.

This gap reflects a limitation of direct prediction: a VLM reasons over two full images without the ability to inspect finer details or verify a candidate location.
We introduce \themodel, a VLM agent that turns an initial correspondence estimate into a prediction verified against visual evidence (Fig.~\ref{fig:teaser}; Sec.~\ref{sec:method}).
The VLM first predicts point coordinates and visibility before seeing optional proposals from expert vision models.
Its understanding of the scene then guides which magnified views, visual comparisons, and geometric measurements it requests, giving it access to the evidence needed for precise localization.
Expert outputs serve as additional evidence: the VLM can revise both coordinates and visibility and depart from every proposal, including by recovering a point that a tracker declares invisible.
Agentic inference improves correspondence for every backbone evaluated with \themodel, most at strict localization tolerances.
With GPT-6 Astra, \themodel surpasses all evaluated expert vision models on all four metrics under matched two-image and full-video input settings (Tab.~\ref{tab:agent-comparison}).
Some GPT-6 Astra predictions locate the queried physical point more accurately than the reference human annotations~\citep{tapvid}.

This also enables continuous point trajectories in long real-world videos.
Growing efforts in robot learning use point tracking for manipulation~\citep{robotap} and policy learning from egocentric human demonstrations~\citep{egomimic,egozero}.
Reliable 3D trajectory annotations could turn egocentric recordings into beneficial supervision for learning how humans interact with the physical world.
However, existing trackers often fail to reconnect points after disappearance, fragmenting trajectories and limiting supervision for long-horizon interactions.
Our pipeline uses trackers to follow visible points and \themodel to recover the same physical points when they reappear, connecting their earlier and later tracks.
On a manually annotated sample of reappearing points, \themodel with GPT-6 Astra achieves 76.9 APD versus 40.1 for the best evaluated tracker (Tab.~\ref{tab:reappearance-human-agreement}).
The resulting \thedataset provides 3D trajectory annotations for egocentric recordings substantially longer than the clips in existing real-video datasets (Sec.~\ref{sec:egotap4d}; Tab.~\ref{tab:dataset-comparison}).
Adding \thedataset to D4RT~\citep{d4rt} training improves average 3D tracking performance (Tab.~\ref{tab:d4rt-data}).

In summary, our work makes three contributions:
(1) \thebench, a diverse cross-domain benchmark that characterizes the emerging crossover between general-purpose VLMs and expert vision models, together with the remaining gaps in localization, visibility, and generalization;
(2) \themodel, a VLM agent that improves correspondence across all evaluated backbones and, with GPT-6 Astra, surpasses all evaluated expert vision models on all four metrics; and
(3) \thedataset, a 3D point tracking dataset from long real-world videos that improves D4RT performance, together with its \mbox{\themodel-based} annotation pipeline.

\let\input\arxivinput
\par\vspace{0pt plus 3\baselineskip}
\setlength{\textfloatsep}{9pt plus 2pt minus 2pt}
\raggedbottom
\section{Related Work}
\label{sec:related-work}

\begin{figure*}[t]
    \centering
    \includegraphics[width=\textwidth,trim=0 24bp 0 0,clip]{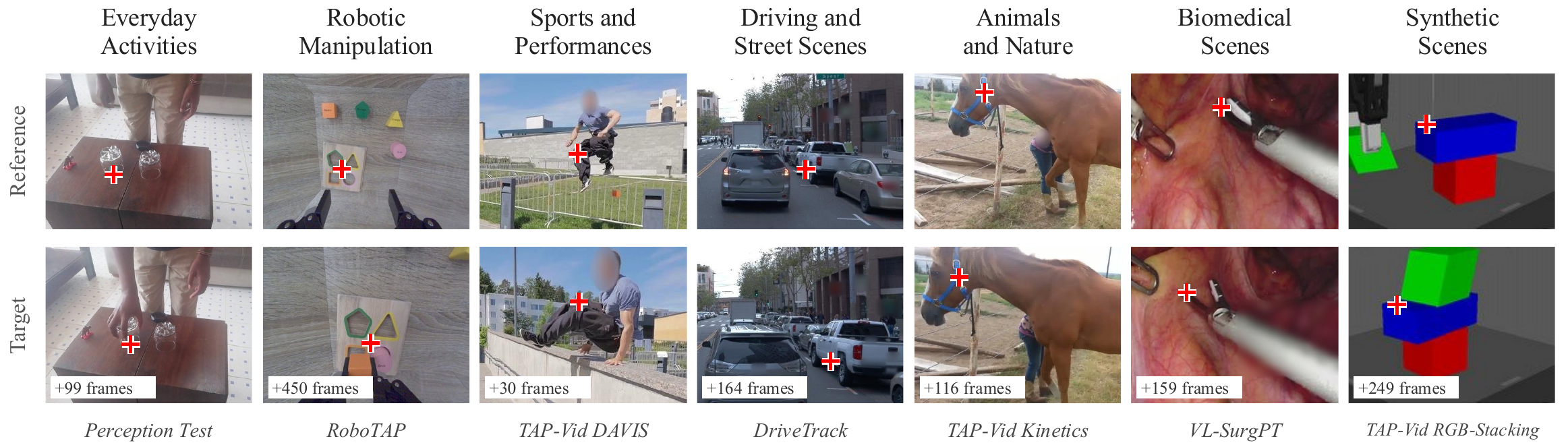}
    \caption{\textbf{Example queries from \thebench.}
    The first six columns show real-video domains~\citep{perceptiontest,robotap,tapvid,drivetrack,vlsurgpt}, and the last shows synthetic data used in the analysis.
    Each column pairs a reference frame (top) with a target frame (bottom).
    Red crosshairs mark the same physical surface point.
    Target crosshairs are shown for illustration only and are not included in model inputs.}
    \label{fig:benchmark-domains}
\end{figure*}

\noindent\textbf{Visual correspondence.}
Image matching establishes correspondences across views using visual descriptors and spatial context~\citep{sift,superglue,loftr,roma,romav2}, and features of large vision models yield correspondences without matching-specific training~\citep{dift,dinov3}.
Point tracking additionally exploits temporal information to maintain physical surface-point identity and estimate visibility through motion and occlusion.
PIPs~\citep{pips} and TAPIR~\citep{tapir} combine per-frame matching with temporal refinement, CoTracker3~\citep{cotracker3} jointly updates tracks across points and time, and AllTracker~\citep{alltracker} extends tracking to dense correspondences between a reference frame and each video frame.
TAPNext~\citep{tapnext,tapnextpp} casts tracking as next-token prediction, and Track-On~\citep{trackon} tracks online with a memory.
Pseudo-labels from real videos improve generalization~\citep{bootstap,cotracker3}, and a learned verifier can select reliable ones among tracker predictions~\citep{aydemir2026}.
Benchmarks on long egocentric and dynamic videos show that points leaving and re-entering the view remain a key failure mode~\citep{egopoints,itto}.
In \themodel, the VLM forms its own correspondence hypothesis, actively selects image regions and visual operations to test it, and revises its prediction. This verification happens at inference and re-identifies points after they reappear, which we use to build \thedataset.

\noindent\textbf{Spatial grounding in VLMs.}
VLMs ground language to image regions~\citep{groundhog,glamm} and predict points for object and affordance localization~\citep{molmo,robopoint}.
Molmo2~\citep{molmo2-report} extends pointing to video and object tracking, while MolmoPoint~\citep{molmopoint} improves localization through dedicated grounding tokens.
Evaluation spans language-guided pointing~\citep{pointarena}, candidate correspondence selection~\citep{blink,mmtbench}, and semantic matching across object instances~\citep{soco}.
CrossPoint-Bench~\citep{crosspoint} evaluates physical point correspondence and visibility in indoor scenes, scoring pointing by whether a prediction falls within the target mask.
Its CroPond model is fine-tuned on geometrically generated correspondence pairs, whereas \themodel reaches precise correspondence with general-purpose VLMs through agentic inference alone.

\noindent\textbf{Agentic visual reasoning.}
Visual programming connects language reasoning to executable vision tools~\citep{vipergpt,visprog}, while Visual Sketchpad~\citep{visualsketchpad} enables iterative reasoning with visual feedback.
V*~\citep{vstar}, ZoomEye~\citep{zoomeye}, and DeepEyes~\citep{deepeyes} make visual inspection adaptive through search, zooming, or learned policies.
ACTrack~\citep{actrack} coordinates expert vision models for object tracking, invoking a VLM under persistent disagreement to arbitrate between candidate boxes.
AgentSTAR~\citep{agentstar} combines VLM reasoning with numerical optimization for shape tracking: the VLM guides pose searches and inspects renderings, while an optimizer produces precise poses.
In \themodel, the VLM itself predicts and revises the point's coordinates and visibility. Tools only supply evidence, so its answer can depart from any proposal.

\par\ifdim\pagetotal>0.5\pagegoal\newpage\fi
\flushbottom
\section{\thebench}
\label{sec:benchmark}

\thebench contains 2,000 two-frame point-correspondence queries
selected from public point-tracking benchmarks, covering 272 video clips across six real-video domains
(Fig.~\ref{fig:benchmark-domains}).
Because public videos may appear in VLM training data, we also capture 15 new videos and annotate a held-out set of 1,814 queries to check that results on \thebench do not reflect data leakage (Sec.~\ref{sec:benchmark-curation}).

\subsection{Problem Setup}
\label{sec:benchmark-setup}
Each query consists of a source image $I_s$, a marked visible point $p_s$, and an unmarked target image $I_t$.
A method predicts whether the same physical surface point is visible in $I_t$, giving a visibility label $\widehat{v}_t\in\{0,1\}$.
If predicted visible, it also returns the point's 2D image coordinates $\widehat{p}_t$. Otherwise, no position is required.
Both the camera and surface may move, so the task requires locating the same surface point, not merely recognizing the same object or part.
A point is invisible if it is occluded, lies outside the image, or belongs to a surface facing away from the camera, even when the object itself remains visible.

\subsection{Benchmark Curation}
\label{sec:benchmark-curation}

\noindent\textbf{Source videos.}
We pool clips from public point-tracking benchmarks~\citep{tapvid,kubric,tapvid3d,drivetrack,egopoints,robotrack,perceptiontest,itto,votsp2026,robotap,vlsurgpt,tapformer} to cover diverse scenes and motions.
Uniform sampling would overrepresent larger sources and their recurring recording conditions.
We therefore allocate source quotas based on visual diversity measured with DINOv3 embeddings~\citep{dinov3}.
Within each source, we cluster clips by appearance, motion, and visibility, then select those nearest the cluster centers.

\noindent\textbf{Benchmark queries.}
We construct a pool of 14,896 queries: 12,496 from real videos and 2,400 from synthetic videos.
For each annotated track, we pair its first visible observation and, when available, one randomly sampled later visible observation with later labeled target frames.
We include invisible targets, exclude missing annotations, and balance displacement, frame gap, and target visibility.
The synthetic queries are used only for real--synthetic comparisons in the analysis.
To reduce the computational cost and runtime of evaluation, we select 2,000 queries from the real-video pool as the main benchmark, stratified by domain, target visibility, whole-track motion, and frame gap.
Across evaluated VLMs, scores and rankings on the 2,000-query benchmark closely agree with those on the full 12,496-query real-video pool.

\noindent\textbf{Held-out annotations.}
We independently capture and annotate 15 in-the-wild videos in three stages:
(1) an annotator selects a visible source point and labels its position or invisibility in target frames;
(2) a second annotator labels the same frames without seeing the first answers;
(3) a third annotator labels disputed targets without seeing either earlier answer.
Two labels agree if both mark the point invisible or both mark it visible within four pixels at the evaluation scale.
For disputed targets, the third label and an agreeing earlier label determine the final annotation.
We average agreeing visible coordinates and exclude unresolved targets and unusable tracks, retaining 1,814 queries.
\mbox{Videos and labels remain hidden.}

\subsection{Evaluation Protocol}
\label{sec:benchmark-protocol}

All main comparisons and ablations use the same 2,000 queries.
Two-image methods receive only the source and target images, while full-video methods can also access other frames.
In direct prediction, a VLM reasons over the inputs and answers in a single call without tools, whereas agents may inspect images and use tools over multiple steps.
Missing or invalid answers count as failures.
\begin{figure*}[!t]
    \centering
    \includegraphics[width=\textwidth]{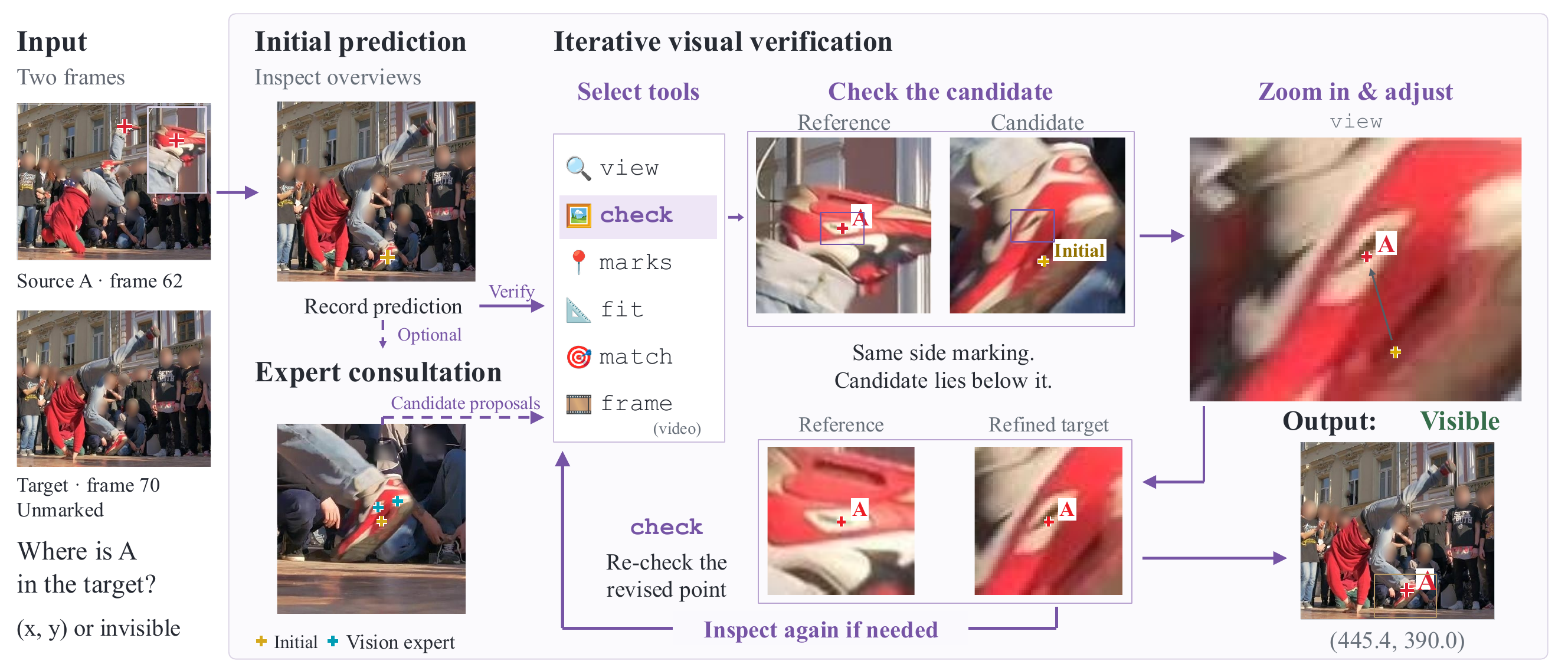}\caption{\textbf{\themodel.} (1) Initial prediction: the VLM agent inspects both views and records its own prediction. (2) Optional expert consultation: expert vision models propose candidate locations. (3) Iterative visual verification: the agent selects visual tools to test its hypothesis, revises the point's location and visibility, and re-checks as needed.}
    \label{fig:agent}
\end{figure*}

\noindent\textbf{Metrics.}
Following TAP-Vid~\citep{tapvid}, we evaluate on a normalized $256\times256$ grid.
We report APD for localization accuracy, AJ for joint localization and visibility, and OA for visibility accuracy.
APD and AJ are averaged over distance thresholds of 1, 2, 4, 8, and 16 pixels.
Unlike TAP-Vid's position-only accuracy, our APD additionally requires a visible prediction.
EPE averages position errors on ground-truth-visible targets, capped at 32 pixels.
All metrics are pooled across queries. Curation, annotation, and evaluation details appear in supplementary Sec.~\ref{sec:supp-trackingbench}.

\section{\themodel}
\label{sec:method}

\subsection{Overview}
\label{sec:method-overview}

Human visual reasoning draws on an understanding of surfaces and motion to guide where to look and which details to compare.
Motivated by this interaction between reasoning and observation, \themodel lets a VLM independently predict point coordinates and visibility, then actively test its prediction (Fig.~\ref{fig:agent}).
The VLM selects visual tools (Sec.~\ref{sec:method-tools}) to inspect image regions, compare local details, and refine the point's location and visibility.
Each observation informs both the prediction and what to inspect next.
Expert vision models optionally provide candidate locations for verification, while the VLM determines the final correspondence.
The procedure requires no \mbox{correspondence-specific VLM fine-tuning}.
\begin{table*}[!t]
\centering
\fontsize{7.0}{8.4}\selectfont
\setlength{\tabcolsep}{0.7pt}
\renewcommand{\arraystretch}{1.0}
\newcommand{\domaindelta}[2]{\raisebox{-0.4ex}{\fontsize{4.8}{5.2}\selectfont\textcolor{#1}{#2}}}
\begin{tabular*}{\textwidth}{@{\extracolsep{\fill}}ll*{22}{c}@{}}
\toprule
& & \multicolumn{4}{c}{\textbf{Overall}} & \multicolumn{3}{c}{\shortstack{\textbf{Everyday}\\\textbf{Activities}}} & \multicolumn{3}{c}{\shortstack{\textbf{Robotic}\\\textbf{Manipulation}}} & \multicolumn{3}{c}{\shortstack{\textbf{Sports and}\\\textbf{Performances}}} & \multicolumn{3}{c}{\shortstack{\textbf{Driving and}\\\textbf{Street Scenes}}} & \multicolumn{3}{c}{\shortstack{\textbf{Animals}\\\textbf{and Nature}}} & \multicolumn{3}{c}{\shortstack{\textbf{Biomedical}\\\textbf{Scenes}}} \\
\cmidrule(lr){3-6}\cmidrule(lr){7-9}\cmidrule(lr){10-12}\cmidrule(lr){13-15}\cmidrule(lr){16-18}\cmidrule(lr){19-21}\cmidrule(lr){22-24}
\textbf{Model} & \textbf{Harness} & AJ & APD & OA & EPE$\downarrow$ & AJ & APD & OA & AJ & APD & OA & AJ & APD & OA & AJ & APD & OA & AJ & APD & OA & AJ & APD & OA \\
\midrule
\rowcolor{black!7}\multicolumn{24}{l}{\textbf{Two images}} \\
\rowcolor{tabband}\multicolumn{24}{l}{\emph{Expert vision models}} \\
RoMa v2~\citep{romav2} &  & 33.9 & 51.0 & 77.6 & 11.2 & 37.3 & 52.2 & 83.3 & 37.5 & 56.9 & 75.5 & 27.2 & 45.1 & 75.6 & 38.6 & 55.3 & 68.7 & 24.6 & 42.4 & 72.6 & 35.4 & 50.5 & \thirdbest{94.0} \\
AllTracker~\citep{alltracker} &  & 46.1 & 57.3 & 73.3 & 10.2 & 49.4 & 58.8 & 77.3 & 46.6 & 59.3 & 72.2 & 46.4 & 57.4 & 73.2 & 43.7 & 57.3 & 66.3 & 36.5 & 49.5 & 69.2 & 43.4 & 53.2 & 78.0 \\
AllTracker, HR &  & 47.9 & 56.6 & 70.8 & 11.2 & 49.7 & 56.2 & 71.3 & 49.1 & 59.0 & 71.9 & 47.9 & 56.5 & 73.0 & 48.5 & 61.2 & 67.0 & 39.8 & 49.7 & 67.3 & 47.0 & 54.1 & 76.0 \\
\rowcolor{tabband}\multicolumn{24}{l}{\emph{VLMs and agents}} \\
\multirow{2}{*}{Qwen3.8-27B~\citep{qwen3.8-27b}} & Direct & 14.9 & 26.0 & 80.1 & 14.7 & 14.8 & 24.8 & 87.6 & 13.4 & 24.6 & 73.5 & 14.1 & 25.5 & 77.2 & 18.9 & 31.9 & 72.0 & 12.9 & 24.3 & 76.4 & 19.6 & 31.8 & 90.0 \\
 & \textbf{\themodel} & 38.6\domaindelta{pos}{+23.7} & 57.5\domaindelta{pos}{+31.4} & 79.6\domaindelta{black!45}{$-$0.5} & 8.0\domaindelta{pos}{$-$6.7} & 40.4 & 56.5 & 85.8 & 38.9 & 60.3 & 73.8 & 37.1 & 57.2 & 75.3 & 41.0 & 62.3 & 77.3 & 30.4 & 50.9 & 73.6 & 45.2 & 59.1 & 92.0 \\
\multirow{2}{*}{Gemma-4-31B-it~\citep{gemma4-31b-it}} & Direct & 17.6 & 28.5 & 76.6 & 13.7 & 16.4 & 25.4 & 80.4 & 16.9 & 28.4 & 71.2 & 21.1 & 35.0 & 75.9 & 19.7 & 31.9 & 71.7 & 14.0 & 24.8 & 75.0 & 20.6 & 31.8 & 92.0 \\
 & \textbf{\themodel} & 32.3\domaindelta{pos}{+14.7} & 50.8\domaindelta{pos}{+22.3} & 79.0\domaindelta{pos}{+2.4} & 10.0\domaindelta{pos}{$-$3.6} & 31.7 & 47.9 & 84.2 & 36.2 & 58.8 & 74.2 & 33.1 & 52.5 & 76.6 & 34.8 & 55.4 & 73.7 & 25.7 & 44.1 & 76.0 & 29.7 & 45.5 & 90.0 \\
\multirow{2}{*}{Inkling 2 Small~\citep{inkling2-small}} & Direct & 20.9 & 33.4 & 79.1 & 12.5 & 19.9 & 30.0 & 84.8 & 20.1 & 33.8 & 74.8 & 17.8 & 30.2 & 76.4 & 30.7 & 48.8 & 74.7 & 19.7 & 32.2 & 74.0 & 18.8 & 29.1 & 86.0 \\
 & \textbf{\themodel} & 43.8\domaindelta{pos}{+22.9} & 64.1\domaindelta{pos}{+30.8} & 81.5\domaindelta{pos}{+2.4} & 6.1\domaindelta{pos}{$-$6.5} & 46.8 & 63.6 & 87.9 & 43.3 & 66.3 & 77.2 & 42.2 & 64.3 & 78.2 & 45.7 & 70.1 & 74.7 & 34.3 & 55.7 & 76.9 & 46.0 & 60.9 & \thirdbest{94.0} \\
\multirow{2}{*}{Muse Wonder Full~\citep{muse-wonder-full}} & Direct & 29.5 & 42.7 & 83.0 & 8.9 & 31.5 & 42.7 & 87.6 & 23.9 & 37.2 & 78.5 & 28.1 & 41.1 & \secondbest{83.2} & 35.6 & 53.1 & 75.7 & 25.1 & 39.6 & \thirdbest{80.3} & 28.1 & 38.6 & \thirdbest{94.0} \\
 & \textbf{\themodel} & 51.4\domaindelta{pos}{+21.9} & 69.2\domaindelta{pos}{+26.5} & 85.4\domaindelta{pos}{+2.4} & 4.9\domaindelta{pos}{$-$4.0} & 53.2 & 67.9 & 90.5 & 51.5 & 71.9 & \secondbest{82.1} & 50.4 & 67.7 & \secondbest{83.2} & 54.4 & \secondbest{77.1} & \thirdbest{79.7} & \thirdbest{42.1} & 62.5 & \secondbest{80.8} & \thirdbest{55.6} & \secondbest{68.6} & \secondbest{96.0} \\
\multirow{2}{*}{Claude Opus 5.5~\citep{claude-opus5.5}} & Direct & 33.0 & 49.8 & 82.2 & 7.8 & 34.2 & 48.8 & 89.2 & 28.5 & 45.8 & 74.5 & 32.3 & 49.0 & 79.5 & 40.4 & 61.9 & 75.3 & 27.0 & 44.7 & 79.3 & 35.2 & 49.1 & \secondbest{96.0} \\
 & \textbf{\themodel} & 49.3\domaindelta{pos}{+16.2} & 68.2\domaindelta{pos}{+18.4} & 84.2\domaindelta{pos}{+2.0} & 5.1\domaindelta{pos}{$-$2.8} & 51.0 & 66.5 & 90.9 & 49.4 & 72.0 & 78.1 & 47.9 & 67.4 & \thirdbest{81.9} & 52.9 & 76.1 & 77.7 & 39.5 & 61.0 & 78.8 & \secondbest{56.7} & \secondbest{68.6} & \best{98.0} \\
\multirow{2}{*}{Claude Fable 5.2~\citep{claude-fable5.2}} & Direct & 35.6 & 52.7 & 83.3 & 6.9 & 36.7 & 50.7 & 89.6 & 33.0 & 52.4 & 77.5 & 34.7 & 51.8 & 80.6 & 41.5 & 63.5 & 78.7 & 29.2 & 47.5 & 76.9 & 40.3 & 54.5 & \secondbest{96.0} \\
 & \textbf{\themodel} & \thirdbest{52.7\domaindelta{pos}{+17.1}} & \best{71.7\domaindelta{pos}{+19.0}} & \thirdbest{86.0\domaindelta{pos}{+2.7}} & \best{4.1\domaindelta{pos}{$-$2.8}} & \thirdbest{54.8} & \thirdbest{70.0} & \best{92.9} & \secondbest{53.7} & \best{76.7} & 79.5 & \thirdbest{52.0} & \best{72.1} & \best{84.8} & \thirdbest{54.7} & \best{77.4} & \secondbest{80.0} & \thirdbest{42.1} & \secondbest{63.0} & 78.4 & \best{57.7} & \best{70.9} & \secondbest{96.0} \\
\multirow{2}{*}{GPT-6.1 Sol~\citep{gpt6.1-sol}} & Direct & 40.6 & 56.8 & 83.7 & 6.8 & 41.5 & 54.9 & 88.9 & 42.7 & 62.6 & 78.8 & 38.7 & 54.7 & \thirdbest{81.9} & 47.1 & 68.6 & 78.0 & 31.1 & 46.7 & \thirdbest{80.3} & 37.7 & 50.5 & \thirdbest{94.0} \\
 & \textbf{\themodel} & \secondbest{53.6\domaindelta{pos}{+13.0}} & \thirdbest{70.8\domaindelta{pos}{+14.0}} & \secondbest{86.1\domaindelta{pos}{+2.4}} & \thirdbest{4.7\domaindelta{pos}{$-$2.1}} & \secondbest{56.3} & \secondbest{70.5} & \secondbest{91.7} & \thirdbest{52.8} & \thirdbest{73.8} & \thirdbest{81.5} & \secondbest{53.7} & \thirdbest{70.7} & \best{84.8} & \secondbest{55.3} & 75.7 & \best{80.7} & \secondbest{43.3} & \thirdbest{62.9} & 79.8 & 53.5 & \thirdbest{65.9} & \secondbest{96.0} \\
\multirow{2}{*}{GPT-6 Astra~\citep{gpt6-astra}} & Direct & 43.2 & 59.4 & 85.5 & 6.5 & 43.9 & 57.3 & 89.7 & 43.7 & 63.4 & 80.1 & 42.2 & 57.5 & \best{84.8} & 49.9 & 71.7 & \thirdbest{79.7} & 34.9 & 51.1 & \best{85.1} & 38.3 & 51.4 & \thirdbest{94.0} \\
 & \textbf{\themodel} & \best{54.9\domaindelta{pos}{+11.7}} & \secondbest{71.4\domaindelta{pos}{+12.0}} & \best{86.3\domaindelta{pos}{+0.8}} & \secondbest{4.6\domaindelta{pos}{$-$1.9}} & \best{57.0} & \best{70.6} & \thirdbest{91.2} & \best{56.1} & \secondbest{75.8} & \best{83.8} & \best{55.0} & \secondbest{71.2} & \best{84.8} & \best{55.8} & \thirdbest{76.5} & \best{80.7} & \best{44.9} & \best{63.5} & \secondbest{80.8} & 54.5 & 65.5 & \thirdbest{94.0} \\
\midrule
\rowcolor{black!7}\multicolumn{24}{l}{\textbf{Full video}} \\
\rowcolor{tabband}\multicolumn{24}{l}{\emph{Expert vision models}} \\
CoTracker3~\citep{cotracker3} &  & 49.4 & 64.0 & 81.3 & 7.2 & 49.9 & 62.2 & 84.3 & 51.1 & 68.9 & 80.1 & 48.0 & 61.3 & 80.3 & 52.5 & 73.5 & 80.0 & 44.0 & 58.1 & 73.1 & 46.4 & 58.6 & 92.0 \\
BootsTAPIR~\citep{bootstap} &  & 47.8 & 59.6 & 76.3 & 8.6 & 49.0 & 58.8 & 78.8 & 49.8 & 62.8 & 76.2 & 47.5 & 59.1 & 78.5 & 49.0 & 64.2 & 68.0 & 40.1 & 52.9 & 72.1 & 48.5 & 60.0 & 90.0 \\
TAPNext++~\citep{tapnextpp} &  & 48.1 & 63.3 & 83.6 & 6.8 & 51.8 & 64.2 & 87.1 & 47.1 & 67.0 & \thirdbest{83.4} & 53.5 & 68.0 & 84.3 & 43.0 & 60.1 & 75.0 & 34.9 & 51.6 & 78.4 & 47.9 & 60.9 & \best{100.0} \\
AllTracker~\citep{alltracker} &  & 53.3 & 66.8 & 81.5 & 6.2 & 54.7 & 65.8 & 85.0 & 57.4 & 74.2 & 82.8 & 52.7 & 66.5 & 81.9 & 53.2 & 70.0 & 76.0 & 44.5 & 58.6 & 73.6 & 50.6 & 60.0 & 82.0 \\
AllTracker, HR &  & \secondbest{56.1} & 67.9 & 80.2 & 6.9 & \thirdbest{57.4} & 67.3 & 83.5 & \thirdbest{57.6} & 71.7 & 79.5 & \secondbest{57.2} & 69.5 & 81.9 & \secondbest{56.1} & 71.6 & 75.7 & \thirdbest{47.3} & 58.9 & 72.1 & 54.6 & 61.4 & 82.0 \\
\rowcolor{tabband}\multicolumn{24}{l}{\emph{VLMs and agents}} \\
Inkling 2 Small~\citep{inkling2-small} & \textbf{\themodel} & 48.5\domaindelta{pos}{+27.5} & 68.5\domaindelta{pos}{+35.1} & 83.2\domaindelta{pos}{+4.1} & 5.2\domaindelta{pos}{$-$7.3} & 50.6 & 67.4 & 89.3 & 48.9 & 72.7 & 78.5 & 47.6 & 68.6 & 80.3 & 50.1 & 73.2 & 75.7 & 39.5 & 61.4 & \thirdbest{80.3} & 50.5 & 64.1 & \thirdbest{96.0} \\
Muse Wonder Full~\citep{muse-wonder-full} & \textbf{\themodel} & \thirdbest{55.5\domaindelta{pos}{+26.1}} & \thirdbest{72.8\domaindelta{pos}{+30.1}} & \thirdbest{86.6\domaindelta{pos}{+3.6}} & \secondbest{4.1\domaindelta{pos}{$-$4.7}} & 57.3 & \thirdbest{71.4} & 90.8 & 56.2 & 76.3 & \secondbest{84.8} & 55.3 & \thirdbest{72.8} & \secondbest{85.3} & 55.3 & \secondbest{77.4} & \thirdbest{80.7} & \secondbest{48.2} & \best{67.7} & \secondbest{81.7} & \best{59.3} & \best{70.9} & \secondbest{98.0} \\
Claude Opus 5.5~\citep{claude-opus5.5} & \textbf{\themodel} & 53.2\domaindelta{pos}{+20.2} & 72.0\domaindelta{pos}{+22.2} & 84.3\domaindelta{pos}{+2.1} & \thirdbest{4.3\domaindelta{pos}{$-$3.5}} & 56.0 & 71.1 & \thirdbest{91.6} & 54.6 & \thirdbest{77.7} & 78.1 & 52.3 & 71.7 & 82.2 & 52.8 & 75.3 & 77.7 & 43.2 & 64.7 & 76.9 & \thirdbest{57.4} & \secondbest{70.5} & \thirdbest{96.0} \\
Claude Fable 5.2~\citep{claude-fable5.2} & \textbf{\themodel} & \secondbest{56.1\domaindelta{pos}{+20.5}} & \best{74.4\domaindelta{pos}{+21.7}} & \secondbest{86.7\domaindelta{pos}{+3.5}} & \best{3.6\domaindelta{pos}{$-$3.3}} & \secondbest{58.3} & \secondbest{72.9} & \best{92.8} & \secondbest{57.9} & \best{79.8} & 82.8 & \thirdbest{56.6} & \best{76.1} & \thirdbest{85.0} & \thirdbest{55.6} & \best{77.5} & \secondbest{81.0} & 46.0 & \secondbest{67.3} & 78.8 & \secondbest{58.6} & \thirdbest{70.0} & \secondbest{98.0} \\
GPT-6 Astra~\citep{gpt6-astra} & \textbf{\themodel} & \best{58.5\domaindelta{pos}{+15.3}} & \secondbest{73.9\domaindelta{pos}{+14.5}} & \best{88.2\domaindelta{pos}{+2.7}} & \secondbest{4.1\domaindelta{pos}{$-$2.4}} & \best{59.8} & \best{73.3} & \secondbest{92.5} & \best{60.2} & \secondbest{78.7} & \best{86.1} & \best{58.8} & \secondbest{74.2} & \best{86.6} & \best{58.3} & \thirdbest{76.8} & \best{82.0} & \best{51.1} & \thirdbest{66.6} & \best{85.1} & 56.9 & 68.6 & \thirdbest{96.0} \\
\bottomrule
\end{tabular*}
\caption{\textbf{Main results on \thebench.} Direct prediction, \themodel, and expert vision models answer the same 2,000 queries. Overall columns evaluate all queries together rather than averaging the six domains, and domain columns omit EPE. Higher is better except for EPE. HR: high resolution. Subscripts give each agent's change from the same VLM's two-image direct prediction, which is also the reference for full-video agents. \textcolor{pos}{Green} marks an improvement and \textcolor{black!45}{gray} a decline. Shading marks the \colorbox{tabfirst}{best}, \colorbox{tabsecond}{second}, and \colorbox{tabthird}{third} results per column within each input block.}
\label{tab:agent-comparison}
\label{tab:vlm-2k}
\end{table*}

\subsection{Visual Tools}
\label{sec:method-tools}

The VLM receives descriptions of the tools in Fig.~\ref{fig:agent} and chooses which to invoke and their inputs.

\noindent\textbf{Image inspection.}
With \toolname{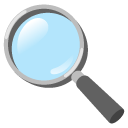}, the VLM chooses regions and scales to inspect scene context, texture, and boundaries, zooming in or out and enhancing contrast as needed.
Grids and point overlays use original image coordinates.
The \toolname{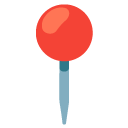} tool returns source-marker coordinates.

\noindent\textbf{Correspondence checking.}
The \toolname{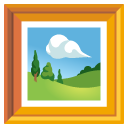} tool displays source and candidate target crops side by side and as an overlay.
Landmarks and boundaries help assess surface identity and spatial offsets.
Rechecking revised coordinates tests whether refinement preserves the queried surface point.

\noindent\textbf{Coordinate estimation.}
The VLM can call \toolname{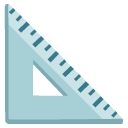} with landmark correspondences it identifies in the source and target images.
The tool fits an affine transformation or homography to these pairs and maps the source point to a target coordinate.
This estimate assumes that the landmarks and query point share a local transformation.
Fitting residuals and a visualization help the VLM assess the resulting estimate.
The \toolname{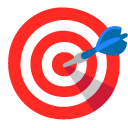} tool uses template matching or SIFT~\citep{sift} within a VLM-selected target region, returning candidates \mbox{for further visual verification}.

\noindent\textbf{Temporal inspection.}
With full-video input, \toolname{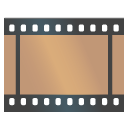} shows VLM-selected intermediate frames and regions.
Surface motion and orientation help resolve point identity and visibility.
Supplementary Sec.~\ref{sec:supp-agentap} provides tool descriptions \mbox{and additional visualization options}.

\subsection{Agentic Inference}
\label{sec:method-inference}

\noindent\textbf{Initial prediction.}
In Stage 1 of Fig.~\ref{fig:agent}, the agent inspects source and target overviews, identifies the queried surface, and records a position or an invisible decision. This prediction precedes expert proposals and intermediate-frame inspection, preserving the agent's independent estimate.

\noindent\textbf{Optional expert consultation.}
When enabled, expert proposals provide alternative correspondence hypotheses for visual verification (Stage 2 of Fig.~\ref{fig:agent}).
The VLM may accept, refine, or reject these proposals, and its final answer need not \mbox{match any supplied candidate}.

\noindent\textbf{Iterative visual verification.}
The VLM tests correspondence hypotheses against visual evidence, choosing subsequent inspections based on what it observes.
The prompt specifies the verification objective and available tools, and the VLM selects each operation and its inputs.

For a visible candidate, the agent compares its position relative to nearby landmarks with that of the source point, as in Stage 3 of Fig.~\ref{fig:agent}.
If the surface correspondence \mbox{appears correct} but the position remains uncertain, the VLM can adjust the coordinate directly or use \toolname{fit} with landmark correspondences.
Crops and geometric estimates retain native image coordinates.
A location obtained from a magnified view or landmark fit can then be marked on the target and checked again.
In the illustrated example, a nearby side marking guides the correction, which \mbox{the agent then re-checks}.

If the visual evidence instead suggests the wrong object, part, or surface, the VLM can choose to revisit the source image and search for a different correspondence.
The marked source point remains fixed, and the VLM revises its interpretation of the surface carrying that point.
Position and visibility are reconsidered together: further inspection may recover a missed point or reveal that the queried surface is hidden despite a plausible match on a visible surface.

The agent returns a target coordinate or an invisible decision after verification.
Supplementary Sec.~\ref{sec:supp-agentap} details the protocol, prompts, \mbox{and example agent sessions}.

\section{Experiments}
\label{sec:experiments}

\noindent\textbf{Setup.}
We compare \themodel with direct prediction by the same VLM, in which the VLM reasons over the two images and answers without tools, and with five expert vision models: RoMa v2~\citep{romav2}, AllTracker~\citep{alltracker}, CoTracker3~\citep{cotracker3}, BootsTAPIR~\citep{bootstap}, and TAPNext++~\citep{tapnextpp}.
With two images, \themodel receives RoMa v2 and AllTracker proposals. With full video, it receives high-resolution AllTracker proposals and intermediate frames.
Settings, prompts, and computational costs are in supplementary Sec.~\ref{sec:supp-agentap}.

\begin{figure*}[!t]
    \centering
    \includegraphics[width=\textwidth]{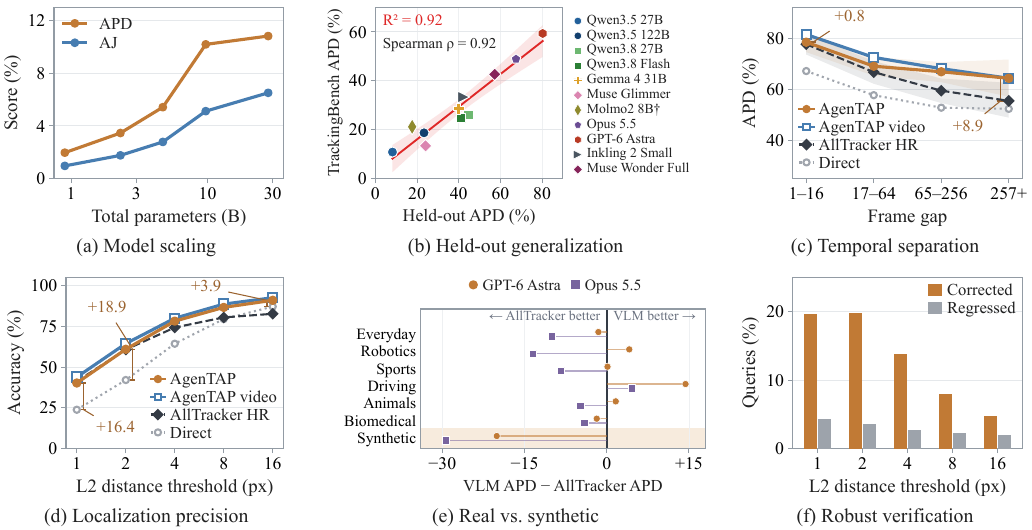}
    \caption{\textbf{Analysis of VLM correspondence.}
    Direct prediction improves with model size (a), transfers to held-out videos (b), and trails trackers on synthetic videos (e).
    \themodel's lead over trackers grows with frame gap (c), its improvement over direct prediction is largest at strict thresholds (d), and its verification is robust (f).
    Agent results use GPT-6 Astra. Red: linear fit. Shading: 95\% CIs. $\dagger$: video input.}
    \label{fig:analysis}
\end{figure*}

\begin{figure}[t]
    \centering
    \includegraphics[trim=0 8 0 4,clip,width=\linewidth]{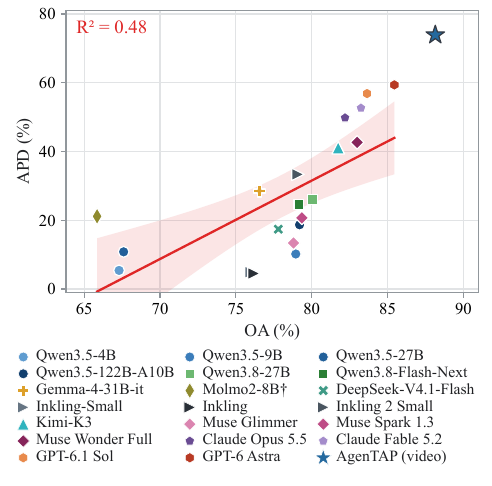}
    \caption{\textbf{Visibility and localization.}
    In direct prediction, VLMs that localize better also judge visibility better.
    $\dagger$: video input.}
    \label{fig:onecall-overview}
\end{figure}

\subsection{Direct Prediction}
\label{sec:direct-prediction}

\noindent\textbf{Visibility and localization.}
VLMs that localize the point better also judge more reliably whether it is still visible (Fig.~\ref{fig:onecall-overview}), suggesting that both draw on the same scene understanding.
Visibility is the harder half.
VLMs rarely call a point hidden: most mark close to nine in ten queries visible, although one in five targets is hidden.
Answering ``visible'' every time would already reach 79.0 OA, and only six VLMs in Fig.~\ref{fig:onecall-overview} beat this by more than two points.
The two skills can still come apart: Inkling, Inkling-Small, and Qwen3.5-9B reach about this OA yet rarely find the exact point, whereas Molmo2, which sees the video, localizes better than its visibility decisions suggest.
The strongest VLM, GPT-6 Astra, already beats every two-image expert model in APD and OA by direct prediction, though not yet in AJ (Tab.~\ref{tab:agent-comparison}).

\noindent\textbf{Model scaling.}
Within the Qwen3.5 dense family, larger models do better, although APD plateaus from 9B to 27B (Fig.~\ref{fig:analysis}a).
Across families, Gemma-4-31B-it beats the much larger Qwen3.5-122B-A10B, so total parameter count alone does not predict correspondence ability.

\noindent\textbf{Held-out generalization.}
Across models, rankings on independently annotated held-out videos closely match those on \thebench (Fig.~\ref{fig:analysis}b).
This agreement supports the benchmark as a reliable evaluation of correspondence ability and eases concerns that strong performance reflects data leakage.
\thebench is also harder: nine of the eleven models score lower on it than on the held-out set.

\subsection{Agentic Inference}
\label{sec:main-results}

\themodel gives the same VLMs visual tools: they still reason, but can now inspect, compare, and verify their answers.

\noindent\textbf{\themodel helps across backbones.}
\themodel improves precision for all backbones (Tab.~\ref{tab:agent-comparison}).
Seven of the eight then beat every two-image expert model in APD, and GPT-6 Astra, from only two frames, beats every full-video tracker in APD and OA.
Given the same input as the expert models, \themodel with GPT-6 Astra beats all of them on all four metrics, including high-resolution AllTracker, whose own predictions it receives as proposals. Claude Fable 5.2 gives the best APD and EPE in both settings.
Applied frame by frame, \themodel turns two-frame correspondences into trajectories that follow the ground-truth motion and keeps points that AllTracker loses (Fig.~\ref{fig:davis-compare}).

\noindent\textbf{\themodel narrows the backbone gap.}
\themodel helps weaker backbones more. Among the six strongest, it adds 30.8 APD to the weakest, Inkling 2 Small, but only 12.0 to the strongest, GPT-6 Astra. Their gap shrinks from 26.0 APD under direct prediction to 7.3 with \themodel: given tools, a much weaker VLM nearly catches up.

\noindent\textbf{Localization precision.}
Without tools, GPT-6 Astra already places 87.1\% of visible points within 16 pixels, but only 23.8\% within one pixel (Fig.~\ref{fig:analysis}d): it finds the right region, not the exact pixel.
\themodel mainly fixes the last few pixels.
Its improvement is small at 16 pixels (+3.9) but large at one- and two-pixel thresholds (+16.4 and +18.9), as magnified views and side-by-side checks reveal details that are lost when the VLM reasons over the full images alone.

\noindent\textbf{Temporal separation.}
All methods lose accuracy as the frame gap grows, but high-resolution AllTracker loses the most, so \themodel's lead over it widens at longer gaps (Fig.~\ref{fig:analysis}c).
When a hidden point reappears in the target frame, full-video trackers call it invisible about three times as often as two-image \themodel (supplementary Sec.~\ref{sec:supp-results}).

\noindent\textbf{Robustness of \themodel.}
\themodel fixes many wrong initial answers while rarely breaking correct ones.
From each session's first prediction, made before any expert proposal, to its final answer, corrections by \themodel with GPT-6 Astra far outnumber regressions at every threshold, mostly by moving a visible point to the right location (Fig.~\ref{fig:analysis}f).

\begin{table}[t]
    \centering
    \fontsize{7.0}{8.4}\selectfont
    \tabcolsep=0.1em
    \newcommand{\yes}{$\checkmark$}
    \newcommand{\no}{$\times$}
    \newcommand{\groupband}[2]{\rowcolor{black!7}\multicolumn{9}{l}{\textbf{#1}\ \ \emph{#2}}}
    \newcommand{\oracle}[1]{\textcolor{gray}{#1}}
    \newcommand{\astra}{\scriptsize GPT-6 Astra}
    \newcommand{\inkling}{\scriptsize Inkling2 S}
    \begin{tabular*}{\linewidth}{@{\extracolsep{\fill}}llccccccc@{}}
        \toprule
        \multirow{2}{*}{\textbf{Harness}} & \multirow{2}{*}{\textbf{Model}} & \scriptsize Inspection & \scriptsize Vision & \scriptsize Intermediate & \multirow{2}{*}{AJ$\uparrow$} & \multirow{2}{*}{APD$\uparrow$} & \multirow{2}{*}{OA$\uparrow$} & \multirow{2}{*}{EPE$\downarrow$} \\
        & & \scriptsize tools & \scriptsize experts & \scriptsize frames & & & & \\
        \midrule
        Direct & \astra & \no & \no & \no & 43.2 & 59.4 & 85.5 & 6.5 \\
        Direct & \inkling & \no & \no & \no & 20.9 & 33.4 & 79.1 & 12.5 \\
        Zoom harness & \astra & \yes & \no & \no & 43.0 & 60.9 & 85.1 & 6.7 \\
        \midrule
        \groupband{Two images}{} \\
        \oracle{Expert oracle} & \oracle{\textemdash} & \oracle{\no} & \oracle{\yes} & \oracle{\no} & \oracle{48.5} & \oracle{59.7} & \oracle{73.3} & \oracle{9.8} \\
        \themodel & \inkling & \yes & \no & \no & 32.0 & 48.5 & 80.4 & 8.7 \\
        \themodel & \inkling & \yes & \yes & \no & 43.8 & 64.1 & 81.5 & 6.1 \\
        \themodel & \astra & \yes & \no & \no & 51.0 & 67.4 & 86.5 & 5.1 \\
        \themodel & \astra & \yes & \yes & \no & 54.9 & 71.4 & 86.3 & 4.6 \\
        \midrule
        \groupband{Full video}{} \\
        \oracle{Expert oracle} & \oracle{\textemdash} & \oracle{\no} & \oracle{\yes} & \oracle{\yes} & \oracle{57.5} & \oracle{69.3} & \oracle{80.2} & \oracle{6.7} \\
        \themodel & \inkling & \yes & \yes & \yes & 48.5 & 68.5 & 83.2 & 5.2 \\
        \textbf{\themodel} & \astra & \yes & \yes & \yes & \textbf{58.5} & \textbf{73.9} & \textbf{88.2} & \textbf{4.1} \\
        \bottomrule
    \end{tabular*}
    \caption{\textbf{Agentic inference ablation.}
    Tool use helps even without expert proposals. Proposals and intermediate frames add further improvements, most for the weaker backbone.}
    \label{tab:method-ablation}
\end{table}

\subsection{Ablations}
\label{sec:ablations}

Tab.~\ref{tab:method-ablation} adds \themodel's components one at a time, with GPT-6 Astra and the open-weight Inkling 2 Small.

\noindent\textbf{Agentic tool use.}
Agentic tool use helps every model.
A fixed zoom harness, which shows the VLM preset close-ups around its direct prediction, barely helps.
Letting the VLM choose its own inspections instead, still without proposals, improves both backbones and already lets GPT-6 Astra beat every two-image expert model on all metrics.

\noindent\textbf{Expert proposals.}
Expert proposals improve the weaker backbone, Inkling 2 Small, more than GPT-6 Astra.
Both models usually retain a supplied proposal, but GPT-6 Astra is more reliable when revising them: its changes usually improve localization, whereas Inkling 2 Small's help in 53.5\% of cases and hurt in 38.7\%.
To assess whether the agents go beyond proposal selection, we evaluate an oracle that chooses the better-localized RoMa v2 or AllTracker proposal for each point while using AllTracker's visibility.
This oracle reaches 59.7 APD.
GPT-6 Astra exceeds it even without proposals, demonstrating strong independent correspondence ability.
Inkling 2 Small exceeds it only with proposals, indicating greater reliance on external candidates.
Even with the true visibility, it reaches only 53.1 AJ and 67.3 APD, below \themodel's 54.9 and 71.4.

\noindent\textbf{Temporal evidence.}
The VLM can exploit video, mainly to judge whether a point is visible: with two images, \themodel with GPT-6 Astra adds 12.0 APD and 0.8 OA over direct prediction, and with the full video OA reaches 88.2.
With the full video, \themodel improves both backbones and, with GPT-6 Astra, surpasses high-resolution AllTracker, whose predictions it receives as proposals.
When both call a point visible, \themodel almost always keeps the tracker's position. Where they disagree about visibility, \themodel is right four times out of five, mostly on visible points the tracker marks invisible.
The frames between the endpoints add 1.6 AJ and 1.1 OA (supplementary Sec.~\ref{sec:supp-agent-ablations}).

\subsection{Analysis}
\label{sec:analysis}
\begin{figure*}[!t]
    \centering
    \includegraphics[width=\textwidth]{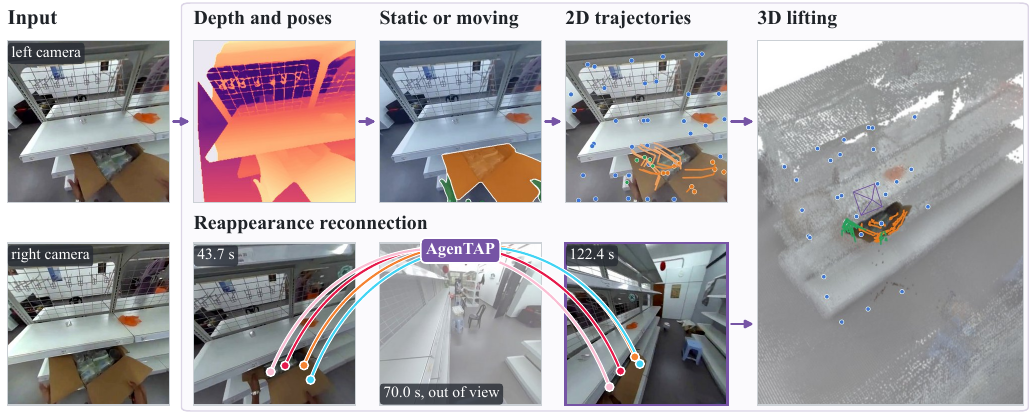}
    \caption{\textbf{\thedataset annotation pipeline.} From a stereo egocentric video with camera poses and hand motion capture, we recover metric depth and camera poses, label each region as static (blue), moving (orange), or hand (green), and build 2D trajectories from the fused scene geometry, AllTracker, and MANO. When an object returns after leaving the view, \themodel with GPT-6 Astra locates its query points again (Fig.~\ref{fig:egotap4d-reconnection}, first row). All trajectories are then lifted to 3D in one world frame.}
    \label{fig:annotation-pipeline}
\end{figure*}

\begingroup
\clubpenalty=10000
\widowpenalty=10000

\noindent\textbf{Real vs.\ synthetic.}
With two-image input, GPT-6 Astra's direct predictions outperform AllTracker in several real-video domains.
On synthetic videos, however, AllTracker clearly outperforms GPT-6 Astra and Opus 5.5 (Fig.~\ref{fig:analysis}e).
Models that perform well on real videos can therefore struggle with synthetic scenes, possibly due to differences in training data.

\noindent\textbf{Improvement by domain.}
\themodel with GPT-6 Astra improves least in driving scenes (+4.8 APD), where direct prediction is strongest, and most in biomedical scenes and sports (+14.1 and +13.7).
It also leads the hardest domain, animals and nature, with 63.5 APD against 49.5 for two-image AllTracker (Tab.~\ref{tab:agent-comparison}).

\noindent\textbf{Semantic-part hallucinations.}
Like humans, VLMs can confuse similar body parts or similar-looking objects as they move.
Even after visual inspection, the agent may place the point on the other eggshell half or gripper finger (\mbox{supplementary Sec.~\ref{sec:supp-results}}).
Surrounding details may also look alike, so a closer look may not reveal the mistake.

\noindent\textbf{Computational cost.}
\themodel reads 141K--184K input tokens per query, and sharing a session among 16 points or three frames saves 56--70\% or 44--52\% of them with similar accuracy (supplementary Secs.~\ref{sec:supp-runtime-cost} and~\ref{sec:supp-scaling}).

\par\endgroup

\ifdefined\arxivbuild\flushbottom
\makeatletter
\setlength{\@dblfptop}{0pt plus 1fil}\setlength{\@dblfpsep}{18pt}\setlength{\@dblfpbot}{0pt plus 1fil}
\makeatother
\fi
\begingroup
\clubpenalty=10000
\widowpenalty=10000
\ifdefined\arxivincludegraphics\else
  \setlength{\textfloatsep}{6pt plus 2pt minus 2pt}
\fi

\section{\thedataset}
\label{sec:egotap4d}

Egocentric video is a stress test for correspondence: as the wearer moves and handles objects, surfaces leave the view, disappear behind hands, and return seconds later.
Expert vision models often lose such points (Sec.~\ref{sec:main-results}), and existing real-video 3D point tracking datasets span only a few seconds per clip (Tab.~\ref{tab:dataset-comparison}).
\thedataset fills this gap with 3D point trajectories over long egocentric recordings, using a VLM to recover each point when it reappears.

\subsection{Annotation Pipeline}
\label{sec:egotap4d-curation}

\begin{table}[t]
    \centering
    \footnotesize
    \tabcolsep=0.25em
    \begin{tabular*}{\linewidth}{@{\extracolsep{\fill}}lccccc@{}}
        \toprule
        \textbf{Dataset} & Duration & Avg.\ clip & Real & Dynamic & Wild \\
        \midrule
        Kubric MOVi-E~\citep{kubric,tapvid} & 21.3\,h & 2.0\,s & $\times$ & $\checkmark$ & $\times$ \\
        PointOdyssey~\citep{pointodyssey} & 1.9\,h & 42\,s & $\times$ & $\checkmark$ & $\times$ \\
        Dynamic Replica~\citep{dynamicstereo} & 1.3\,h & 10\,s & $\times$ & $\checkmark$ & $\times$ \\
        DriveTrack~\citep{drivetrack} & 23.3\,h & 8.4\,s & $\checkmark$ & rigid & $\checkmark$ \\
        Stereo4D~\citep{stereo4d} & 121.2\,h & 4.9\,s & $\checkmark$ & $\checkmark$ & $\checkmark$ \\
        MolmoMotion-1M, real~\citep{molmomotion} & 267.4\,h & 1.6\,s & $\checkmark$ & $\checkmark$ & partial \\
        \midrule
        \textbf{\thedataset} & 297.2\,h & 198\,s & $\checkmark$ & $\checkmark$ & $\checkmark$ \\
        \bottomrule
    \end{tabular*}
    \caption{\textbf{Comparison of 3D point tracking datasets.}
    \thedataset has the longest average video duration among the compared real-video datasets.
    Duration is the total video length.}
    \label{tab:dataset-comparison}
\end{table}

We use different tools for different parts of the annotation process (Fig.~\ref{fig:annotation-pipeline}).
We annotate 3,740 recordings from Xperience-10M~\citep{xperience10m} and 1,659 from Ego-Exo4D~\citep{egoexo4d}, 5,399 videos and 297.2\,h in total, with an average duration of 198\,s.
Depth and camera poses map observations from all frames into a shared metric coordinate system.
After a per-recording calibration of the Xperience-10M stereo depth, its median error stays within $\pm$1\% up to 3\,m (supplementary Sec.~\ref{sec:supp-metric-geometry}).
\emph{Geometry} handles static surfaces: GMOS object masks~\citep{gmos} separate them from moving ones, and every point on the fused static surface projects into every frame, so static points can be sampled anywhere, at any density.
\emph{A tracker} handles moving surfaces while they stay in view: AllTracker~\citep{alltracker} follows them and bridges gaps of up to a second, and motion capture provides hands where available.
\emph{A VLM} handles what both miss.
When an object returns after at least three seconds, \themodel with GPT-6 Astra locates 10 of its points from before the gap in a frame after the return.
We accept a prediction only if it lies on the same object, has valid depth, and is not on a hand, and an AllTracker cross-check on a second post-return frame rejects inconsistent matches.
Of 14,091 reappearance image pairs with 10 points each, 7,057 pass filtering, and 23,072 of their 70,570 query points pass these checks and enter the dataset (supplementary Sec.~\ref{sec:supp-reconnection}). Fig.~\ref{fig:annotation-examples} shows trajectories from six recordings.

\subsection{Annotation Quality}
\label{sec:egotap4d-human-evaluation}
\label{sec:egotap4d-training}
\begin{figure*}[p]
    \centering
    \includegraphics[width=\textwidth]{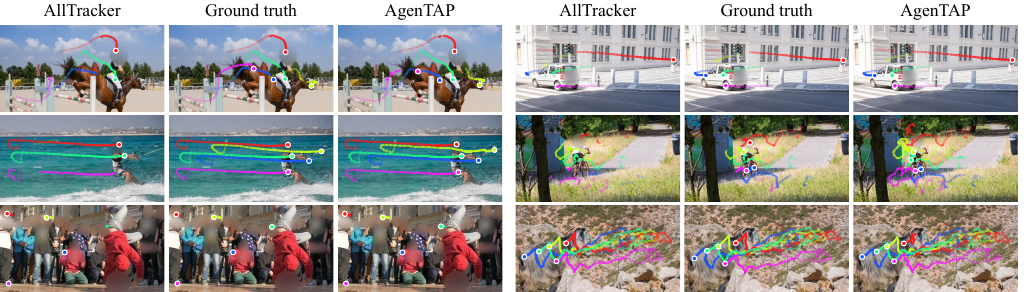}
    \caption{\textbf{\themodel and AllTracker on DAVIS.} Trails show each point from frame 0 to the shown frame. Each answer uses only that frame and frame 0, and \themodel also receives the RoMa v2 and AllTracker answers. A point predicted invisible has no dot.}
    \label{fig:davis-compare}
\end{figure*}

\begin{figure*}[p]
    \centering
    \includegraphics[width=\textwidth,trim=0 152.3 0 0,clip]{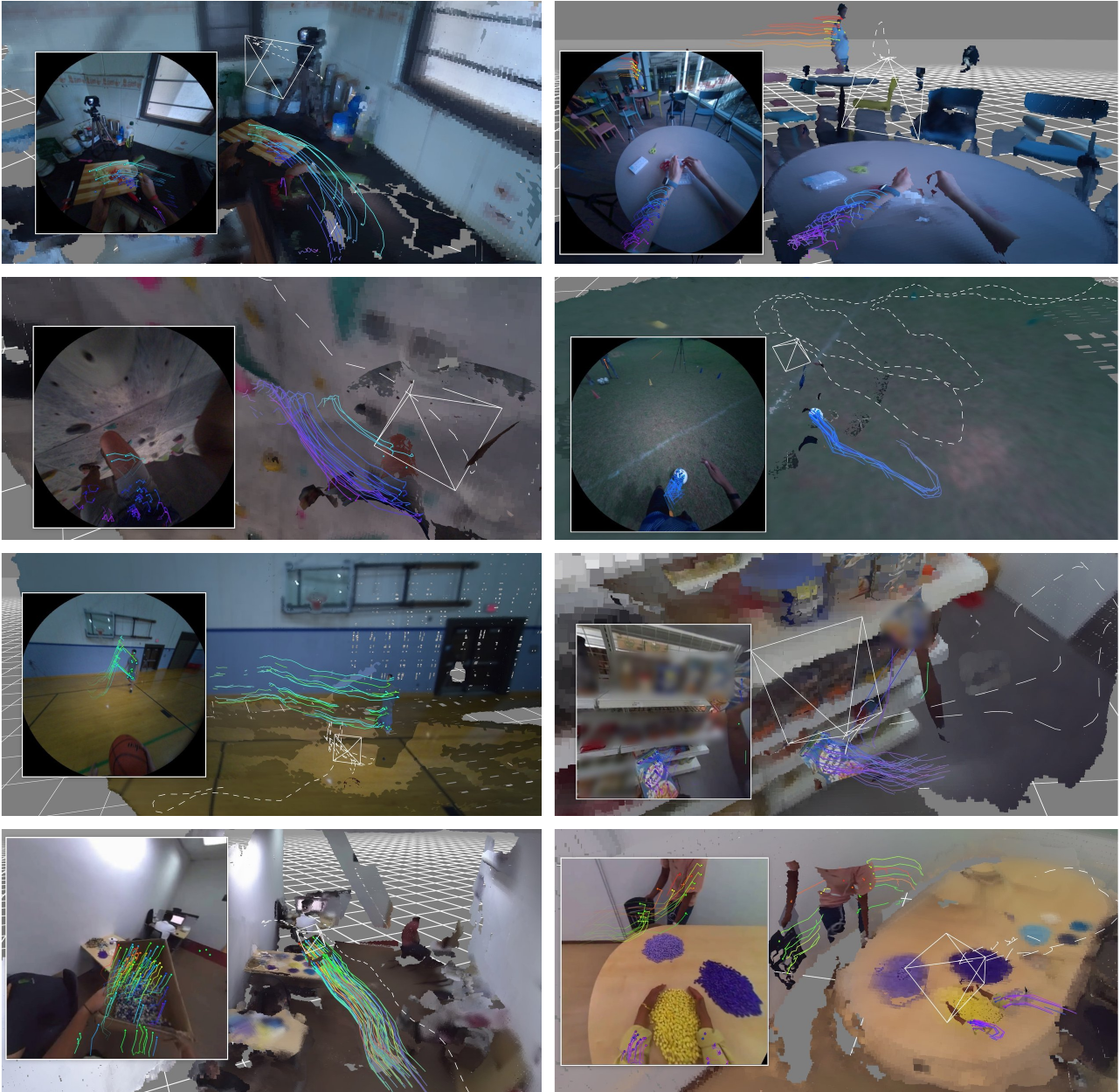}
    \caption{\textbf{\thedataset trajectories.} Each panel shows one recording. The main view shows 3D trajectories in world coordinates together with the reconstructed scene, and the inset shows the corresponding video frame with the 2D trajectories. The same color marks the same track in both views. The white frustum shows the current camera pose, and the dashed white line shows the camera trajectory.}
    \label{fig:annotation-examples}
\end{figure*}

\begin{figure*}[t]
    \centering
    \includegraphics[width=0.92\textwidth,trim=0 233.2bp 0 0,clip]{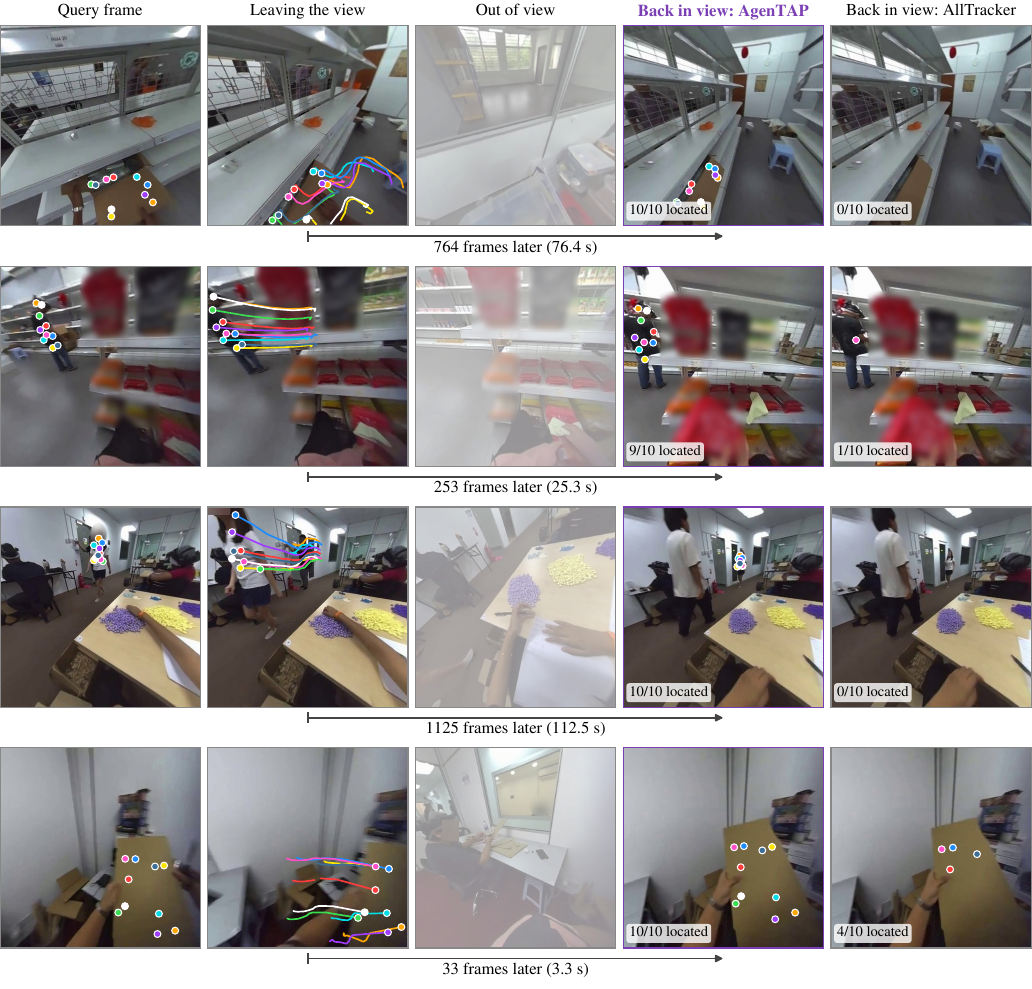}
    \caption{\textbf{Point reconnection after reappearance.} Two Xperience-10M sequences in which an object or a person exits the field of view and later re-enters it. From left to right, the columns show the query points, their tracks shortly before the exit, a frame in which they are out of view (dimmed), and the predictions of \themodel and AllTracker after the return. Colors denote point identity. Points predicted invisible are omitted, and each prediction panel reports how many of the ten query points are predicted visible. The arrow gives the time between exit and re-entry. Supplementary Fig.~\ref{fig:annotation-reappearance} shows two more sequences.}
    \label{fig:egotap4d-reconnection}
\end{figure*}

\begin{table*}[t!]
    \centering
    \footnotesize
    \tabcolsep=0.2em
    \newcommand{\gain}[1]{\raisebox{-0.4ex}{\fontsize{5.5}{6}\selectfont\textcolor{pos}{#1}}}
    \begin{tabular*}{\linewidth}{@{\extracolsep{\fill}}l*{15}{c}@{}}
        \toprule
        & \multicolumn{3}{c}{\textbf{Overall}} & \multicolumn{3}{c}{\textbf{ADT}~\citep{adt}} & \multicolumn{3}{c}{\textbf{DexYCB}~\citep{dexycb}} & \multicolumn{3}{c}{\textbf{PStudio}~\citep{panoptic}} & \multicolumn{3}{c}{\textbf{DROID}~\citep{droid}} \\
        \cmidrule(lr){2-4}\cmidrule(lr){5-7}\cmidrule(lr){8-10}\cmidrule(lr){11-13}\cmidrule(lr){14-16}
        \textbf{Method} & AJ~$\uparrow$ & $\mathrm{APD}_{\mathrm{3D}}$~$\uparrow$ & OA~$\uparrow$ & AJ~$\uparrow$ & $\mathrm{APD}_{\mathrm{3D}}$~$\uparrow$ & OA~$\uparrow$ & AJ~$\uparrow$ & $\mathrm{APD}_{\mathrm{3D}}$~$\uparrow$ & OA~$\uparrow$ & AJ~$\uparrow$ & $\mathrm{APD}_{\mathrm{3D}}$~$\uparrow$ & OA~$\uparrow$ & AJ~$\uparrow$ & $\mathrm{APD}_{\mathrm{3D}}$~$\uparrow$ & OA~$\uparrow$ \\
        \midrule
        D4RT$^\dagger$~\citep{d4rt} & 39.74 & 54.26 & 84.45 & 18.28 & 27.80 & 90.23 & 63.80 & 77.86 & 89.46 & 45.62 & 57.77 & 88.63 & 31.27 & 53.62 & 69.49 \\
        \quad + \thedataset & \textbf{41.82}\gain{+2.08} & \textbf{56.55}\gain{+2.29} & \textbf{85.76}\gain{+1.31} & 21.16 & 31.04 & 90.67 & 63.21 & 76.95 & 89.83 & 49.28 & 61.94 & 90.71 & 33.64 & 56.28 & 71.83 \\
        \bottomrule
    \end{tabular*}
    \caption{\textbf{Training D4RT with \thedataset.}
    Adding \thedataset improves average 3D localization and visibility.
    Overall is the unweighted mean across the four datasets, with improvements in green. $^\dagger$ denotes our D4RT reproduction.}
    \label{tab:d4rt-data}
\end{table*}

\noindent\textbf{Annotations approach human agreement.}
Expert vision models fail to re-identify reappearing points, so we evaluate this step directly.
Two annotators independently label 582 point--frame pairs after reappearance, 428 on people and moved objects (supplementary Sec.~\ref{sec:supp-egotap4d}).
\themodel with GPT-6 Astra reaches 76.9 APD, close to the annotators' 79.3 agreement, whereas the best expert model reaches 40.1 (Tab.~\ref{tab:reappearance-human-agreement}).
The open-weight Inkling 2 Small also beats every expert, so re-identification needs no closed-weight model.
Fig.~\ref{fig:egotap4d-reconnection} shows two returns after 76 and 25\,s, where \themodel predicts 10 and 9 of the 10 points visible and AllTracker only 0 and 1.

\begin{table}[t]
    \centering
    \fontsize{7.0}{8.4}\selectfont
    \setlength{\tabcolsep}{0.8pt}
    \renewcommand{\arraystretch}{1.0}
    \newcommand{\groupband}[1]{\rowcolor{tabband}\multicolumn{10}{l}{\emph{#1}}}
    \begin{tabular*}{\linewidth}{@{\extracolsep{\fill}}lccccccccc@{}}
        \toprule
        & \multicolumn{3}{c}{All} & \multicolumn{3}{c}{Dynamic} & \multicolumn{3}{c}{Static} \\
        \cmidrule(lr){2-4}\cmidrule(lr){5-7}\cmidrule(l){8-10}
        Model & AJ$\uparrow$ & APD$\uparrow$ & OA$\uparrow$ & AJ$\uparrow$ & APD$\uparrow$ & OA$\uparrow$ & AJ$\uparrow$ & APD$\uparrow$ & OA$\uparrow$ \\
        \midrule
        human & 68.7 & 79.3 & 92.4 & 66.1 & 77.8 & 91.1 & 75.3 & 82.7 & 96.1 \\
        \groupband{Expert vision models} \\
        RoMa v2~\citep{romav2} & 24.2 & 37.4 & \thirdbest{73.2} & 11.3 & 19.0 & \thirdbest{65.0} & 65.6 & \secondbest{79.3} & \secondbest{96.1} \\
        CoTracker3~\citep{cotracker3} & 30.1 & 33.1 & 55.3 & 14.7 & 15.6 & 42.1 & 62.2 & 72.9 & 92.2 \\
        BootsTAPIR~\citep{bootstap} & \thirdbest{36.4} & 39.4 & 58.2 & \thirdbest{21.9} & \thirdbest{23.6} & 48.6 & \thirdbest{68.9} & 75.4 & 85.1 \\
        TAPNext++~\citep{tapnextpp} & 13.8 & 14.1 & 33.7 & 0.7 & 0.7 & 26.2 & 42.3 & 44.9 & 54.5 \\
        AllTracker & 34.8 & \thirdbest{40.1} & 60.1 & 20.0 & 23.2 & 48.6 & \secondbest{69.6} & \thirdbest{78.6} & 92.2 \\
        AllTracker, HR & 26.7 & 28.4 & 47.4 & 15.9 & 16.7 & 40.9 & 50.8 & 54.8 & 65.6 \\
        \groupband{VLMs and agents} \\
        \themodel{} (Inkling 2 Small) & \secondbest{42.5} & \secondbest{59.7} & \secondbest{86.0} & \secondbest{35.4} & \secondbest{51.9} & \secondbest{83.4} & 63.2 & 77.5 & \thirdbest{93.2} \\
        \themodel{} (GPT-6 Astra) & \best{64.4} & \best{76.9} & \best{92.6} & \best{60.5} & \best{73.6} & \best{90.7} & \best{74.3} & \best{84.7} & \best{98.1} \\
        \bottomrule
    \end{tabular*}
    \caption{\textbf{Human evaluation of reappearing points.} Full-video \themodel inspects only the endpoint images, with full-video AllTracker proposals. With GPT-6 Astra, \themodel approaches human agreement. Even with the open-weight Inkling 2 Small, it beats every expert vision model on all and dynamic points.}
    \label{tab:reappearance-human-agreement}
\end{table}

\noindent\textbf{Downstream validation.}
As long-horizon 4D trackers are not yet available, we check label quality by training D4RT~\citep{d4rt} with and without \thedataset (supplementary Sec.~\ref{sec:supp-d4rt}).
Both models use the D4RT architecture and train for 200k steps.
The baseline uses eight public datasets, the second model adds \thedataset, and all datasets are sampled with equal weights.
We evaluate on ADT and PStudio from TAPVid-3D~\citep{tapvid3d}, DexYCB~\citep{dexycb}, and DROID~\citep{droid} with the TAPVid-3D metrics.
Adding \thedataset improves average 3D localization, AJ, and visibility on most benchmarks (Tab.~\ref{tab:d4rt-data}). Localization drops slightly only on DexYCB, whose stationary tabletop views differ most from our egocentric videos.
\endgroup

\titlespacing*{\section}{0pt}{8pt plus 2pt minus 2pt}{8pt}
\section{Conclusion}
\label{sec:conclusion}

Finding the same physical point across views and time is a longstanding problem in computer vision.
We approached it as visual reasoning: with \themodel, a general-purpose VLM forms a hypothesis about the point, inspects the images to test it, and revises its answer.
With GPT-6 Astra, \themodel surpasses expert vision models on \thebench, especially on reappearing points.
This shows that VLM reasoning has advanced from objects to individual pixels.
We used \themodel to reconnect reappearing points in long egocentric videos for \thedataset.
Applied at scale, it can provide 3D training data for expert vision models and for robots that learn from human demonstrations.
We hope the community will build on agentic tracking to develop models with stronger spatial intelligence.

\endgroup
\flushbottom

\section*{Acknowledgments}

The authors acknowledge the computational resources from the University of Virginia Research Computing and Data Analytics Center, the Advanced Cyberinfrastructure Coordination Ecosystem: Services \& Support (ACCESS) program, and the National Artificial Intelligence Research Resource (NAIRR) Pilot, including the Anvil supercomputer at Purdue University (National Science Foundation award OAC 2005632) and the Delta and DeltaAI advanced computing resources (National Science Foundation awards OAC 2005572 and OAC 2320345).
This publication was made using the Waymo Open Dataset, provided by Waymo LLC under license terms available at \url{https://waymo.com/open}.
Frames from Perception Test~\citep{perceptiontest} and RoboTAP~\citep{robotap} are used under the Creative Commons Attribution 4.0 license.
The authors thank Yihan Wang, Zhao Dong, and Jakob Engel for proofreading.

{
    \small
    \def\UrlBreaks{\do\/\do-\do.\do_}
    \bibliographystyle{cvpr2027/ieeenat_fullname}
    \bibliography{paper}
}

\appendix
\clearpage
\pagenumbering{arabic}
\renewcommand{\thepage}{S\arabic{page}}
\pdfbookmark[0]{Supplementary Material}{supplementary-contents}
\twocolumn[{%
\begingroup
\definecolor{suppheading}{RGB}{24,48,76}
\setlength{\parindent}{0pt}
\setlength{\parskip}{0pt}
\hypersetup{linkcolor=suppheading}
{\fontsize{18}{22}\selectfont\sffamily\bfseries\color{suppheading}
Supplementary Material\par}
\vspace{4pt}
{\large\papertitle\par}
\vspace{12pt}
This supplement provides implementation details, evaluation protocols, and additional evidence for the main paper.
Section~\ref{sec:supp-trackingbench} describes how \thebench is built and evaluated.
Section~\ref{sec:supp-agentap} documents the tools, inference settings, and computational costs of \themodel.
Section~\ref{sec:supp-results} provides additional results, including agent ablations and qualitative examples.
Section~\ref{sec:supp-egotap4d} details \thedataset, including geometry recovery, trajectory annotation, human evaluation, and the D4RT training comparison.
\par\vspace{18pt}
\input{sections/supplement_contents}
\endgroup
}]
\setcounter{table}{0}\renewcommand{\thetable}{S\arabic{table}}\renewcommand{\theHtable}{supp.\arabic{table}}%

\raggedbottom
\setcounter{topnumber}{4}
\setcounter{bottomnumber}{4}
\setcounter{totalnumber}{8}
\setcounter{dbltopnumber}{4}
\renewcommand{\topfraction}{0.95}
\renewcommand{\dbltopfraction}{0.95}
\renewcommand{\textfraction}{0.05}
\renewcommand{\floatpagefraction}{0.65}
\renewcommand{\dblfloatpagefraction}{0.65}
\makeatletter
\let\supp@xfloat\@xfloat
\def\@xfloat#1[#2]{\supp@xfloat#1[!tbp]}
\def\@xdblfloat#1[#2]{%
  \ifx\@xfloat\supp@xfloat\expandafter\@firstoftwo\else\expandafter\@secondoftwo\fi
  {\supp@xfloat{#1}[#2]}{\supp@xfloat{#1}[!tp]}%
  \hsize\textwidth\linewidth\textwidth}
\setlength{\@fpsep}{12pt plus 2pt minus 2pt}
\setlength{\@dblfpsep}{12pt plus 2pt minus 2pt}
\setlength{\@fptop}{0pt}
\setlength{\@dblfptop}{0pt}
\makeatother

\setcounter{figure}{0}
\renewcommand{\thefigure}{S\arabic{figure}}
\renewcommand{\thetable}{S\arabic{table}}
\renewcommand{\theHfigure}{supp.\arabic{figure}}
\renewcommand{\theHtable}{supp.\arabic{table}}

\clearpage
\section{\thebench}
\label{sec:supp-trackingbench}

\thebench is built in three steps. First, we select video clips from public point-tracking benchmarks (Sec.~\ref{sec:supp-video-selection}). Second, we build a pool of 14,896 queries from their released tracks (Sec.~\ref{sec:supp-query-pool}). Third, we sample the 2,000-query main benchmark from the 12,496 real-video queries of this pool (Sec.~\ref{sec:supp-benchmark-composition}). Because public videos may appear in VLM training data, we also annotate 1,814 held-out queries on 15 new videos (Sec.~\ref{sec:supp-heldout-annotation}). Sec.~\ref{sec:supp-evaluation-protocol} details the evaluation protocol.

\subsection{Source Videos}
\label{sec:supp-video-selection}

We draw 7,172 candidate clips from public point-tracking benchmarks (Tab.~\ref{tab:benchmark-sources}), allocate a quota to each source, and select representative clips within each source.

\paragraph{Source quotas.}
We represent each clip by the mean DINOv3 ViT-L~\citep{dinov3} features of 16 frames, normalized to unit length. A greedy covering procedure~\citep{gonzalez1985} counts how many representative clips are needed to cover each source within a distance of 0.7. Each quota grows with the square root of this count, so visually diverse sources receive more clips while large sources do not dominate. Every source contributes at least eight clips when available, and DriveTrack and ADT together contribute at most 50.

\paragraph{Clip selection.}
Within each source, we cluster the appearance features together with five tracking attributes: camera rotation, visible fraction, points per labeled frame, clip length, and dynamic fraction. The attributes add variation in tracking conditions that appearance alone misses. We log-transform points per labeled frame and clip length, standardize all five attributes, scale them by 0.5, and concatenate them with the appearance features. With a fixed seed, we then select the unused clip nearest to each cluster center (Fig.~\ref{fig:benchmark-selection}).

\paragraph{Final clips.}
We select 300 clips, remove the 8 STIR clips because their landmarks lack verified temporal identities~\citep{stir}, and add 20 TAP-Vid Kubric validation clips for synthetic analysis~\citep{tapvid,kubric}, giving 312 clips. The 28 Kubric and RGB-Stacking clips are synthetic and stay out of the main benchmark.

{\makeatletter\let\@xfloat\supp@xfloat\makeatother
\begin{table}[t]
    \centering
    \fontsize{7.0}{8.4}\selectfont
    \tabcolsep=0.5em
    \begin{tabular*}{\linewidth}{@{\extracolsep{\fill}}lccc@{}}
    \toprule
    \textbf{Source} & Candidates & Selected & Evaluated \\
    \midrule
    TAP-Vid Kinetics~\citep{tapvid} & 1,144 & 75 & 75 \\
    EgoPoints~\citep{egopoints} & 517 & 39 & 39 \\
    DriveTrack~\citep{drivetrack,tapvid3d} & 2,457 & 33 & 33 \\
    RoboTrack Real~\citep{robotrack} & 318 & 27 & 27 \\
    TAP-Vid Kubric~\citep{tapvid,kubric} & \textemdash & 0 & 20 \\
    Perception Test~\citep{perceptiontest} & 73 & 18 & 18 \\
    TAPVid-3D ADT~\citep{tapvid3d} & 1,956 & 17 & 17 \\
    ITTO~\citep{itto} & 75 & 16 & 16 \\
    VOTSp2026~\citep{votsp2026} & 50 & 15 & 15 \\
    TAP-Vid DAVIS~\citep{tapvid} & 30 & 13 & 13 \\
    RoboTAP~\citep{robotap} & 265 & 10 & 10 \\
    STIR~\citep{stir} & 64 & 8 & 0 \\
    TAPVid-3D PStudio~\citep{tapvid3d} & 156 & 8 & 8 \\
    TAP-Vid RGB-Stacking~\citep{tapvid} & 50 & 8 & 8 \\
    VL-SurgPT~\citep{vlsurgpt} & 12 & 8 & 8 \\
    DrivTAP~\citep{tapformer} & 5 & 5 & 5 \\
    \midrule
    Total & 7,172 & 300 & 312 \\
    \bottomrule
    \end{tabular*}
    \caption{\textbf{Candidate sources and selected clips.} Candidates: clips considered for selection (Fig.~\ref{fig:benchmark-selection}). Selected: clips chosen by source quota and clustering. Evaluated: clips of the 14,896-query pool, after removing STIR and adding 20 Kubric clips, which were not candidates.}
    \label{tab:benchmark-sources}
\end{table}
}
{\makeatletter\let\@xfloat\supp@xfloat\makeatother
\begin{figure}[t]
    \centering
    \includegraphics[width=\linewidth]{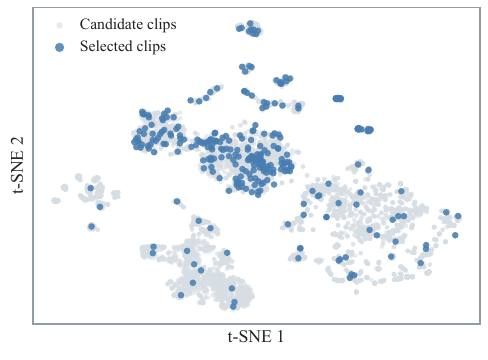}
    \caption{\textbf{Appearance-based video selection.} Selected clips (blue) and candidate clips (gray), projected from DINOv3 embeddings~\citep{dinov3} using t-SNE~\citep{tsne}. The projection only illustrates appearance neighborhoods. Selection uses the full-dimensional features and tracking attributes.}
    \label{fig:benchmark-selection}
\end{figure}
}
\begin{figure*}[t]
    \centering
    \includegraphics[width=\textwidth]{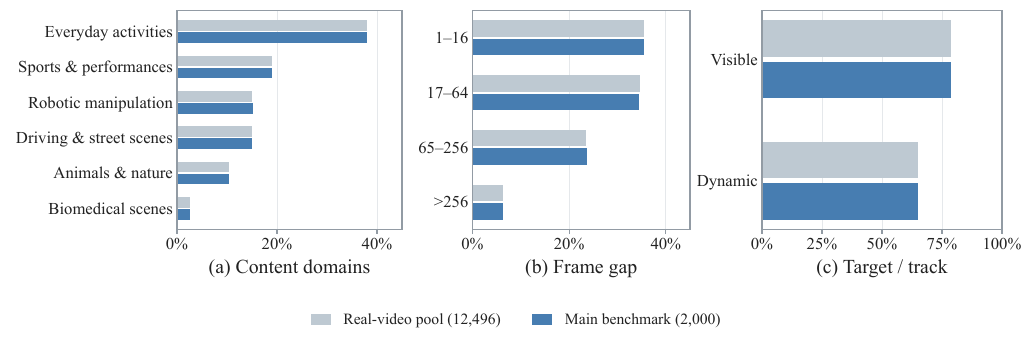}
    \caption{\textbf{Main benchmark and real-video pool.} Query proportions in the 2,000-query benchmark and among the 12,496 real-video queries of the pool. Stratified sampling preserves the domain, frame-gap, visibility, and motion composition.}
    \label{fig:benchmark-frame-gaps}
\end{figure*}

\subsection{Query Pool}
\label{sec:supp-query-pool}
\label{sec:supp-query-annotation}
\label{sec:supp-released-track-queries}

Each query pairs a reference observation of a released track with a target frame, as described in Sec.~\ref{sec:benchmark-curation}. Query points thus inherit their identities from the released tracks, as in TAP-Vid~\citep{tapvid}. To avoid concentrating on nearby frames and nearly static points, we sample reference--target pairs across strata of displacement, frame gap, target visibility, and intermediate invisibility. The smallest-displacement stratum receives half weight. At most four targets per reference and 40 track--reference pairs per clip limit repeated queries. This yields 14,896 queries from 5,958 tracks, 12,496 of them from real videos and 2,400 from synthetic ones.

\subsection{Main Benchmark}
\label{sec:supp-benchmark-composition}
\label{sec:supp-benchmark-validation}

To reduce the computational cost of evaluation, all main comparisons use 2,000 queries sampled from the 12,496 real-video queries. We fix the sampling rule before evaluation, then check that the sample represents the pool and examine its labels.

\paragraph{Sampling.}
We stratify by content domain, target visibility, whole-track motion (static or dynamic, defined below), and frame gap (1--16, 17--64, 65--256, and more than 256 frames). Each of the 85 nonempty strata receives a share of the 2,000 queries proportional to its size in the pool. We sample uniformly within each stratum with seed 0. The rule and the seed were fixed before any score on the benchmark was observed.

\paragraph{Composition.}
The benchmark covers 272 clips from 13 real-video sources and six content domains: everyday activities, sports and performances, robotic manipulation, driving and street scenes, animals and nature, and biomedical scenes. Domains describe scene content, so one source can span several domains. The benchmark has 1,579 visible and 421 invisible targets, and 1,296 queries on dynamic tracks and 704 on static ones. A track is dynamic if any of its visible annotations lies on an object that GMOS~\citep{gmos} identifies as moving, and we review these labels manually. The label thus describes the whole track rather than the queried interval. Frame gaps count frames rather than seconds, since frame rates differ across sources.

\paragraph{Representativeness.}
The benchmark matches the pool in domain, frame gap, visibility, and motion (Fig.~\ref{fig:benchmark-frame-gaps}). Scores match as well. Across the 29 systems scored on both sets, APD rankings on the benchmark and the pool have Kendall $\tau=0.995$, and APD differs by at most 1.1 points (Fig.~\ref{fig:pool-validation}). Models run only on the benchmark, including all agents, GPT-6.1 Sol, Claude Fable 5.2, and Inkling 2 Small, have no pool scores and are not included. Over 200 alternative sampling seeds, the 95\% range of a system's APD stays within $\pm1.3$ points of its pool score for the median system and within $\pm2.3$ points for every system.

{\makeatletter\let\@xfloat\supp@xfloat\makeatother
\begin{figure}[t]
    \centering
    \includegraphics{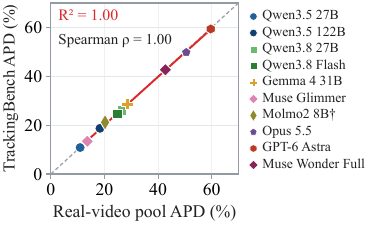}
    \caption{\textbf{Benchmark validation against the larger real-video pool.} APD on the 12,496-query pool (x) and the 2,000-query benchmark (y) for the VLMs of Fig.~\ref{fig:analysis}b with pool scores. Red: linear fit with 95\% CI. Dashed: equal scores. $\dagger$: video input. All 29 systems scored on both sets differ by at most 1.1 APD.}
    \label{fig:pool-validation}
\end{figure}
}
{\makeatletter\let\@xfloat\supp@xfloat\makeatother
\begin{figure*}[!t]
    \centering
    \includegraphics[width=\textwidth]{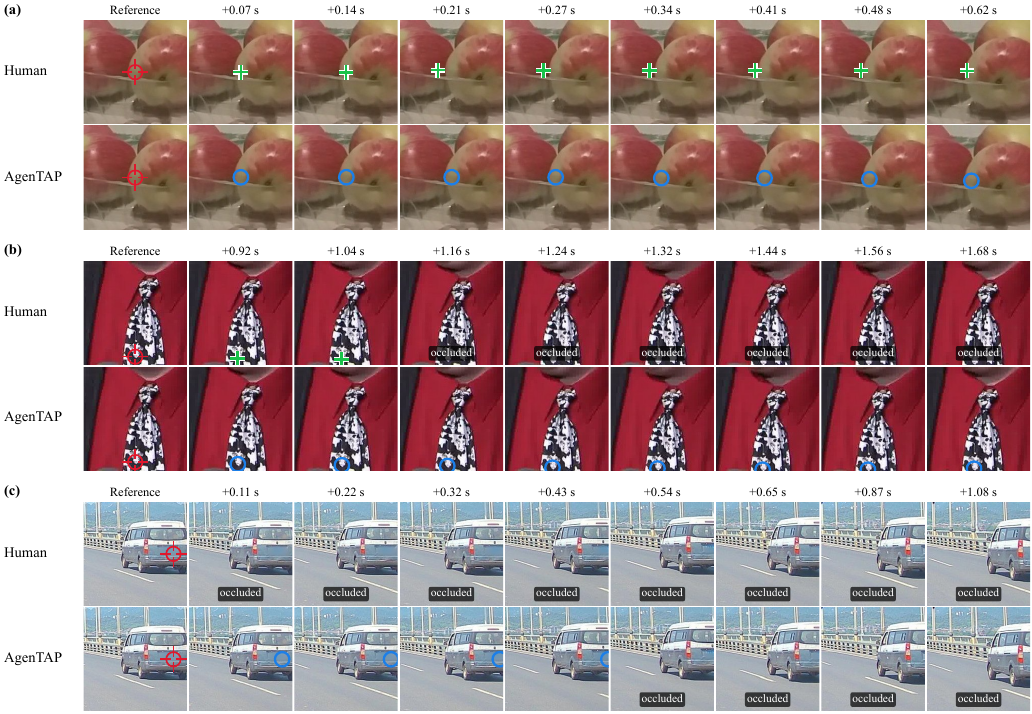}
    \caption{\textbf{Human label failures.} Released annotations (upper rows, green crosses) and predictions of \themodel with GPT-6 Astra (lower rows, blue circles) on consecutive frames. The first column shows the reference frame and query point. (a)~The label moves to a neighboring apple, while \themodel stays on the queried one. (b)~A visible point is labeled occluded, while \themodel continues to locate it. (c)~The point beside the license plate is labeled occluded while still visible, and \themodel follows it to the image border and marks it invisible once it leaves the frame.}
    \label{fig:human-failure-sequences}
\end{figure*}
}
{\makeatletter\let\@xfloat\supp@xfloat\makeatother
\begin{table*}[!t]
\centering
\fontsize{7.0}{8.4}\selectfont
\setlength{\tabcolsep}{0.7pt}
\renewcommand{\arraystretch}{1.0}
\newcommand{\domaindelta}[2]{\raisebox{-0.4ex}{\fontsize{4.8}{5.2}\selectfont\textcolor{#1}{#2}}}
\begin{tabular*}{\textwidth}{@{\extracolsep{\fill}}ll*{22}{c}@{}}
\toprule
& & \multicolumn{4}{c}{\textbf{Overall}} & \multicolumn{3}{c}{\shortstack{\textbf{Everyday}\\\textbf{Activities}}} & \multicolumn{3}{c}{\shortstack{\textbf{Robotic}\\\textbf{Manipulation}}} & \multicolumn{3}{c}{\shortstack{\textbf{Sports and}\\\textbf{Performances}}} & \multicolumn{3}{c}{\shortstack{\textbf{Driving and}\\\textbf{Street Scenes}}} & \multicolumn{3}{c}{\shortstack{\textbf{Animals}\\\textbf{and Nature}}} & \multicolumn{3}{c}{\shortstack{\textbf{Biomedical}\\\textbf{Scenes}}} \\
\cmidrule(lr){3-6}\cmidrule(lr){7-9}\cmidrule(lr){10-12}\cmidrule(lr){13-15}\cmidrule(lr){16-18}\cmidrule(lr){19-21}\cmidrule(lr){22-24}
\textbf{Model} & \textbf{Harness} & AJ & APD & OA & EPE$\downarrow$ & AJ & APD & OA & AJ & APD & OA & AJ & APD & OA & AJ & APD & OA & AJ & APD & OA & AJ & APD & OA \\
\midrule
\rowcolor{black!7}\multicolumn{24}{l}{\textbf{Two images}} \\
\rowcolor{tabband}\multicolumn{24}{l}{\emph{Expert vision models}} \\
RoMa v2~\citep{romav2} &  & 20.5 & 33.7 & 77.6 & 16.0 & 22.0 & 34.0 & 83.3 & 29.3 & 46.8 & 75.5 & 16.9 & 30.1 & 75.6 & 19.4 & 31.4 & 68.7 & 13.6 & 25.4 & 72.6 & 16.6 & 27.7 & \thirdbest{94.0} \\
AllTracker~\citep{alltracker} &  & 27.2 & 37.3 & 73.3 & 13.7 & 27.2 & 36.6 & 77.3 & 36.2 & 48.4 & 72.2 & 30.1 & 40.1 & 73.2 & 21.7 & 31.2 & 66.3 & 20.6 & 30.1 & 69.2 & 21.8 & 29.5 & 78.0 \\
AllTracker, HR &  & 30.3 & 39.1 & 70.8 & 14.0 & 29.5 & 37.4 & 71.3 & 39.4 & 49.2 & 71.9 & \thirdbest{32.8} & 41.4 & 73.0 & 26.5 & 37.0 & 67.0 & \secondbest{24.7} & 32.7 & 67.3 & \thirdbest{26.7} & 34.1 & 76.0 \\
\rowcolor{tabband}\multicolumn{24}{l}{\emph{VLMs and agents}} \\
\multirow{2}{*}{Qwen3.8-27B~\citep{qwen3.8-27b}} & Direct & 4.0 & 8.1 & 80.1 & 24.2 & 3.6 & 7.0 & 87.6 & 6.5 & 13.2 & 73.5 & 4.5 & 9.2 & 77.2 & 2.8 & 5.9 & 72.0 & 2.9 & 6.2 & 76.4 & 5.7 & 10.5 & 90.0 \\
 & \textbf{\themodel} & 21.4\domaindelta{pos}{+17.4} & 35.7\domaindelta{pos}{+27.6} & 79.6\domaindelta{black!45}{$-$0.5} & 13.9\domaindelta{pos}{$-$10.4} & 21.0 & 33.6 & 85.8 & 29.6 & 48.2 & 73.8 & 22.1 & 37.4 & 75.3 & 18.8 & 32.9 & 77.3 & 15.6 & 28.7 & 73.6 & 21.9 & 32.7 & 92.0 \\
\multirow{2}{*}{Gemma-4-31B-it~\citep{gemma4-31b-it}} & Direct & 5.1 & 9.7 & 76.6 & 23.1 & 4.2 & 7.5 & 80.4 & 8.6 & 16.0 & 71.2 & 7.6 & 14.5 & 75.9 & 2.7 & 5.6 & 71.7 & 3.3 & 6.7 & 75.0 & 5.9 & 10.9 & 92.0 \\
 & \textbf{\themodel} & 18.3\domaindelta{pos}{+13.2} & 31.8\domaindelta{pos}{+22.1} & 79.0\domaindelta{pos}{+2.4} & 15.9\domaindelta{pos}{$-$7.2} & 16.7 & 28.3 & 84.2 & 27.3 & 46.8 & 74.2 & 19.3 & 33.7 & 76.6 & 17.2 & 30.8 & 73.7 & 13.7 & 25.6 & 76.0 & 14.8 & 25.0 & 90.0 \\
\multirow{2}{*}{Inkling 2 Small~\citep{inkling2-small}} & Direct & 7.1 & 13.2 & 79.1 & 22.0 & 5.6 & 10.0 & 84.8 & 12.3 & 22.4 & 74.8 & 6.8 & 13.1 & 76.4 & 8.2 & 15.8 & 74.7 & 5.8 & 11.1 & 74.0 & 5.4 & 9.5 & 86.0 \\
 & \textbf{\themodel} & 24.6\domaindelta{pos}{+17.5} & 40.4\domaindelta{pos}{+27.2} & 81.5\domaindelta{pos}{+2.4} & 11.6\domaindelta{pos}{$-$10.4} & 25.2 & 39.1 & 87.9 & 32.3 & 52.4 & 77.2 & 25.0 & 42.2 & 78.2 & 21.1 & 37.1 & 74.7 & 18.0 & 32.3 & 76.9 & 21.0 & 32.7 & \thirdbest{94.0} \\
\multirow{2}{*}{Muse Wonder Full~\citep{muse-wonder-full}} & Direct & 9.6 & 16.4 & 83.0 & 17.9 & 9.8 & 15.9 & 87.6 & 12.8 & 21.6 & 78.5 & 10.2 & 17.2 & \secondbest{83.2} & 8.9 & 16.0 & 75.7 & 6.1 & 11.9 & \thirdbest{80.3} & 6.6 & 10.9 & \thirdbest{94.0} \\
 & \textbf{\themodel} & 29.3\domaindelta{pos}{+19.7} & 44.6\domaindelta{pos}{+28.2} & 85.4\domaindelta{pos}{+2.4} & 10.0\domaindelta{pos}{$-$7.9} & 28.9 & 42.6 & 90.5 & 39.1 & 58.0 & \secondbest{82.1} & 30.8 & 45.8 & \secondbest{83.2} & 26.9 & 43.7 & \thirdbest{79.7} & 21.4 & 35.8 & \secondbest{80.8} & 25.8 & \thirdbest{37.3} & \secondbest{96.0} \\
\multirow{2}{*}{Claude Opus 5.5~\citep{claude-opus5.5}} & Direct & 14.0 & 24.1 & 82.2 & 15.6 & 13.5 & 22.5 & 89.2 & 17.9 & 31.2 & 74.5 & 14.5 & 24.9 & 79.5 & 14.3 & 25.6 & 75.3 & 9.6 & 18.2 & 79.3 & 15.0 & 22.7 & \secondbest{96.0} \\
 & \textbf{\themodel} & 27.7\domaindelta{pos}{+13.7} & 43.1\domaindelta{pos}{+18.9} & 84.2\domaindelta{pos}{+2.0} & 10.3\domaindelta{pos}{$-$5.4} & 27.0 & 40.7 & 90.9 & 37.4 & 57.3 & 78.1 & 28.4 & 44.0 & \thirdbest{81.9} & 25.7 & 41.8 & 77.7 & 20.0 & 34.7 & 78.8 & \secondbest{28.0} & \secondbest{39.1} & \best{98.0} \\
\multirow{2}{*}{Claude Fable 5.2~\citep{claude-fable5.2}} & Direct & 15.7 & 26.4 & 83.3 & 14.4 & 14.3 & 23.3 & 89.6 & 21.8 & 36.4 & 77.5 & 16.6 & 28.0 & 80.6 & 15.6 & 27.6 & 78.7 & 11.4 & 20.8 & 76.9 & 17.1 & 26.4 & \secondbest{96.0} \\
 & \textbf{\themodel} & \thirdbest{30.7\domaindelta{pos}{+15.0}} & \secondbest{46.8\domaindelta{pos}{+20.3}} & \thirdbest{86.0\domaindelta{pos}{+2.7}} & \best{9.1\domaindelta{pos}{$-$5.3}} & \thirdbest{29.6} & \thirdbest{43.5} & \best{92.9} & \secondbest{41.8} & \best{62.7} & 79.5 & \thirdbest{32.8} & \best{49.6} & \best{84.8} & \thirdbest{28.0} & \secondbest{45.2} & \secondbest{80.0} & 21.8 & \thirdbest{36.7} & 78.4 & \best{30.1} & \best{41.4} & \secondbest{96.0} \\
\multirow{2}{*}{GPT-6.1 Sol~\citep{gpt6.1-sol}} & Direct & 19.6 & 31.1 & 83.7 & 13.3 & 18.3 & 28.1 & 88.9 & 30.5 & 47.0 & 78.8 & 19.6 & 31.1 & \thirdbest{81.9} & 19.7 & 33.0 & 78.0 & 12.5 & 21.6 & \thirdbest{80.3} & 14.9 & 22.3 & \thirdbest{94.0} \\
 & \textbf{\themodel} & \secondbest{31.3\domaindelta{pos}{+11.7}} & \thirdbest{46.4\domaindelta{pos}{+15.3}} & \secondbest{86.1\domaindelta{pos}{+2.4}} & \thirdbest{9.4\domaindelta{pos}{$-$3.9}} & \secondbest{30.6} & \best{44.4} & \secondbest{91.7} & \thirdbest{40.6} & \thirdbest{59.7} & \thirdbest{81.5} & \secondbest{33.7} & \thirdbest{48.5} & \best{84.8} & \secondbest{28.8} & \thirdbest{44.8} & \best{80.7} & \thirdbest{23.1} & \secondbest{37.5} & 79.8 & 26.2 & 36.8 & \secondbest{96.0} \\
\multirow{2}{*}{GPT-6 Astra~\citep{gpt6-astra}} & Direct & 21.8 & 33.7 & 85.5 & 12.5 & 20.3 & 30.3 & 89.7 & 31.4 & 48.0 & 80.1 & 23.0 & 34.6 & \best{84.8} & 22.4 & 37.1 & \thirdbest{79.7} & 14.4 & 24.4 & \best{85.1} & 16.7 & 24.5 & \thirdbest{94.0} \\
 & \textbf{\themodel} & \best{32.1\domaindelta{pos}{+10.3}} & \best{46.9\domaindelta{pos}{+13.2}} & \best{86.3\domaindelta{pos}{+0.8}} & \secondbest{9.2\domaindelta{pos}{$-$3.3}} & \best{30.9} & \secondbest{44.3} & \thirdbest{91.2} & \best{43.0} & \secondbest{61.2} & \best{83.8} & \best{34.5} & \secondbest{48.7} & \best{84.8} & \best{29.0} & \best{45.4} & \best{80.7} & \best{25.4} & \best{39.9} & \secondbest{80.8} & 26.5 & 36.4 & \thirdbest{94.0} \\
\midrule
\rowcolor{black!7}\multicolumn{24}{l}{\textbf{Full video}} \\
\rowcolor{tabband}\multicolumn{24}{l}{\emph{Expert vision models}} \\
CoTracker3~\citep{cotracker3} &  & 28.7 & 41.4 & 81.3 & 11.5 & 27.3 & 38.5 & 84.3 & 39.5 & 55.9 & 80.1 & 30.1 & 42.1 & 80.3 & 25.8 & 41.3 & 80.0 & 24.2 & 34.9 & 73.1 & 23.5 & 33.2 & 92.0 \\
BootsTAPIR~\citep{bootstap} &  & 26.3 & 36.6 & 76.3 & 12.8 & 24.6 & 33.8 & 78.8 & 36.6 & 48.6 & 76.2 & 29.9 & 40.7 & 78.5 & 22.4 & 32.9 & 68.0 & 20.4 & 29.7 & 72.1 & 25.1 & 34.5 & 90.0 \\
TAPNext++~\citep{tapnextpp} &  & 26.9 & 40.1 & 83.6 & 12.0 & 27.8 & 39.7 & 87.1 & 35.7 & 53.4 & \thirdbest{83.4} & 33.5 & 47.0 & 84.3 & 18.5 & 30.2 & 75.0 & 15.9 & 26.8 & 78.4 & 21.5 & 32.3 & \best{100.0} \\
AllTracker~\citep{alltracker} &  & 30.6 & 43.2 & 81.5 & 10.6 & 29.5 & 40.7 & 85.0 & 43.5 & 60.0 & 82.8 & 33.3 & 46.3 & 81.9 & 25.3 & 37.9 & 76.0 & 23.5 & 34.7 & 73.6 & 24.5 & 32.3 & 82.0 \\
AllTracker, HR &  & \secondbest{34.1} & 46.0 & 80.2 & 10.4 & \secondbest{32.8} & 43.8 & 83.5 & \thirdbest{45.0} & 58.9 & 79.5 & \best{38.0} & 50.0 & 81.9 & \secondbest{28.7} & 41.9 & 75.7 & \secondbest{28.1} & 37.8 & 72.1 & 29.1 & 37.3 & 82.0 \\
\rowcolor{tabband}\multicolumn{24}{l}{\emph{VLMs and agents}} \\
Inkling 2 Small~\citep{inkling2-small} & \textbf{\themodel} & 28.1\domaindelta{pos}{+21.0} & 44.6\domaindelta{pos}{+31.5} & 83.2\domaindelta{pos}{+4.1} & 10.5\domaindelta{pos}{$-$11.5} & 27.7 & 42.4 & 89.3 & 38.1 & 59.5 & 78.5 & 29.7 & 46.7 & 80.3 & 24.8 & 41.8 & 75.7 & 20.6 & 35.9 & \thirdbest{80.3} & 23.5 & 35.5 & \thirdbest{96.0} \\
Muse Wonder Full~\citep{muse-wonder-full} & \textbf{\themodel} & 32.7\domaindelta{pos}{+23.1} & \thirdbest{48.3\domaindelta{pos}{+31.9}} & \thirdbest{86.6\domaindelta{pos}{+3.6}} & \thirdbest{8.9\domaindelta{pos}{$-$9.0}} & 32.0 & \thirdbest{45.9} & 90.8 & 43.7 & 62.5 & \secondbest{84.8} & 35.3 & \thirdbest{51.2} & \secondbest{85.3} & 26.9 & 43.7 & \thirdbest{80.7} & \thirdbest{27.0} & \best{42.3} & \secondbest{81.7} & \secondbest{29.7} & \secondbest{40.9} & \secondbest{98.0} \\
Claude Opus 5.5~\citep{claude-opus5.5} & \textbf{\themodel} & 31.5\domaindelta{pos}{+17.5} & 47.9\domaindelta{pos}{+23.7} & 84.3\domaindelta{pos}{+2.1} & 9.0\domaindelta{pos}{$-$6.6} & 30.6 & 45.0 & \thirdbest{91.6} & 43.0 & \thirdbest{64.1} & 78.1 & 33.6 & 50.2 & 82.2 & \thirdbest{27.1} & \thirdbest{44.2} & 77.7 & 24.3 & \thirdbest{40.1} & 76.9 & \thirdbest{29.4} & \best{41.4} & \thirdbest{96.0} \\
Claude Fable 5.2~\citep{claude-fable5.2} & \textbf{\themodel} & \thirdbest{33.5\domaindelta{pos}{+17.8}} & \best{49.7\domaindelta{pos}{+23.3}} & \secondbest{86.7\domaindelta{pos}{+3.5}} & \best{8.2\domaindelta{pos}{$-$6.2}} & \thirdbest{32.4} & \secondbest{46.7} & \best{92.8} & \secondbest{45.6} & \best{65.9} & 82.8 & \thirdbest{36.5} & \best{53.5} & \thirdbest{85.0} & \secondbest{28.7} & \best{45.8} & \secondbest{81.0} & 25.6 & \secondbest{41.3} & 78.8 & \best{29.9} & \best{41.4} & \secondbest{98.0} \\
GPT-6 Astra~\citep{gpt6-astra} & \textbf{\themodel} & \best{34.9\domaindelta{pos}{+13.1}} & \secondbest{49.6\domaindelta{pos}{+15.9}} & \best{88.2\domaindelta{pos}{+2.7}} & \secondbest{8.5\domaindelta{pos}{$-$4.0}} & \best{33.3} & \best{47.2} & \secondbest{92.5} & \best{47.6} & \secondbest{65.2} & \best{86.1} & \secondbest{37.9} & \secondbest{52.3} & \best{86.6} & \best{29.6} & \secondbest{44.9} & \best{82.0} & \best{29.4} & \best{42.3} & \best{85.1} & \thirdbest{29.4} & \thirdbest{40.5} & \thirdbest{96.0} \\
\bottomrule
\end{tabular*}
\caption{\textbf{Results at original image resolution.} The rows of Tab.~\ref{tab:agent-comparison} of the main paper scored in native pixels instead of on the $256\times256$ grid, with the same thresholds of 1--16 pixels and an EPE cap of 32 pixels. Overall columns evaluate all queries together, and domain columns omit EPE. HR: high resolution. Subscripts give each agent's change from the same VLM's two-image direct prediction. \textcolor{pos}{Green} marks an improvement and \textcolor{black!45}{gray} a decline. Shading marks the \colorbox{tabfirst}{best}, \colorbox{tabsecond}{second}, and \colorbox{tabthird}{third} results per column within each input block.}
\label{tab:native-resolution}
\end{table*}
}

\paragraph{Label quality.}
We keep the released point identities and visibility labels, accounting for source-specific media alignment and invalid frames. Label quality varies across sources, and RoboTrack Real marks its annotations as unreviewed~\citep{robotrack}. Three annotators reviewed 81 queries on which the free-form GPT-6.1 Sol agent (Sec.~\ref{sec:supp-agent-ablations}) missed by 16 pixels or more, one annotator per query. They judged 19 released labels wrong and 30 targets impossible to label from the two images.

\paragraph{Examples of label errors.}
\phantomsection\label{sec:supp-human-failures}
A separate review of disagreements between \themodel with GPT-6 Astra and the released annotations reveals cases where the label appears wrong (Fig.~\ref{fig:human-failure-sequences}). In the apple sequence, the label moves to a neighboring apple, while \themodel stays on the queried one. In the other two sequences, the labels mark visible points as occluded, while \themodel continues to locate them. In the vehicle sequence, it follows the point to the image border and marks it invisible once it leaves the frame. We evaluate against the released labels without modification, so these predictions are scored as errors.

\subsection{Held-out Annotations}
\label{sec:supp-heldout-annotation}

To check that results on \thebench do not reflect data leakage, we capture 15 new videos, totaling 33.7 minutes of animals and human activities, and annotate them in the three stages of Sec.~\ref{sec:benchmark-curation}. Four annotators take part, and three of them contribute videos.

\paragraph{(1) Initial annotation.}
The uploader selects a reference frame and a physical surface point, then labels the point in about 20 target frames before and after the reference. A visible point is clicked, and a hidden one is marked occluded or outside the image. This gives 96 tracks and 1,870 targets.

\paragraph{(2) Independent verification.}
A second annotator labels the same targets without seeing the first answers and may reject an unsuitable reference point. One rejection removes 19 targets. The two annotators agree on 89.6\% of the remaining 1,851 targets. Where both mark the point visible, their median distance is 0.42 pixels on the $256\times256$ grid.

\paragraph{(3) Adjudication.}
A third annotator labels the 192 disputed targets without seeing either earlier answer, and the final label comes from the third label and the earlier label it agrees with (Sec.~\ref{sec:benchmark-curation}). We exclude 18 unresolved targets and a second unsuitable reference point with its 19 targets. The final set has 1,814 queries on 94 tracks, with 1,364 visible targets and 631 targets before the reference. The median gap between reference and target is 5.5 seconds, and the 90th percentile is 59.7 seconds. The videos and labels remain hidden. Fig.~\ref{fig:analysis}b of the main paper compares direct prediction on this set with the benchmark.

\subsection{Evaluation Protocol}
\label{sec:supp-evaluation-protocol}

All methods are scored with the protocol of Sec.~\ref{sec:benchmark-protocol}. This section defines its metrics and the settings of the expert vision models.

\paragraph{Errors.}
Each query $q$ has a ground-truth visibility $v_q\in\{0,1\}$ and, if $v_q=1$, a position $p_q$ in the native pixels of a $W_q\times H_q$ target image. A method returns a predicted visibility $\hat{v}_q$ and, if $\hat{v}_q=1$, a position $\hat{p}_q$. A missing or unreadable answer counts as a wrong visibility decision. When $v_q=\hat{v}_q=1$, the error is measured on the $256\times256$ grid of TAP-Vid~\citep{tapvid},
\begin{equation*}
    d_q=\left\lVert \operatorname{diag}\!\left(\tfrac{255}{W_q-1},\tfrac{255}{H_q-1}\right)\left(\hat{p}_q-p_q\right)\right\rVert_2,
\end{equation*}
and otherwise $d_q=\infty$. This scaling does not constrain the image resolution given to a model.

\paragraph{Metrics.}
A query is a hit at threshold $x$ if $v_q=\hat{v}_q=1$ and $d_q<x$. With the thresholds $X=\{1,2,4,8,16\}$,
\begin{equation*}
    \delta^x=\frac{\#\{\text{hits at }x\}}{\#\{q: v_q=1\}},\qquad
    \mathrm{APD}=\frac{1}{|X|}\sum_{x\in X}\delta^x.
\end{equation*}
AJ is the average Jaccard of TAP-Vid,
\begin{equation*}
    \mathrm{AJ}=\frac{1}{|X|}\sum_{x\in X}\frac{\mathrm{TP}_x}{\mathrm{TP}_x+\mathrm{FP}_x+\mathrm{FN}_x},
\end{equation*}
where $\mathrm{TP}_x$ counts hits, $\mathrm{FP}_x$ counts queries predicted visible that are not hits, and $\mathrm{FN}_x$ counts visible targets that are not hits. A missing answer on an invisible target also counts as a false positive. OA, the occlusion accuracy, is the share of queries whose predicted visibility matches the ground truth, with missing answers counted as wrong. EPE averages $\min(d_q,32)$ over the visible targets, so a visible target predicted invisible or left unanswered contributes 32 pixels. All metrics pool the queries of a set rather than averaging over clips.

\paragraph{Original resolution.}
Tab.~\ref{tab:native-resolution} rescores the same predictions in native pixels, with $d_q$ computed without the rescaling above and the same thresholds and EPE cap. These scores depend on image resolution, so we compare them only within that table. \themodel still achieves the best score on every metric in both input settings, and with GPT-6 Astra it surpasses all expert vision models on all four metrics.

\paragraph{Expert vision models.}
RoMa v2~\citep{romav2} runs at its default precise setting and predicts the point visible when its overlap probability is at least 0.05. AllTracker~\citep{alltracker} uses its official evaluation settings at $384\times512$, and its high-resolution variant processes each clip with the long side at 1,024 pixels. It predicts the point visible when the product of its visibility and confidence is at least 0.6. CoTracker3~\citep{cotracker3} runs offline at $384\times512$ and predicts the point visible when its visibility is at least 0.6. With full video, the trackers receive every frame from the reference frame to the end of the clip, and with two images only the reference and target frames.

\paragraph{BootsTAPIR and TAPNext++.}
\label{sec:supp-tracker-baselines}
BootsTAPIR~\citep{bootstap} and TAPNext++~\citep{tapnextpp} in \Cref{tab:agent-comparison} use the released PyTorch checkpoints at $256\times256$ in FP32, without fine-tuning. For TAPNext++, this is the public PointOdyssey/Kubric-1024 checkpoint, not the final model of its paper trained with DynHumans. Both run on the full video. BootsTAPIR uses offline temporal refinement, and TAPNext++ runs causally, carrying its state across 16-frame chunks. BootsTAPIR predicts visible when $(1-\sigma(o))(1-\sigma(d))>0.5$, where $o$ and $d$ are its occlusion and expected-distance logits. TAPNext++ thresholds its visibility probability at 0.5. Predictions are mapped back to native coordinates and scored like all other methods.

\section{\themodel}
\label{sec:supp-agentap}

In \themodel, a VLM answers each query by calling a small image tool and reasoning over what it returns. This section describes the tool (Sec.~\ref{sec:supp-tool-interface}), the staged protocol and its prompt (Sec.~\ref{sec:supp-inference-protocol}), how each backbone is run (Sec.~\ref{sec:supp-agent-settings} and Tab.~\ref{tab:agent-settings}), its computational cost (Sec.~\ref{sec:supp-runtime-cost} and Tab.~\ref{tab:agent-cost}), and example sessions (Sec.~\ref{sec:supp-example-sessions}).

\subsection{Tools}
\label{sec:supp-tool-interface}

\begin{figure*}[tp]
    \centering
    \footnotesize
    \newlength{\toolpic}\setlength{\toolpic}{1.04in}
    \ifdefined\arxivbuild\setlength{\toolpic}{1.09in}\fi
    \newcommand{\toolline}[2]{{\scriptsize\ttfamily\frenchspacing\hangindent=1.35em\hangafter=1\noindent\makebox[1.35em][l]{\textcolor{black!50}{#1}}#2\par}}
    \newcommand{\toolcall}[1]{\toolline{\$}{#1}}
    \newcommand{\toolreturn}[1]{\toolline{$\rightarrow$}{#1}}
    \newcommand{\toolentry}[4]{%
        \noindent
        \begin{minipage}[t]{\toolpic}\vspace{0pt}\includegraphics[width=\linewidth]{figures/graphics/agent_tool_examples_blurred/#1.jpg}\end{minipage}\hspace{0.1in}%
        \begin{minipage}[t][\toolpic][t]{\dimexpr\linewidth-\toolpic-0.1in\relax}\vspace{-1.5pt}
            \raggedright\setlength{\parskip}{0pt}
            {\small\textbf{\texttt{#2}}}\par\vspace{2pt} #3\par\vspace{4pt} #4
        \end{minipage}\par}
    \begin{minipage}[t]{0.488\textwidth}
        \toolentry{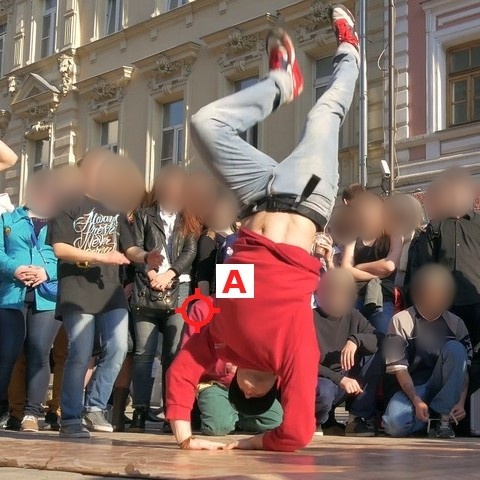}{marks}{Returns the exact centers of the markers drawn on the reference image.}
            {\toolcall{marks}\toolreturn{\{"center": [257.0, 310.5], \ldots\}}}
        \vspace{7pt}
        \toolentry{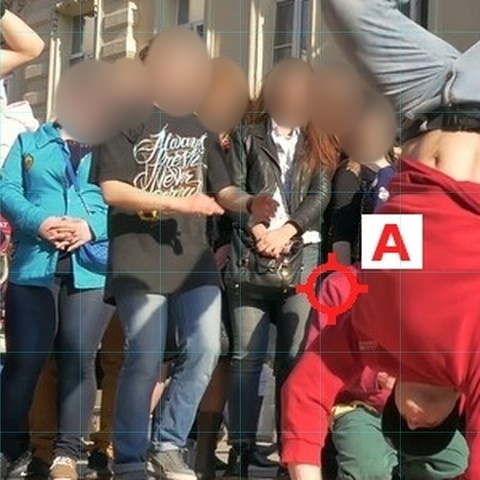}{view}{Renders a chosen region of the reference or the target under a grid in native pixel coordinates.}
            {\toolcall{view reference -{}-center 200 280 -{}-size 300}}
        \vspace{7pt}
        \toolentry{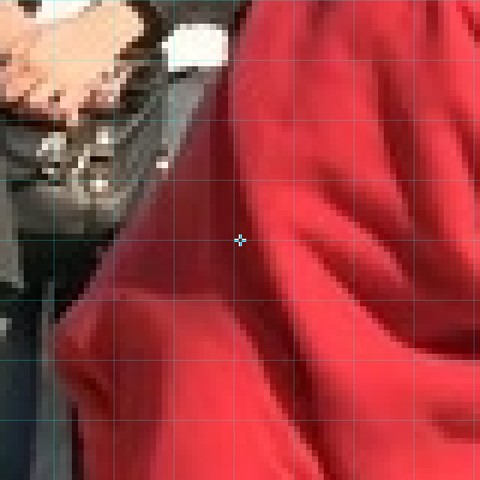}{view -{}-mark}{The same command can mark candidate coordinates (\texttt{-{}-mark}) or enhance the contrast (\texttt{-{}-enhance}).}
            {\toolcall{view target -{}-center 251 310 -{}-size 80 -{}-mark 251 310}}
        \vspace{7pt}
        \toolentry{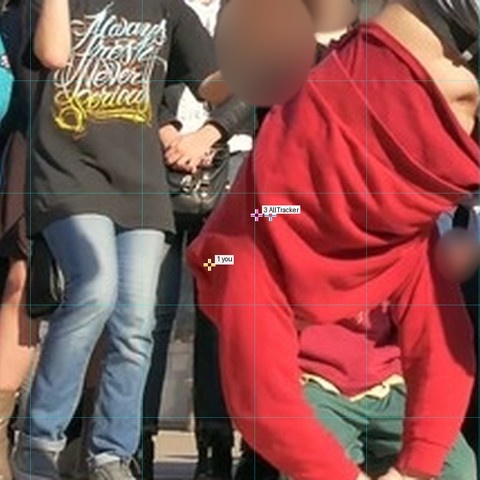}{commit}{Records the initial prediction and returns it next to the expert proposals.}
            {\toolcall{commit 230 332}}
    \end{minipage}\hfill
    \begin{minipage}[t]{0.488\textwidth}
        \toolentry{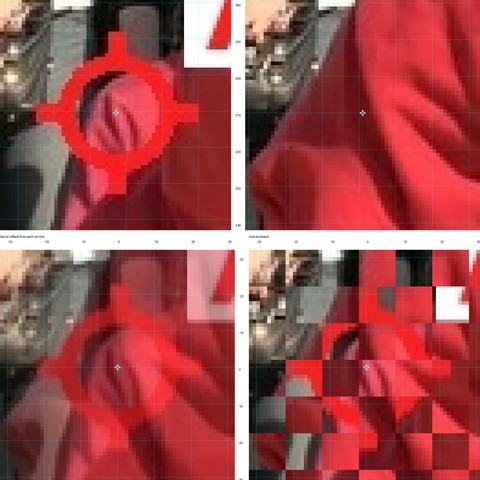}{check}{Compares the patches around A and around a candidate: side by side, blended, and as a checkerboard.}
            {\toolcall{check -{}-ref 257 310.5 -{}-tgt 251 310 -{}-size 64}}
        \vspace{7pt}
        \toolentry{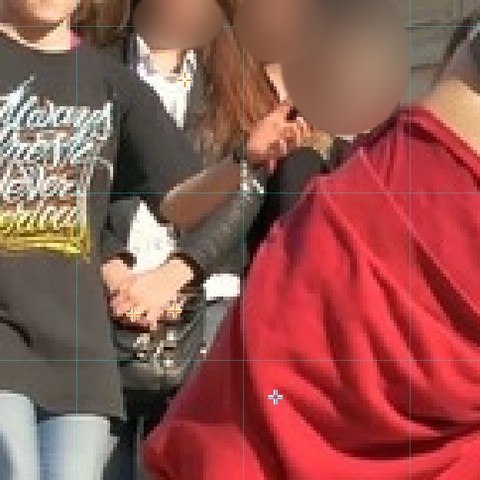}{fit}{Maps A through an affine transform or a homography fitted to landmark pairs chosen by the agent.}
            {\toolcall{fit -{}-pair 229.7 285.4 229.4 285.1 \ldots\ -{}-point 257 310.5}}
        \vspace{7pt}
        \toolentry{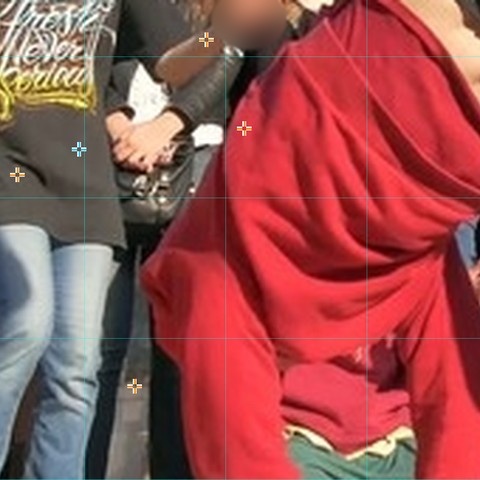}{match}{Searches a target box for the patch around A, by template matching or SIFT.}
            {\toolcall{match -{}-ref 257 310.5 -{}-size 48 -{}-search 150 230 360 400}}
        \vspace{7pt}
        \toolentry{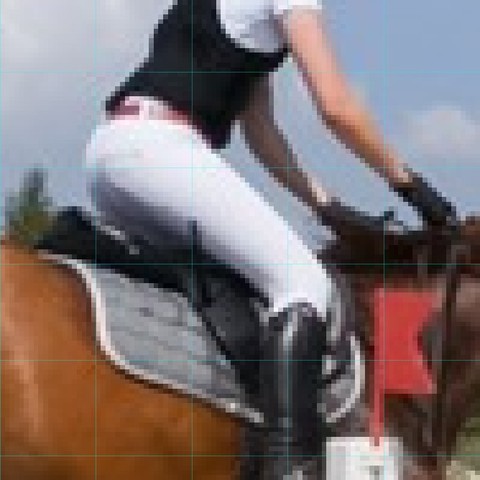}{frames{\normalfont,} frame}{List the intermediate frames and render one of them, or a region of it.}
            {\toolcall{frames}\toolreturn{\{"available\_frames": [1, 2, \ldots, 22], \ldots\}}\toolcall{frame 21 -{}-center 370 155 -{}-size 100}}
    \end{minipage}
    \caption{\textbf{Commands of the visual tool.} Each entry shows a command, an example call, and the returned image, cropped to a square. \texttt{marks} and \texttt{frames} return text (abridged). \texttt{marks} is shown with the reference image, and the \texttt{fit} call has four landmark pairs. The examples use a TAP-Vid DAVIS \emph{breakdance} pair (frames 0 and 67). \texttt{frames} and \texttt{frame} require intermediate-frame access, and their example is from the session in Fig.~\ref{fig:agent-session}.}
    \label{fig:agent-tool-examples}
\end{figure*}

Each query runs in its own container, which holds only the images of the query and the tool, with no network access and no dataset or ground-truth annotations. The agent calls the tool from the command line (Fig.~\ref{fig:agent-tool-examples}). The tools of Sec.~\ref{sec:method-tools} are commands of this one program, \texttt{zoom.py}. Two further commands, \texttt{commit} and \texttt{frames}, run the staged protocol and list the intermediate frames. Most commands return an image, which we call a view, overlaid with a coordinate grid in native image pixels (a 0--1000 grid for the two locally served models of Sec.~\ref{sec:supp-agent-settings}). \texttt{view} can mark a candidate position on the target or enhance contrast, and \texttt{check} shows the reference point next to a candidate position, side by side, blended, and as a checkerboard. \texttt{fit} and \texttt{match} return proposed coordinates that the agent can inspect, revise, or discard.

{\makeatletter\let\@xfloat\supp@xfloat\makeatother
\newlength{\promptskip}
\begingroup
    \setlength{\promptskip}{3pt}
    \newcommand{\promptsize}{\footnotesize}
    \newcommand{\promptbox}[2]{%
        \begingroup
        \setlength{\fboxrule}{0.4pt}\setlength{\fboxsep}{0pt}%
        \fcolorbox{black!30}{white}{%
            \begin{minipage}{\dimexpr\linewidth-2\fboxrule\relax}
                \setlength{\fboxsep}{4pt}%
                \colorbox{tabband}{\makebox[\dimexpr\linewidth-2\fboxsep\relax][l]{\strut\footnotesize\textbf{#1}}}\par
                \vspace{4pt}%
                \hspace*{7pt}\begin{minipage}{\dimexpr\linewidth-14pt\relax}
                    \promptsize\raggedright
                    \setlength{\parskip}{\promptskip}\setlength{\parindent}{0pt}%
                    #2
                \end{minipage}\par
                \vspace{5pt}%
            \end{minipage}}%
        \endgroup}
    \newcommand{\codesize}{\scriptsize}
    \newcommand{\code}[1]{\leavevmode{\ttfamily\codesize #1}}
    \newcommand{\note}[1]{\leavevmode{\codesize\color{black!60}#1}}
    \newcommand{\phead}[1]{\textbf{#1}}
    \newcommand{\ph}[1]{$\langle$\textit{#1}$\rangle$}
    \newcommand{\optionalstage}[1]{\par\noindent\colorbox{black!6}{\begin{minipage}{\dimexpr\linewidth-2\fboxsep\relax}\raggedright\setlength{\parskip}{3pt}\setlength{\parindent}{0pt}#1\end{minipage}}\par}
    \newenvironment{promptitems}{\begin{itemize}[leftmargin=1.3em,itemsep=0pt,parsep=0pt,topsep=0pt,partopsep=0pt]}{\end{itemize}}
    \newenvironment{commands}{\par\vspace{1pt}\begin{tabular}{@{\hspace{0.4em}}>{\raggedright\arraybackslash}p{0.645\linewidth}@{\hspace{0.4em}}>{\raggedright\arraybackslash}p{0.335\linewidth}@{}}}{\end{tabular}\par}
    \newcommand{\wide}[1]{\multicolumn{2}{@{\hspace{0.4em}}l@{}}{#1}}

\begin{figure*}[!t]
    \centering
    \renewcommand{\arraystretch}{1.1}
    \promptbox{Agent instructions}{%
        You are a visual-correspondence agent working in a sandbox. Your working directory holds exactly two images and one tool:
        \begin{promptitems}
            \item \code{reference.jpg}: IMAGE 1, the reference frame, with crosshair A and a magnified inset around it.
            \item \code{target.jpg}: IMAGE 2, the target frame, unmarked. It is \ph{W} x \ph{H} pixels.
            \item \code{zoom.py}: renders views of the images (\code{python3 zoom.py -h} lists its commands).
        \end{promptitems}
        There is nothing else to use: no internet, no other files, no dataset. Both images are attached to this message; look at them first.

        \phead{Goal}\\
        A marks the physical surface point at the exact centre of the crosshair, on one particular object or surface. Find that same point in IMAGE 2:
        \begin{promptitems}
            \item Follow the object, not the location: if it moved, turned or deformed, the point moved with it.
            \item Same instance, not a look-alike. Look-alike parts are easy to swap: left and right feet or hands, the faces of a cube. When the object or the camera turned, tell them apart by what is next to them.
            \item On a boundary, the point belongs to the surface at the exact centre.
            \item If the point itself is hidden, outside IMAGE 2, or facing away, answer occluded; otherwise give its position.
        \end{promptitems}

        Make a view with the tool, then open the file it prints with the \code{view\_image} tool. Every view has a grid labelled in that image's own native pixel coordinates, so you can read positions directly from it. All of your own vision tools are allowed too: \code{view\_image} opens any image here, and with detail ``original'' it shows the file at its full resolution instead of resized (useful for \code{reference.jpg} and \code{target.jpg} themselves).

        You work in three stages, in this order; zoom.py enforces them.

        \phead{Stage 1}, a first answer (at most 2 views, overviews only): is A visible in IMAGE 2, and if so roughly where?
        \begin{commands}
            \code{python3 zoom.py marks} & \note{red marks in the reference with exact centres (find A)} \\
            \code{python3 zoom.py view reference|target -{}-box X0 Y0 X1 Y1} & \note{a region showing at least a third of the image each way} \\
            \code{python3 zoom.py view reference|target -{}-center X Y -{}-size S [-{}-mark X Y]} & \note{the same, as a square; \code{-{}-mark} draws a point} \\
            \code{python3 zoom.py commit X Y} & \note{visible, roughly here; this ends stage 1} \\
            \code{python3 zoom.py commit occluded} & \note{not visible: hidden, outside IMAGE 2, or turned away} \\
        \end{commands}

        \optionalstage{%
        \phead{Stage 2}, the candidates (after commit X Y only; one view, shown by commit): where two automatic matchers put A, from the same two frames, drawn beside your first answer; a matcher appears only where it is confident of its point.
        \begin{promptitems}
            \item[] RoMa v2 (dense image matcher): very accurate on the static scene, usually wrong on objects that moved.
            \item[] AllTracker (point tracker, given the two frames): better on moving or deforming objects, but often off.
            \item[] When the two agree they are usually right. They place A; whether A is visible is yours to judge.
        \end{promptitems}
        After commit occluded there are no candidates: go on to stage 3 and look closer.}

        \phead{Stage 3}, check and answer (at most 5 views; every command, close zooms included):
        \begin{commands}
            \code{python3 zoom.py check -{}-ref X Y -{}-tgt X Y -{}-size 64} & \note{A and a candidate side by side, and blended/checkered} \\
            \code{python3 zoom.py view target -{}-center X Y -{}-size 80 -{}-mark X Y} & \note{zoom in; a cyan crosshair at X, Y} \\
            \code{python3 zoom.py view target -{}-center X Y -{}-size 120 -{}-enhance auto} & \note{brighter / more contrast (also bright, gamma, edges, gray)} \\
            \wide{\code{python3 zoom.py fit -{}-pair RX RY TX TY -{}-pair ... -{}-point X Y [-{}-model affine|homography]}} \\
            \wide{\hspace{1.5em}\note{map the reference point through a transform fitted to landmark pairs you are sure of}} \\
            \wide{\code{python3 zoom.py match -{}-ref X Y -{}-size 48 -{}-search X0 Y0 X1 Y1 [-{}-method template|sift]}} \\
            \wide{\hspace{1.5em}\note{appearance matching of the patch around A in a target region; proposals only}} \\
        \end{commands}

        Then answer: a point (yours, a candidate or a refined one) or occluded. Change whether A is visible from your first answer only for what you see yourself in a view.

        Your final message is the answer only: JSON in the schema below, with x and y in native target-image pixels.
    }

    \caption{\textbf{Agent instructions.} First part of the two-image prompt of \themodel, followed by the task text in Fig.~\ref{fig:agent-task-prompt}. Angle brackets denote query-specific values, and gray text is the comment of a command. The shaded stage concerns the optional expert proposals. The full-video prompt describes AllTracker as a full-video tracker and also lists the \texttt{frames} and \texttt{frame} commands.}
    \label{fig:agent-prompt}
\end{figure*}

\begin{figure*}[!t]
    \centering
    \promptbox{Task text}{%
        IMAGE 1 is the reference frame (frame \ph{r}). A single physical surface point is marked with a red crosshair labeled A. The inset panel shows a 2.75$\times$ magnified view of the 180 $\times$ 180 pixel region centered on A.

        All remaining images are unmarked target frames:
        \begin{promptitems}
            \item IMAGE 2: id \code{'}\ph{target id}\code{'}, frame \ph{t} (\ph{t$-$r} frames after the reference)
        \end{promptitems}

        \vspace{4pt}
        \begin{minipage}[t]{0.26\linewidth}\raggedright\setlength{\parskip}{3pt}
        \phead{Task}

        For each target frame, locate the same physical surface point marked as A in the reference frame.

        The apparent position, scale, orientation, and perspective of the object may change between frames. Your goal is to identify the same underlying surface point, defined by the center of the marked reference pixel.
        \end{minipage}\hfill
        \begin{minipage}[t]{0.38\linewidth}\raggedright\setlength{\parskip}{3pt}
        \phead{Guidelines}
        \begin{promptitems}
            \item Track the physical surface point, not its image coordinates.
            \item Use all available cues, including: local texture, semantic structure, geometry, neighboring features, and object shape.
            \item Ignore changes in camera viewpoint, scale, lighting, or partial deformation.
            \item If the point is occluded, outside the image, or on a surface that has rotated completely out of view, return occluded for that frame rather than estimating its location.
            \item Coordinates are native target-image pixels: origin (0, 0) at top-left, x right, y down, \code{0 <= x <} \ph{W}, \code{0 <= y <} \ph{H}.
        \end{promptitems}
        \end{minipage}\hfill
        \begin{minipage}[t]{0.30\linewidth}\raggedright\setlength{\parskip}{3pt}
        \phead{Output}
        \begin{promptitems}
            \item Return exactly one result for each target frame.
            \item Copy each target frame ID exactly as provided.
            \item Do not omit, reorder, or duplicate any target.
            \item Answer with JSON only, no other text:\\
            \code{\{"targets":[\{"id":"<target id>",}\\
            \code{\ "x":<number or null>,}\\
            \code{\ "y":<number or null>,}\\
            \code{\ "occluded":<true or false>\}]\}}\\
            Use null for x and y whenever occluded is true.
        \end{promptitems}
        \end{minipage}
    }
    \caption{\textbf{Task text.} Second part of the prompt, which the models also receive for direct prediction. Angle brackets denote query-specific values.}
    \label{fig:agent-task-prompt}
\end{figure*}
\endgroup
}
{\makeatletter\let\@xfloat\supp@xfloat\makeatother
\begin{table*}[t]
    \begin{minipage}[t]{0.485\textwidth}
        \vspace{0pt}
        \centering
            \fontsize{7.0}{8.4}\selectfont
    \tabcolsep=0.2em
    \newcommand{\yes}{$\checkmark$}
    \newcommand{\no}{\textemdash}
    \begin{tabular*}{\linewidth}{@{\extracolsep{\fill}}llllrcc@{}}
        \toprule
        \textbf{Backbone} & Serving & Agent program & Effort & Limit (s) & Proposals & Grid \\
        \midrule
        Qwen3.8-27B & local, FP8 & Codex & xhigh & 3,600 & \yes & 0--1000 \\
        Gemma-4-31B-it & local, FP8 & Codex & on & 3,600 & \yes & 0--1000 \\
        Inkling 2 Small & Tinker & Claude Code & high & 600 & \yes & native \\
        Muse Wonder Full & provider & Muse Code & high & 600 & \yes & native \\
        Claude Opus 5.5 & provider & Claude Code & high & 600 & \yes & native \\
        Claude Fable 5.2 & provider & Claude Code & high & 600 & \yes & native \\
        GPT-6.1 Sol & provider & Codex & high & 600 & \yes & native \\
        GPT-6 Astra & provider & Codex & high & 600 & \yes & native \\
        \bottomrule
    \end{tabular*}
    \caption{\textbf{Settings of \themodel per backbone.} Agent program: the harness that runs the model and executes its commands. Effort: reasoning effort. Limit: time per session. Proposals: staged protocol with RoMa v2 and AllTracker proposals. Grid: coordinates read and written by the agent. All backbones use the same inspection commands and at most eight rendered views per session.}
    \label{tab:agent-settings}

    \end{minipage}\hfill
    \begin{minipage}[t]{0.485\textwidth}
        \vspace{0pt}
        \centering
            \fontsize{7.0}{8.4}\selectfont
    \tabcolsep=0.25em
    \newcommand{\groupband}[1]{\rowcolor{tabband}\multicolumn{5}{l}{\emph{#1}}}
    \begin{tabular*}{\linewidth}{@{\extracolsep{\fill}}lcccc@{}}
        \toprule
        \textbf{Method} & Tok.\ in & Tok.\ out & Time (s) & AJ$\uparrow$ \\
        \midrule
        \groupband{Expert vision model} \\
        AllTracker HR, video & \textemdash & \textemdash & 4.6 & \secondbest{56.1} \\ \midrule
        \groupband{GPT-6.1 Sol} \\
        Direct & 3.2K & 1.1K & 23 & 40.6 \\
        \themodel & 166K & 1.1K & 91 & 53.6 \\ \midrule
        \groupband{GPT-6 Astra} \\
        \themodel & 141K & 0.7K & 40 & \thirdbest{54.9} \\
        \themodel, video & 153K & 0.6K & 44 & \best{58.5} \\ \midrule
        \groupband{Inkling 2 Small} \\
        Direct & 2.1K & 13.9K & 105 & 20.9 \\
        \themodel & 184K & 9.2K & 130 & 43.8 \\
        \themodel, video & 183K & 7.6K & 121 & 48.5 \\
        \bottomrule
    \end{tabular*}
    \caption{\textbf{Computational cost per query.} Mean tokens per query on the 2,000-query benchmark (Sec.~\ref{sec:supp-runtime-cost}). Time: median seconds per VLM session and mean GPU-seconds per AllTracker query. Sessions run in parallel. Inkling 2 Small ran 32 sessions at a time through Tinker and so took longer. Shading marks the \colorbox{tabfirst}{best}, \colorbox{tabsecond}{second}, and \colorbox{tabthird}{third} AJ.}
    \label{tab:agent-cost}

    \end{minipage}
\end{table*}
}

\subsection{Protocol}
\label{sec:supp-inference-protocol}

The protocol follows the three stages of Fig.~\ref{fig:agent}.

\paragraph{Stage 1: initial prediction.}
The agent opens at most two overviews, each covering at least a third of the image in both dimensions, and records its initial prediction with \texttt{commit}.

\paragraph{Stage 2: expert proposals.}
If the initial prediction is visible, \texttt{commit} returns one view that shows it next to the RoMa v2 and AllTracker proposals. An expert's proposal is shown only if that expert predicts the point visible and inside the image. The agent receives no other information about the experts' visibility decisions. After an invisible initial prediction, no proposals are shown.

\paragraph{Stage 3: verification.}
The agent then has five more views to verify its answer and may change both the position and the visibility. These limits count only views rendered by the tool, and the agent can also open the two original images. The final answer is a target position or an invisible decision, and a missing or invalid answer counts as a failure (Sec.~\ref{sec:supp-evaluation-protocol}).

\paragraph{Prompt.}
The prompt is the agent instructions (Fig.~\ref{fig:agent-prompt}) followed by the task text of the query (Fig.~\ref{fig:agent-task-prompt}). Direct prediction receives only the task text.

\paragraph{Full video.}
With full video, the RoMa v2 proposal stays the same, the AllTracker proposal comes from its high-resolution prediction on the whole video, and after the initial prediction the agent can also inspect intermediate frames with \texttt{frames} and \texttt{frame}. These views count toward the five verification views. The intermediate-frame ablation of Sec.~\ref{sec:supp-agent-ablations} keeps this proposal but shows only the two endpoint images.

\subsection{Settings}
\label{sec:supp-agent-settings}

Tab.~\ref{tab:agent-settings} lists how each backbone runs. Every backbone uses the same inspection commands with at most eight rendered views per session, and a session that ends without a valid answer counts as a failure. The six hosted backbones follow the staged protocol above. The two open-weight models served locally, Qwen3.8-27B and Gemma-4-31B-it, follow the same protocol with the same proposals on a 0--1000 coordinate grid, with longer time limits because a local model takes longer per step.

\subsection{Runtime and Computational Cost}
\label{sec:supp-runtime-cost}

\paragraph{Computational cost.}
Tab.~\ref{tab:agent-cost} reports tokens and time per query for GPT-6.1 Sol, GPT-6 Astra, and Inkling 2 Small. GPT-6 Astra opens on average 4.3 views per two-image session and 4.5 with video access, and GPT-6.1 Sol opens 5.2. The expert proposals add less than 7 GPU-seconds per query and are not included. For the free-form GPT-6.1 Sol agent of Sec.~\ref{sec:supp-multipoint}, sharing a session among 16 points lowers the mean tokens per query by 56--70\%. Sessions are independent, so they run in parallel. With 8 concurrent sessions, GPT-6 Astra with video access answers all 2,000 queries in 4.0 hours, about 500 queries per hour. While this computational cost is high, distilling the agent into an efficient model, such as a compact VLM or AllTracker, is a natural next step and could enable real-time tracking with \themodel.

\subsection{Example Sessions}
\label{sec:supp-example-sessions}

{\makeatletter\let\@xfloat\supp@xfloat\makeatother
\begin{figure*}[!t]
    \centering
    \includegraphics[width=\textwidth]{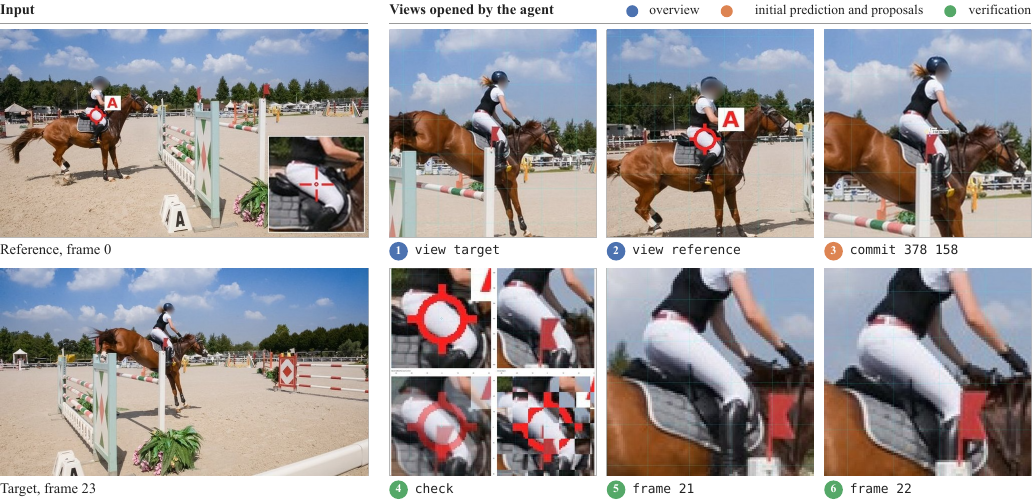}
    \caption{\textbf{Session 1: a target point that is hidden.} TAP-Vid DAVIS \emph{horsejump-high}, frames 0 and 23, with A on the rider's breeches.}
    \label{fig:agent-session}
\end{figure*}

\begin{figure*}[!t]
    \centering
    \includegraphics[width=\textwidth]{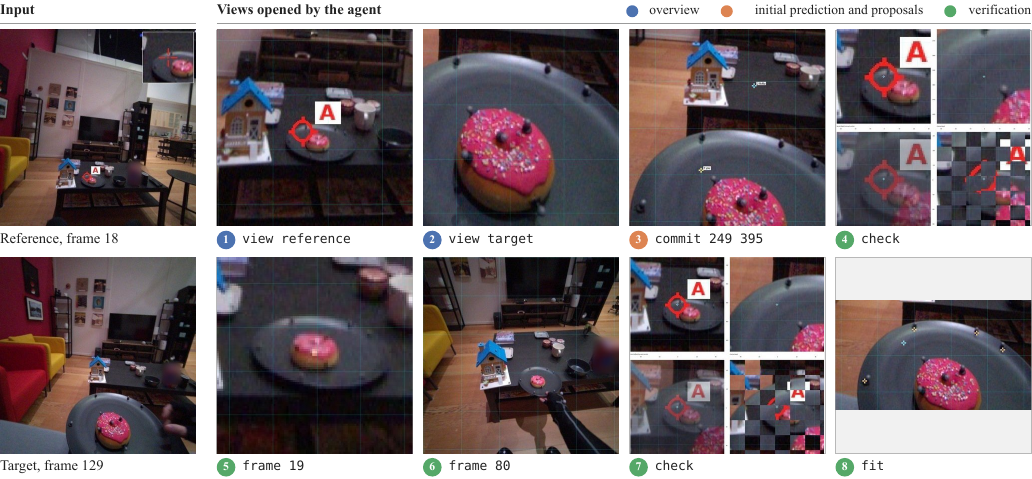}
    \caption{\textbf{Session 2: refinement with \texttt{fit}.} TAPVid-3D ADT, frames 18 and 129, with A on the plate behind the doughnut.}
    \label{fig:agent-session-fit}
\end{figure*}

\begin{figure*}[!t]
    \centering
    \includegraphics[width=\textwidth]{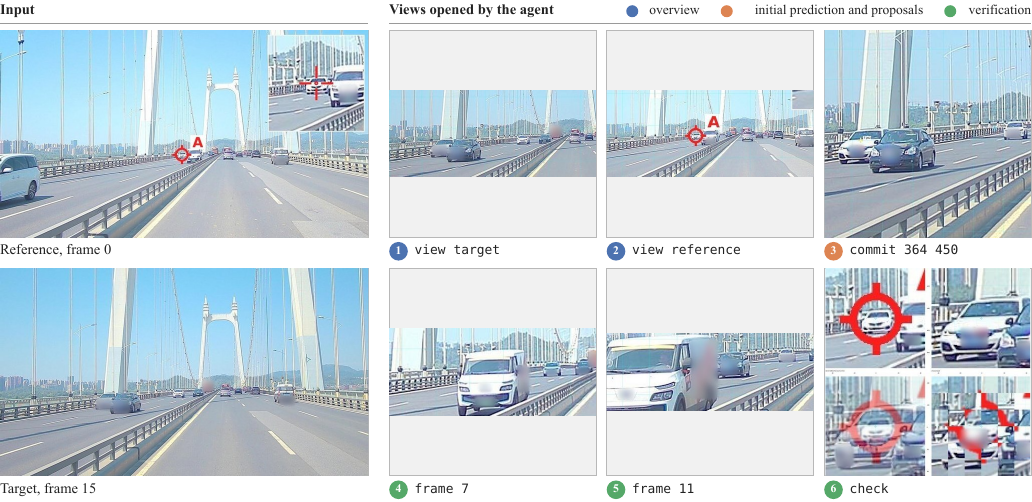}
    \caption{\textbf{Session 3: a point that reappears.} DrivTAP \emph{Day\_Bridge}, frames 0 and 15, with A on the front of a white sedan.}
    \label{fig:agent-session-reappear}
\end{figure*}
}

We show three full-video sessions of GPT-6 Astra on the 2,000-query benchmark. Each figure shows the views the agent opened and its final answer.

\paragraph{Session 1: a target point that is hidden.}
In Fig.~\ref{fig:agent-session}, the initial prediction, (378, 158), lies next to the full-video AllTracker proposal at (381, 161), and RoMa v2 gives no proposal. After the check (view~4), the agent notes that ``the point is very close to the top of the red jump flag'' and inspects frames 21 and 22. Its final answer is invisible, as in the ground-truth label, whereas AllTracker predicts the point visible.

\paragraph{Session 2: refinement with \texttt{fit}.}
In Fig.~\ref{fig:agent-session-fit}, the initial prediction is (249, 395). RoMa v2 proposes (306, 305), which the agent places on the table, and AllTracker predicts the point invisible. After inspecting frames 19 and 80, the agent fits a homography to four landmarks on the plate rim (view~8). The final answer, (235.7, 382.8), is 1.5 pixels from the ground-truth label on the $256\times256$ evaluation grid.

\paragraph{Session 3: a point that reappears.}
In Fig.~\ref{fig:agent-session-reappear}, the initial prediction is (364, 450). RoMa v2 proposes (585, 430), and AllTracker predicts the point invisible. In frames 7 and 11, the agent finds that ``the van briefly hides the marked car in the intermediate frames'' and that the sedan ``reappears to the left of the dark car in the target''. The final answer, (365, 451), is 1.0 pixel from the ground-truth label on the $256\times256$ evaluation grid.

\ifdefined\arxivbuild\null\newpage\fi
\section{Additional Results}
\label{sec:supp-results}
\renewcommand{\floatpagefraction}{0.85}
\renewcommand{\dblfloatpagefraction}{0.85}

This section extends Sec.~\ref{sec:experiments} of the main paper. Sec.~\ref{sec:supp-complete-results} completes the direct-prediction results, Sec.~\ref{sec:supp-localization-reappearance} analyzes \themodel, Sec.~\ref{sec:supp-agent-ablations} extends its ablations, Sec.~\ref{sec:supp-scaling} shares one session among several query points or target frames, and Sec.~\ref{sec:supp-qualitative} shows qualitative examples.

\subsection{Direct Prediction}
\label{sec:supp-complete-results}
\label{sec:supp-model-inputs}
\label{sec:supp-scaling-domains}

\paragraph{Inputs.}
Under direct prediction, VLMs receive the marked reference image and the target image with the task text of Fig.~\ref{fig:agent-task-prompt}. Two-image expert models receive the same two frames, and full-video trackers the video. The exception is Molmo2, which receives up to 16 frames sampled uniformly from reference to target, both included, with the query coordinates as text. Tables and figures mark it.

\paragraph{Models.}
The tables cite the model cards or provider documentation. Closed-weight model identifiers include \texttt{muse-wonder-full}, \texttt{claude-opus-5-5}, \texttt{claude-fable-5-2}, \texttt{gpt-6-astra}, and \texttt{gpt-6.1-sol}~\citep{muse-wonder-full,claude-opus5.5,claude-fable5.2,gpt6-astra,gpt6.1-sol}. For direct prediction, Claude Fable 5.2 runs in Claude Code, where it answers in one turn without tools, at high effort and with 600 seconds per query. The open-weight Inkling 2 Small~\citep{inkling2-small} is queried through the Tinker service~\citep{tinker}, and the other open-weight models run locally. On the held-out set of Sec.~\ref{sec:supp-heldout-annotation} (Fig.~\ref{fig:analysis}b), Claude Opus 5.5 and Inkling 2 Small run in Claude Code at high effort, GPT-6 Astra runs through a ChatGPT subscription backend, and Muse Wonder Full runs through its API at maximum reasoning effort, with its 48 failed requests counted as failures. For that figure, Claude Opus 5.5 runs in Claude Code on both query sets, which gives 32.3 AJ and 48.9 APD on the benchmark instead of 33.0 and 49.8.

\paragraph{Reporting.}
Methods in a comparison are scored on the same queries, with invalid answers counted as failures. Metrics are rounded once, half up, to one decimal in most tables, and improvements are computed before rounding.

\begin{table}[t]
\centering
\fontsize{7.0}{8.4}\selectfont
\setlength{\tabcolsep}{2pt}
\begin{tabular*}{\linewidth}{@{\extracolsep{\fill}}l*{4}{c}@{}}
\toprule
\textbf{Model} & AJ$\uparrow$ & APD$\uparrow$ & OA$\uparrow$ & EPE$\downarrow$ \\
\midrule
\rowcolor{black!7}\multicolumn{5}{l}{\textbf{Two images}} \\
Qwen3.5-0.8B~\citep{qwen3.5-0.8b} & 1.0 & 2.0 & 65.2 & 30.1 \\
Qwen3.5-2B~\citep{qwen3.5-2b} & 1.8 & 3.5 & 67.6 & 28.3 \\
Qwen3.5-4B~\citep{qwen3.5-4b} & 2.8 & 5.4 & 67.3 & 25.5 \\
Qwen3.5-9B~\citep{qwen3.5-9b} & 5.1 & 10.2 & 79.0 & 21.7 \\
Qwen3.5-27B~\citep{qwen3.5} & 6.5 & 10.8 & 67.6 & 22.1 \\
Qwen3.5-122B-A10B~\citep{qwen3.5-122b-a10b} & 10.3 & 18.7 & 79.2 & 17.5 \\
Qwen3.8-27B~\citep{qwen3.8-27b} & 14.9 & 26.0 & 80.1 & 14.7 \\
Qwen3.8-Flash-Next~\citep{qwen3.8-flash-next} & 14.6 & 24.6 & 79.2 & 15.5 \\
\addlinespace[2pt]
Gemma-4-31B-it~\citep{gemma4-31b-it} & 17.6 & 28.5 & 76.6 & 13.7 \\
DeepSeek-V4.1-Flash~\citep{deepseek-v4.1-flash} & 9.8 & 17.4 & 77.8 & 18.1 \\
\addlinespace[2pt]
Inkling-Small~\citep{inkling-small} & 2.5 & 4.9 & 76.0 & 25.3 \\
Inkling~\citep{inkling} & 2.3 & 4.5 & 76.2 & 25.9 \\
Inkling 2 Small~\citep{inkling2-small} & 20.9 & 33.4 & 79.1 & 12.5 \\
\addlinespace[2pt]
Kimi-K3~\citep{kimi-k3} & 26.3 & 41.1 & 81.8 & 10.2 \\
\addlinespace[2pt]
Muse Glimmer~\citep{muse-glimmer} & 7.2 & 13.4 & 78.8 & 19.6 \\
Muse Spark 1.3~\citep{muse-spark} & 12.4 & 20.7 & 79.4 & 16.9 \\
Muse Wonder Full~\citep{muse-wonder-full} & 29.5 & 42.7 & 83.0 & 8.9 \\
\addlinespace[2pt]
Claude Opus 5.5~\citep{claude-opus5.5} & 33.0 & 49.8 & 82.2 & 7.8 \\
Claude Fable 5.2~\citep{claude-fable5.2} & \cellcolor{tabthird}35.6 & \cellcolor{tabthird}52.7 & \cellcolor{tabthird}83.3 & \cellcolor{tabthird}6.9 \\
GPT-6.1 Sol~\citep{gpt6.1-sol} & \cellcolor{tabsecond}40.6 & \cellcolor{tabsecond}56.8 & \cellcolor{tabsecond}83.7 & \cellcolor{tabsecond}6.8 \\
GPT-6 Astra~\citep{gpt6-astra} & \cellcolor{tabfirst}43.2 & \cellcolor{tabfirst}59.4 & \cellcolor{tabfirst}85.5 & \cellcolor{tabfirst}6.5 \\
\midrule
\rowcolor{black!7}\multicolumn{5}{l}{\textbf{Video input}} \\
Molmo2-8B~\citep{molmo2-8b} & 12.6 & 21.2 & 65.9 & 19.1 \\
\bottomrule
\end{tabular*}
\caption{\textbf{All direct-prediction results.} All 22 VLMs on the 2,000-query benchmark, including Qwen3.5-0.8B and Qwen3.5-2B, which Fig.~\ref{fig:onecall-overview} of the main paper omits. Molmo2 receives up to 16 frames, and all other models receive two images. Shading marks the \colorbox{tabfirst}{best}, \colorbox{tabsecond}{second}, and \colorbox{tabthird}{third} results per column among the two-image models.}
\label{tab:onecall-all-2k}
\end{table}

\paragraph{All models.}
Tab.~\ref{tab:onecall-all-2k} gives AJ, APD, OA, and EPE of all 22 VLMs on the 2,000-query benchmark, including Qwen3.5-0.8B and Qwen3.5-2B, which Fig.~\ref{fig:onecall-overview} omits. Direct prediction rarely calls a point hidden. Of the 21 two-image VLMs (all but Molmo2), 17 predict at least 85\% of the queries visible (median 89.1\%), although 21.0\% of the targets are hidden.

\begin{table}[t]
    \centering
    \fontsize{7.0}{8.4}\selectfont
    \tabcolsep=0.3em
    \begin{tabular*}{\linewidth}{@{\extracolsep{\fill}}lcccc@{}}
        \toprule
        \textbf{Model} & \textbf{Real} & \textbf{Kubric} & \textbf{RGB-Stacking} & \textbf{Synthetic} \\
        \midrule
        \rowcolor{black!7}\multicolumn{5}{l}{\textbf{Two images}} \\
        \rowcolor{tabband}\multicolumn{5}{l}{\emph{Expert vision models}} \\
        RoMa v2 & 52.1 & \thirdbest{59.5} & 60.4 & \thirdbest{59.9} \\
        AllTracker & \secondbest{58.2} & \best{66.2} & \best{70.3} & \best{68.2} \\
        AllTracker, HR & \thirdbest{57.0} & \secondbest{65.4} & \secondbest{65.6} & \secondbest{65.5} \\
        \rowcolor{tabband}\multicolumn{5}{l}{\emph{VLMs}} \\
        Qwen3.8-27B & 26.4 & 8.1 & 7.8 & 8.0 \\
        Kimi-K3 & 41.2 & 9.1 & 19.2 & 14.0 \\
        Claude Opus 5.5 & 50.5 & 28.8 & 49.7 & 38.8 \\
        GPT-6 Astra & \best{59.7} & 35.2 & \thirdbest{62.0} & 48.1 \\
        \midrule
        \rowcolor{black!7}\multicolumn{5}{l}{\textbf{Full video}} \\
        \rowcolor{tabband}\multicolumn{5}{l}{\emph{Expert vision models}} \\
        CoTracker3 & \thirdbest{64.3} & \thirdbest{74.4} & \thirdbest{71.1} & \thirdbest{72.8} \\
        AllTracker & \secondbest{67.6} & \secondbest{75.3} & \best{81.1} & \best{78.1} \\
        AllTracker, HR & \best{68.2} & \best{77.2} & \secondbest{75.0} & \secondbest{76.2} \\
        \bottomrule
    \end{tabular*}
    \caption{\textbf{Real and synthetic correspondence.} Two-image GPT-6 Astra leads AllTracker on real videos but trails it on synthetic ones. APD (\%) on 12,496 real and 2,400 synthetic queries (Kubric and RGB-Stacking, 1,200 each). All VLMs use direct prediction. HR: high resolution. Shading marks the \colorbox{tabfirst}{best}, \colorbox{tabsecond}{second}, and \colorbox{tabthird}{third} results per column within an input block.}
    \label{tab:synthetic-comparison}
\end{table}

\paragraph{Real and synthetic videos.}
\phantomsection\label{sec:supp-real-synthetic}
Synthetic videos change the ordering of VLMs and expert vision models (Tab.~\ref{tab:synthetic-comparison}). With the same two images, GPT-6 Astra slightly leads AllTracker in APD on real videos, but AllTracker leads clearly on the synthetic subset and on each synthetic source. Synthetic imagery is not uniformly hard for VLMs, since GPT-6 Astra scores higher on RGB-Stacking than on real videos. Full-video trackers score higher on both synthetic sources than on real videos. The synthetic column pools the queries of both sources.

\subsection{Agentic Inference}
\label{sec:supp-localization-reappearance}

\paragraph{Reappearance.}
\phantomsection\label{sec:supp-reappearance-selection}
A visible target is reappearing if an intermediate annotation marks it invisible, and continuously visible if every intermediate frame is labeled visible. Of the 1,579 visible targets, 933 are continuously visible, 456 reappear, and 190 have unknown histories and are left out of this split. On the 456 reappearing targets, the five full-video trackers (CoTracker3, BootsTAPIR, TAPNext++, and AllTracker at both resolutions) predict 15--28\% invisible, 22\% on average, against 8\% for two-image \themodel with GPT-6 Astra. Since APD and AJ count these predictions as misses, \themodel leads the trackers by 7.6--15.7 APD and 3.1--11.6 AJ. During continuous visibility, high-resolution AllTracker and two-image \themodel have similar APD (75.4 and 75.9), while the tracker leads in AJ (67.7 and 65.7) and one-pixel accuracy (49.3 and 46.8).

Full-video \themodel improves reappearance correspondence further. On all 456 reappearing targets it reaches 67.4 APD and 57.7 AJ, against 57.5 and 52.1 for high-resolution AllTracker, and predicts 38 of them invisible (with one invalid answer), where the tracker rejects 124. Over the whole benchmark, the two differ in visibility on 270 queries, and the agent is right on 215, 192 of them visible targets that the tracker marks invisible. Where both predict visible, their positions agree within one pixel on the $256\times256$ grid in 1,388 of 1,445 cases.

\begin{figure*}[!t]
    \centering
    \includegraphics[width=\textwidth]{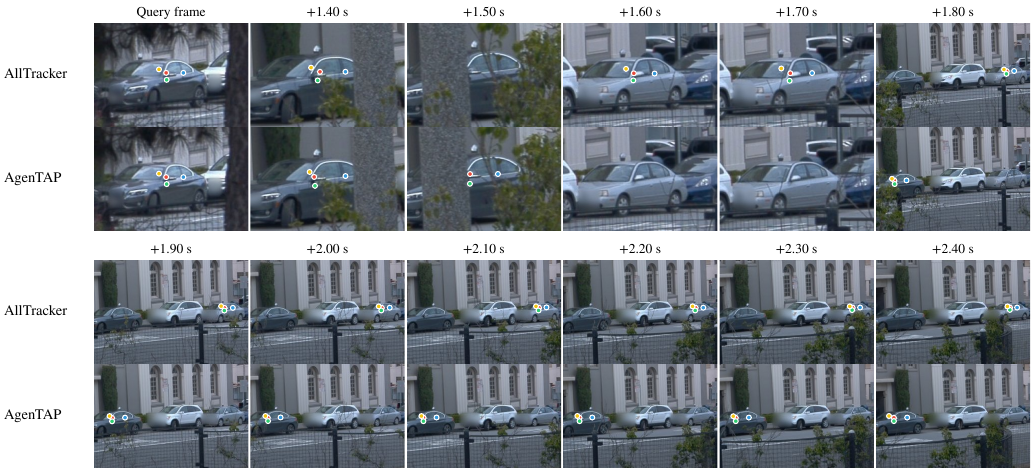}
    \caption{\textbf{A car seen again after an occlusion.} AllTracker (upper rows) and full-video \themodel with GPT-6 Astra (lower rows) predict the same query points on a parked car filmed from a moving vehicle (DriveTrack). After the car reappears, AllTracker places the points on another car, while \themodel finds the queried one again. The query frame is followed by the last eleven of 24 evaluated target frames. Every panel is cropped around both methods' predictions for four of the six query points. Points predicted invisible are not drawn.}
    \label{fig:tracker-comparison-driving}
\end{figure*}

\begin{figure*}[!t]
    \centering
    \includegraphics[width=\textwidth]{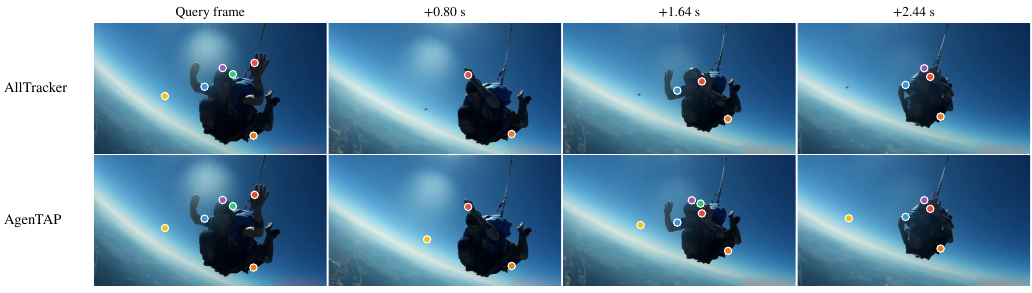}\\[4pt]
    \includegraphics[width=\textwidth]{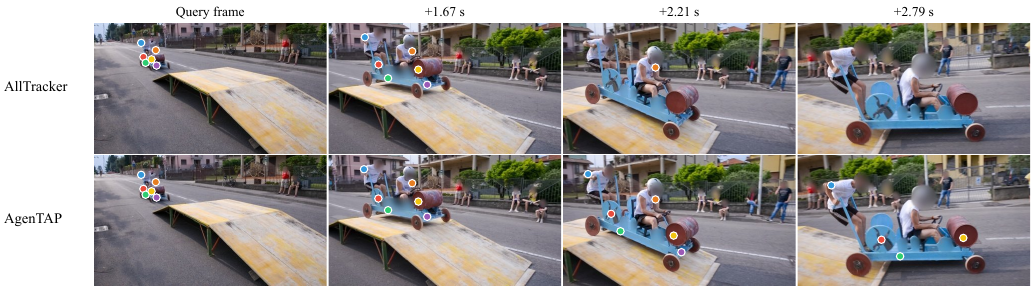}
    \caption{\textbf{Visible points during rotation.} AllTracker (upper rows) and full-video \themodel with GPT-6 Astra (lower rows) predict the same query points, with a reference frame and three evaluated target frames per clip. Top: a tumbling skydiver (TAP-Vid Kinetics). AllTracker marks the point on the distant second skydiver (yellow) as invisible, while \themodel continues to locate it. Bottom: a soapbox cart on a ramp (TAP-Vid DAVIS). As the cart turns, AllTracker marks most of the visible points as invisible, while \themodel still locates them. Dots show predicted positions, and points predicted invisible are not drawn.}
    \label{fig:tracker-comparison-rotation}
\end{figure*}

\paragraph{Comparison with AllTracker.}
\phantomsection\label{sec:supp-tracker-comparison}
Figs.~\ref{fig:tracker-comparison-driving} and~\ref{fig:tracker-comparison-rotation} compare full-video \themodel with GPT-6 Astra and AllTracker on the same video and query points. When the car reappears, AllTracker places the points on another car, while \themodel finds the queried car again. As the skydiver tumbles and the soapbox cart turns, AllTracker marks points that remain visible as invisible, while \themodel still locates them.

\paragraph{From initial prediction to final answer.}
\label{sec:supp-initial-to-final}
Fig.~\ref{fig:analysis}f of the main paper follows the 2,000 two-image GPT-6 Astra sessions from the initial prediction to the final answer. An answer is correct if it is visible and within the threshold for a visible target, and invisible for an invisible one. The six invalid final answers count as incorrect. At 4 pixels, 276 of 769 initial errors become correct and 53 of 1,231 correct answers regress, raising joint correctness from 61.55\% to 72.70\%. The corrected errors comprise 223 of 475 position errors, 3 of 47 missed-visible targets, and 50 of 247 false-visible predictions, so most recoveries come from localization. For visible targets with a visible initial prediction and every shown expert proposal wrong, the agent keeps 90 of 129 correct initial predictions (69.8\%) and corrects 32 of 226 wrong ones (14.2\%). These correct final answers depart from every proposal.

\subsection{Ablations}
\label{sec:supp-agent-ablations}

\paragraph{Fine and broad thresholds.}
\Cref{tab:agent-components-details} adds the finest and broadest thresholds to the ablation of Tab.~\ref{tab:method-ablation} on the same queries and runs, together with the initial prediction each session commits before any expert proposal is shown. The free-form agent, used in Sec.~\ref{sec:supp-scaling} and for the label audit of Sec.~\ref{sec:supp-benchmark-composition}, runs GPT-6.1 Sol with the same commands but without the stages of Sec.~\ref{sec:supp-inference-protocol}. The initial prediction of GPT-6 Astra, committed after at most two overviews, places 23.8\% of visible targets within one pixel, as direct prediction does, and 90.4\% within 16 pixels, above the 87.1\% of direct prediction. Its final answer places 40.2\% within one pixel and 91.0\% within 16 pixels, and the agent without proposals 34.4\% and 90.2\%. Inkling 2 Small depends more on the proposals. Its initial prediction places only 2.7\% within one pixel, and proposals raise its final share from 12.7\% to 33.6\%. The improvements thus lie mainly in fine localization.

\begin{table}[t]
    \centering
    \fontsize{7.0}{8.4}\selectfont
    \tabcolsep=0.5em
    \newcommand{\groupband}[1]{\rowcolor{tabband}\multicolumn{3}{l}{\emph{#1}}}
    \begin{tabular*}{\linewidth}{@{\extracolsep{\fill}}lcc@{}}
        \toprule
        \textbf{Setting} & $\delta^{1}\uparrow$ & $\delta^{16}\uparrow$ \\
        \midrule
        \groupband{Expert vision models} \\
        Expert oracle & 37.6 & 73.1 \\ \midrule
        \groupband{GPT-6 Astra} \\
        Initial prediction (stage 1) & 23.8 & 90.4 \\
        Without proposals & 34.4 & 90.2 \\
        \themodel & 40.2 & 91.0 \\ \midrule
        \groupband{GPT-6.1 Sol} \\
        Free-form agent & 32.8 & 90.1 \\ \midrule
        \groupband{Inkling 2 Small} \\
        Initial prediction (stage 1) & 2.7 & 68.7 \\
        Without proposals & 12.7 & 81.4 \\
        \themodel & 33.6 & 86.3 \\
        \bottomrule
    \end{tabular*}
    \caption{\textbf{Expert proposals and fine localization.} Most of the agent's improvement is in fine localization. $\delta^{1}$, $\delta^{16}$: shares of visible targets predicted visible within 1 and 16 pixels on the 2,000-query benchmark. AJ and APD of the settings shared with Tab.~\ref{tab:method-ablation} of the main paper are in that table. Initial prediction: the answer the agent commits before any proposal is shown. Oracle: whichever of RoMa v2 and two-frame AllTracker is closer to the ground truth, with AllTracker's visibility.}
    \label{tab:agent-components-details}
\end{table}

\paragraph{Zoom harness.}
The zoom harness in Tab.~\ref{tab:method-ablation} is a fixed pipeline. Starting from a direct prediction, it shows the VLM marked close-ups of the reference, target overviews, and target close-ups, for at most two rounds of refinement. Unlike in \themodel, the VLM does not choose what to inspect.

\paragraph{Use of proposals.}
When a proposal is shown, two-image \themodel with GPT-6 Astra keeps one within 1 pixel on the $256\times256$ grid in 76.0\% of its visible answers. Its revisions on visible targets end closer to the ground truth than the nearest proposal in 74.1\% of cases, farther in 19.0\%, and within 0.5 pixel in the rest. Inkling 2 Small keeps a proposal in 67.3\% of its visible answers, and its revisions end closer in 53.5\% and farther in 38.7\%.

\paragraph{Expert oracle.}
Beyond the two-image expert oracle of Sec.~\ref{sec:ablations}, the full-video oracle of Tab.~\ref{tab:method-ablation} chooses in the same way between RoMa v2 and high-resolution full-video AllTracker, with the visibility of that tracker. Even with ground-truth visibility, the two-image oracle reaches only 53.1 AJ and 67.3 APD, below two-image \themodel with GPT-6 Astra (54.9 AJ and 71.4 APD).

\paragraph{Intermediate frames.}
With the full-video AllTracker proposal but only the endpoint images, GPT-6 Astra reaches 56.9 AJ, 73.2 APD, 87.1 OA, and 4.3 EPE. Inspecting intermediate frames, with the same proposals, raises these to 58.5, 73.9, 88.2, and 4.1.

\begingroup
\clubpenalty=10000
\widowpenalty=10000
\subsection{Scaling Reference Points and Target Frames}
\label{sec:supp-scaling}

The benchmark results of the main paper ask about one query point in one target frame per session. Here we test whether sessions that share several query points in the reference frame, or several target frames of one query point, keep this accuracy, and how their token usage changes.

\paragraph{Scaling queries.}
We construct a separate query pool for multi-point and multi-frame tracking from the clips and ground-truth tracks of \thebench. For each clip, we select one reference frame and 16 points that remain visible at target frames 18, 36, and 54 frames later. We choose points farthest-first, keeping them at least 5\% of the image width apart so that their markers do not overlap. The pool has three subsets: points with displacement, points with negligible displacement, and clips containing both. Tab.~\ref{tab:multi-point-queries} reports their sizes, displacement thresholds, median displacements, and source datasets. Each query pairs a point with a target frame. Both studies use this query pool, grouping points or target frames into sessions. Their sessions use the free-form agent of Tab.~\ref{tab:agent-components-details}, not the staged protocol of Sec.~\ref{sec:supp-inference-protocol}: the agent calls the tool's commands in any order, without \texttt{commit} or expert proposals.

\paragraph{Session settings.}
For multi-point and multi-frame sessions, we increase the budget by 3 rendered views, 10 model requests, and 300 seconds for each additional point or target frame.

\begin{table}[!t]
\centering
\fontsize{7.0}{8.4}\selectfont
\setlength{\tabcolsep}{1.5pt}
\renewcommand{\arraystretch}{1.1}
\begin{tabular*}{\linewidth}{@{\extracolsep{\fill}}lcccccl@{}}
\toprule
\textbf{Subset} & \textbf{Clips} & \textbf{Points} & \textbf{Queries} & \shortstack{\textbf{Rule}\\\textbf{(px)}} & \shortstack{\textbf{Median}\\\textbf{disp. (px)}} & \textbf{Sources} \\
\midrule
Displacement & 16 & 256 & 768 & $\geq$8 & 40.3 & ADT 15, PT 1 \\
\addlinespace[2pt]
\begin{tabular}[c]{@{}l@{}}Negligible\\displacement\end{tabular} & 11 & 176 & 528 & $<$2 & 0.1 & PT 10, RoboTAP 1 \\
\addlinespace[2pt]
Mixed & 21 & 336 & 1,008 & $\geq$6 or $<$3 & 12.4 / 0.2 & \begin{tabular}[c]{@{}l@{}}PT 13, ADT 4,\\RoboTAP 4\end{tabular} \\
\bottomrule
\end{tabular*}
\caption{\textbf{Queries of the multi-point and multi-frame study.} Each clip contributes 16 ground-truth points and three target frames per point. Rule: displacement from the reference position required at every target frame. Median disp.: median displacement over the queries. Both are at the 256-pixel scale. For the mixed subset the median is given for its 91 points with displacement and 245 points with negligible displacement. PT: Perception Test.}
\label{tab:multi-point-queries}
\end{table}

\subsubsection{Multi-point Tracking}
\label{sec:supp-multipoint}

\paragraph{Tracking performance with more query points.}
We run the free-form agent with GPT-6.1 Sol at high reasoning effort, grouping the same point--target queries into sessions of 1, 2, 4, 8, or 16 points: 768 queries with point displacement, 528 with negligible displacement, and 1,008 in a mixed subset. For sessions of 2, 4, or 8 points, we group the points so that those in a session are as far apart as possible. In all three subsets, APD at every multi-point setting stays close to or slightly above the single-point baseline (Tab.~\ref{tab:multi-point}).

{\makeatletter\let\@xfloat\supp@xfloat\makeatother
\begin{table}[!t]
\centering
\fontsize{7.0}{8.4}\selectfont
\begin{tabular*}{\linewidth}{@{\extracolsep{\fill}}cccc@{}}
\toprule
\textbf{Points/session} & AJ$\uparrow$ & APD$\uparrow$ & OA$\uparrow$ \\
\midrule
\rowcolor{black!7}\multicolumn{4}{l}{\textbf{Points with displacement}} \\
1 & 64.58 & 72.81 & 96.48 \\
2 & 65.61 & 74.09 & 97.27 \\
4 & 64.39 & 73.31 & 96.61 \\
8 & 64.49 & 73.15 & 96.48 \\
16 & 65.43 & 74.30 & 97.27 \\
\midrule
\rowcolor{black!7}\multicolumn{4}{l}{\textbf{Points with negligible displacement}} \\
1 & 94.01 & 96.33 & 99.24 \\
2 & 94.21 & 96.55 & 99.43 \\
4 & 94.57 & 96.78 & 99.43 \\
8 & 94.75 & 96.86 & 99.43 \\
16 & 94.71 & 96.86 & 99.43 \\
\midrule
\rowcolor{black!7}\multicolumn{4}{l}{\textbf{Mixed}} \\
1 & 81.01 & 87.70 & 98.21 \\
2 & 81.52 & 88.39 & 98.61 \\
4 & 81.12 & 87.92 & 98.31 \\
8 & 82.60 & 88.69 & 97.72 \\
16 & 82.45 & 88.81 & 98.21 \\
\bottomrule
\end{tabular*}
\caption{\textbf{Accuracy with more points.} Accuracy holds from 1 to 16 points per session. Free-form agent, GPT-6.1 Sol, same point--target queries. AJ, APD, and OA in \%. All evaluated targets are visible, so OA measures visible-target recall, with invalid outputs counted as errors.}
\label{tab:multi-point}
\end{table}
}

\paragraph{Token efficiency under reference scaling.}
Tab.~\ref{tab:multi-point-tokens} divides each run's usage by its query count $N$. Input tokens include cached tokens and output tokens include reasoning, so the mean total \mbox{per query is}
\begin{equation}
    \bar{T}=\frac{T_{\mathrm{in}}+T_{\mathrm{out}}}{N}.
    \label{eq:multi-point-tokens}
\end{equation}
From one to sixteen points, this mean falls by 56--70\% across the three subsets.

\begin{table}[!t]
\centering
\fontsize{7.0}{8.4}\selectfont
\setlength{\tabcolsep}{2pt}
\begin{tabular*}{\linewidth}{@{\extracolsep{\fill}}crrrrr@{}}
\toprule
& \multicolumn{4}{c}{\textbf{Tokens per query}} & \\
\cmidrule(lr){2-5}
\textbf{Points/session} & Input & Uncached & Output & Total & Ratio \\
\midrule
\rowcolor{black!7}\multicolumn{6}{l}{\textbf{Points with displacement}} \\
1 & 201,147 & 20,173 & 2,309 & 203,456 & 1.00$\times$ \\
2 & 144,747 & 13,318 & 1,928 & 146,675 & 1.39$\times$ \\
4 & 98,824 & 10,661 & 1,334 & 100,159 & 2.03$\times$ \\
8 & 88,991 & 6,594 & 1,047 & 90,039 & 2.26$\times$ \\
16 & 88,808 & 5,402 & 829 & 89,637 & 2.27$\times$ \\
\midrule
\rowcolor{black!7}\multicolumn{6}{l}{\textbf{Points with negligible displacement}} \\
1 & 132,500 & 17,382 & 653 & 133,153 & 1.00$\times$ \\
2 & 92,557 & 12,514 & 560 & 93,117 & 1.43$\times$ \\
4 & 65,317 & 9,051 & 447 & 65,764 & 2.02$\times$ \\
8 & 45,613 & 5,547 & 309 & 45,922 & 2.90$\times$ \\
16 & 40,137 & 3,693 & 249 & 40,386 & 3.30$\times$ \\
\midrule
\rowcolor{black!7}\multicolumn{6}{l}{\textbf{Mixed}} \\
1 & 149,023 & 19,522 & 1,031 & 150,054 & 1.00$\times$ \\
2 & 107,463 & 12,817 & 915 & 108,378 & 1.38$\times$ \\
4 & 75,763 & 9,395 & 694 & 76,457 & 1.96$\times$ \\
8 & 62,628 & 6,288 & 566 & 63,194 & 2.37$\times$ \\
16 & 59,351 & 4,776 & 453 & 59,804 & 2.51$\times$ \\
\bottomrule
\end{tabular*}
\caption{\textbf{Tokens with more points.} Tokens per query fall as more points share a session. Input includes cached input, and Uncached is its uncached part. Output includes reasoning. Total is $\bar{T}$ of Eq.~\ref{eq:multi-point-tokens}. Ratio: single-point Total over the current Total in the same subset.}
\label{tab:multi-point-tokens}
\end{table}

\begin{table*}[t]
    \centering
    \newcommand{\framedelta}[1]{\raisebox{-0.4ex}{\fontsize{4.8}{5.2}\selectfont\textcolor{black!60}{#1}}}
    \fontsize{7.0}{8.4}\selectfont
    \setlength{\tabcolsep}{3pt}
    \begin{tabular*}{\textwidth}{@{\extracolsep{\fill}}clccc@{}}
        \toprule
        & & \multicolumn{1}{c}{\textbf{First target frame (gap 18)}} & \multicolumn{2}{c}{\textbf{All target frames}} \\
        \cmidrule(lr){3-3}\cmidrule(l){4-5}
        \textbf{Frames} & \textbf{Gaps} & APD$\uparrow$ & \shortstack{Separately APD$\uparrow$\\{\fontsize{6}{7}\selectfont One target/session}} & \shortstack{Together APD$\uparrow$\\{\fontsize{6}{7}\selectfont All targets/session}} \\
        \midrule
        \rowcolor{black!7}\multicolumn{5}{l}{\textbf{Points with displacement}\ \ \emph{256 points}} \\
        1 & \{18\} & 78.7 & \textemdash & \textemdash \\
        2 & \{18, 36\} & \best{84.8\framedelta{$+$6.1}} & 76.6 & \best{82.0\framedelta{$+$5.4}} \\
        3 & \{18, 36, 54\} & 83.9\framedelta{$-$0.9} & 72.8 & \best{79.6\framedelta{$+$6.8}} \\
        \midrule
        \rowcolor{black!7}\multicolumn{5}{l}{\textbf{Points with negligible displacement}\ \ \emph{176 points}} \\
        1 & \{18\} & 97.4 & \textemdash & \textemdash \\
        2 & \{18, 36\} & \best{97.5\framedelta{$+$0.1}} & 96.9 & \best{97.3\framedelta{$+$0.4}} \\
        3 & \{18, 36, 54\} & 97.4\framedelta{$-$0.1} & 96.3 & \best{96.6\framedelta{$+$0.3}} \\
        \midrule
        \rowcolor{black!7}\multicolumn{5}{l}{\textbf{Mixed}\ \ \emph{336 points}} \\
        1 & \{18\} & 89.4 & \textemdash & \textemdash \\
        2 & \{18, 36\} & 90.8\framedelta{$+$1.4} & 87.9 & \best{90.1\framedelta{$+$2.2}} \\
        3 & \{18, 36, 54\} & \best{91.4\framedelta{$+$0.6}} & 87.7 & \best{90.1\framedelta{$+$2.4}} \\
        \bottomrule
    \end{tabular*}
    \caption{\textbf{Accuracy with more target frames per session.} Adding frames keeps first-frame accuracy, and frames predicted together are more accurate than predicted separately. All rows use the same points. Left: APD on the first target frame (gap 18). Subscripts show the change from the preceding frame-count setting within each subset. Right: APD on identical point--frame pairs. Separately requests one target frame per session, and Together requests all target frames in one session. Subscripts on Together show Together minus Separately. Shading marks the highest first-frame APD in each subset and \mbox{the higher of Separately and Together}.}
    \label{tab:multi-frame-accuracy}
\end{table*}

\begin{table}[!t]
\centering
\fontsize{7.0}{8.4}\selectfont
\setlength{\tabcolsep}{2pt}
\begin{tabular*}{\linewidth}{@{\extracolsep{\fill}}crrrrr@{}}
\toprule
& \multicolumn{4}{c}{\textbf{Tokens per answer}} & \\
\cmidrule(lr){2-5}
\textbf{Frames/session} & Input & Uncached & Output & Total & Ratio \\
\midrule
\rowcolor{black!7}\multicolumn{6}{l}{\textbf{Points with displacement}} \\
1 & 201,147 & 20,173 & 2,309 & 203,456 & 1.00$\times$ \\
2 & 135,791 & 13,985 & 1,890 & 137,681 & 1.48$\times$ \\
3 & 113,174 & 10,903 & 1,690 & 114,864 & 1.77$\times$ \\
\midrule
\rowcolor{black!7}\multicolumn{6}{l}{\textbf{Points with negligible displacement}} \\
1 & 132,500 & 17,382 & 653 & 133,153 & 1.00$\times$ \\
2 & 80,307 & 10,968 & 470 & 80,777 & 1.65$\times$ \\
3 & 63,483 & 8,756 & 383 & 63,866 & 2.08$\times$ \\
\midrule
\rowcolor{black!7}\multicolumn{6}{l}{\textbf{Mixed}} \\
1 & 149,023 & 19,522 & 1,031 & 150,054 & 1.00$\times$ \\
2 & 96,673 & 13,007 & 835 & 97,508 & 1.54$\times$ \\
3 & 73,941 & 10,033 & 635 & 74,576 & 2.01$\times$ \\
\bottomrule
\end{tabular*}
\caption{\textbf{Tokens with more target frames per session.} Tokens per answer fall as more target frames share a session. An answer is one point at one target frame. Columns as in Tab.~\ref{tab:multi-point-tokens}. Ratio: one-frame Total over the current Total in the same subset.}
\label{tab:multi-frame-tokens}
\end{table}

\subsubsection{Multi-frame Tracking}
\label{sec:supp-multiframe}

Having examined scaling the number of query points in the reference frame, we now turn to scaling the number of target frames for a single query point. We evaluate whether accuracy on a fixed target frame is preserved as more targets are requested. We also compare predicting target frames together or separately on the same point--frame pairs to assess the benefit of processing target frames together. A session with one target frame shows the reference and target images, and a session with several target frames shows the reference with all of its target frames.

\paragraph{First-frame stability under target scaling.}
We examine whether requesting additional targets affects accuracy on the first target frame. We evaluate GPT-6.1 Sol with a single query point. Sessions request the target at gap 18, add the target at gap 36 in the two-frame setting, and also include gap 54 in the three-frame setting. Accuracy on the same first target stays at or above the single-frame baseline across all three subsets (Tab.~\ref{tab:multi-frame-accuracy}, left). Points with displacement benefit most from adding the second target. Adding the third changes APD by less than one point in every subset.

\paragraph{Prediction accuracy with multiple target frames.}
We compare prediction accuracy on the same point--frame pairs when target frames are predicted together or separately. Asking the targets together improves APD for both the two- and three-frame sets (Tab.~\ref{tab:multi-frame-accuracy}, right). The improvements are largest for points with displacement and smaller for the mixed subset. Points with negligible displacement stay close to their already high baseline.

\paragraph{Token efficiency under target scaling.}
Using the same token accounting as in the multi-point study, we measure usage per answer, where each answer corresponds to one point at one target frame (Tab.~\ref{tab:multi-frame-tokens}). From one to three target frames per session, mean total tokens per answer fall by 44--52\% across the three subsets, and input, uncached input, and output tokens all decrease. Tokens per answer still fall from two to three frames in every subset.

\par\endgroup

\subsection{Qualitative Examples}
\label{sec:supp-qualitative}

{\makeatletter\let\@xfloat\supp@xfloat\makeatother
\begin{figure}[!t]
    \centering
    \includegraphics[width=\columnwidth]{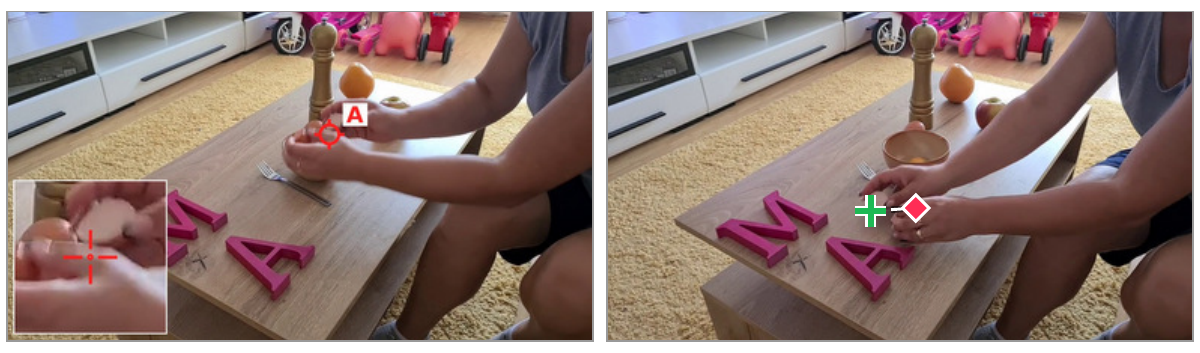}\\[2pt]
    \includegraphics[width=\columnwidth]{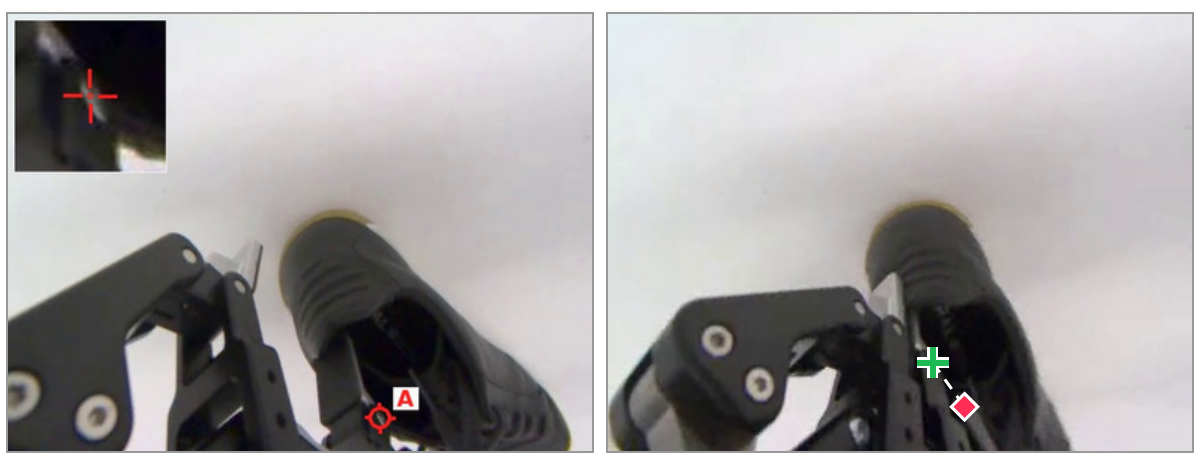}
    \caption{\textbf{Confusion between look-alike parts.} Final predictions of \themodel with GPT-6 Astra. Top (Perception Test): the queried point lies on an egg that is then cracked open, and it follows the half of the shell in the left hand. The prediction lands on the right hand. Bottom (RoboTAP): the queried point lies on one finger of a robot gripper, and the prediction lands on the other finger. Left: reference frame with query marker A. Right: target frame with the ground-truth position (green cross) and final prediction (red diamond).}
    \label{fig:pair-cases}
\end{figure}
}

\paragraph{Look-alike parts.}
\phantomsection\label{sec:supp-failure-cases}
Fig.~\ref{fig:pair-cases} shows two failures of \themodel with GPT-6 Astra. In the first, a point on an egg follows the half of the shell in the left hand after the egg is cracked open, but the agent places it on the right hand. In the second, a point on one finger of a robot gripper ends on the other finger, which looks the same. Both points stay visible in every frame between the reference and the target. The agent inspects several views in each session and still keeps the wrong part, ending 19.8 and 29.0 pixels from the target.

\renewcommand{\floatpagefraction}{0.65}
\renewcommand{\dblfloatpagefraction}{0.65}

\section{\thedataset}
\label{sec:supp-egotap4d}

Fig.~\ref{fig:annotation-pipeline} of the main paper summarizes how geometry, tracking, and VLM reconnection produce \thedataset annotations. This section describes the data sources (Sec.~\ref{sec:supp-recordings}), the annotation pipeline (Sec.~\ref{sec:supp-trajectory-annotation}), the annotation quality (Sec.~\ref{sec:supp-annotation-quality}), and the D4RT experiments (Sec.~\ref{sec:supp-d4rt}).

\subsection{Dataset Sources}
\label{sec:supp-recordings}
 
We combine Xperience-10M~\citep{xperience10m} and Ego-Exo4D~\citep{egoexo4d}, processing both at 10\,fps. From Xperience-10M, we use 3,740 episodes totaling 253.1\,h, with the left camera view, calibrated stereo depth, SLAM camera poses, and hand motion capture where available. From Ego-Exo4D, we use 1,659 takes totaling 44.1\,h, which provide no hand motion capture. In total, \thedataset contains 5,399 videos totaling 297.2\,h, with an average duration of 198\,s.

\subsection{Annotation Pipeline}
\label{sec:supp-trajectory-annotation} 

We annotate each recording in five steps (Fig.~\ref{fig:annotation-pipeline}). We first recover metric depth and camera poses for every frame. We then label each object as static or moving in every frame and use these labels to build trajectories separately for static surfaces, moving objects, and hands. Next, we reconnect the tracks of objects that leave the field of view and return. Finally, we lift the trajectories to 3D, remove unreliable 3D positions, and store the annotations.

\subsubsection{Metric Depth and Camera Poses}
\label{sec:supp-metric-geometry}

For every frame, we need a metric depth map $d_t$ and a camera-to-world pose $T_t^w$, with all poses of a recording in a common world coordinate system. A pixel $p_t$ of track $a$ is then back-projected to the 3D point
\begin{equation}
    X_a(t) = T_t^w\,\Pi^{-1}\!\left(p_t, d_t(p_t)\right),
    \label{eq:lift}
\end{equation}
where $\Pi^{-1}$ is the inverse projection with the camera intrinsics. Xperience-10M provides stereo depth and SLAM camera poses, whereas Ego-Exo4D provides camera poses from Aria Machine Perception Services (MPS)~\citep{engel2023project} but no dense depth. We therefore process the two sources separately.

\paragraph{Xperience-10M.}
We downsample the original 20\,fps rectified left video to 10\,fps. For camera poses, we use the SLAM trajectory provided with the dataset. We obtain the camera-to-world transform $T_t^w$ of each frame by composing the headset pose with the camera-to-headset extrinsics. To detect frames with inaccurate poses, we back-project background pixels outside the GMOS masks to 3D with the stereo depth and project them into frames at least one second away. A projection is inconsistent if it lies in front of the observed surface or falls inside an object mask. Frames with an unusually high fraction of inconsistent projections are marked unreliable and excluded from annotation.

\begin{table}[t]
    \centering
    \fontsize{7.0}{8.4}\selectfont
    \begin{tabular*}{\linewidth}{@{\extracolsep{\fill}}lcc@{}}
        \toprule
        \textbf{Landmark depth} & Stereo depth & Calibrated \\
        \midrule
        0--0.5\,m & $+1.3$ & $+0.7$ \\
        0.5--1\,m & $+1.7$ & $+0.2$ \\
        1--1.5\,m & $+4.0$ & $+0.1$ \\
        1.5--3\,m & $+6.9$ & $-0.2$ \\
        \bottomrule
    \end{tabular*}
    \caption{\textbf{Depth calibration accuracy on Xperience-10M.} Median relative depth error (\%) over all recordings, measured against held-out SLAM landmarks. Positive values indicate overestimated depth.}
    \label{tab:depth-calibration}
\end{table}

We find that the stereo depth provided with Xperience-10M systematically overestimates distance, and the error grows with distance. The bias differs across recordings and is consistent with imperfect rectification of the stereo pair. Such errors perturb the disparity, which is inversely proportional to depth. We therefore correct the stereo depth $d$ of each recording with an affine model in inverse depth,
\begin{equation}
    \frac{1}{z} = \frac{a}{d} + b,
    \label{eq:depth-calibration}
\end{equation}
where $z$ is the corrected depth. We fit $a$ and $b$ to sparse 3D points that do not rely on the stereo depth: points triangulated from Kanade--Lucas--Tomasi (KLT) tracks~\citep{lucas1981iterative} with the SLAM poses, and the SLAM landmarks provided with the dataset. We use half of these points for fitting and hold out the other half for evaluation. On the held-out landmarks, the stereo depth overestimates distance by 1--2\% within 1\,m and by 7\% at 1.5--3\,m, with a quarter of the recordings exceeding 16\% (Tab.~\ref{tab:depth-calibration}). After calibration, the median error stays within $\pm$1\% in every depth range up to 3\,m, and the held-out KLT points give consistent results.

\paragraph{Ego-Exo4D.}
We take every third frame of the Aria RGB video to obtain 10\,fps and undistort each fisheye frame to a $512\times512$ pinhole image. Camera poses come from the MPS closed-loop trajectory~\citep{engel2023project}. Since Ego-Exo4D provides no dense depth, we combine monocular metric depth from MoGe-3~\citep{kong2026fine}, stereo depth from FoundationStereo~\citep{wen2025stereo} on two stereo pairs of Aria cameras, and the semi-dense MPS points. For static regions, we correct the monocular depth in inverse depth as in the Xperience-10M calibration (Eq.~\ref{eq:depth-calibration}), but fit the correction separately for each frame to the MPS points visible in it. For moving objects, which the MPS points do not cover, we rescale the monocular depth to match nearby stereo depth and refine the depth along each point track by optimizing stereo, multi-view photometric, motion smoothness, and monocular prior terms.

We discard low-confidence stereo depth, depth near object boundaries, and depth outside 0.15--3.5\,m. For both sources, we fuse the static scene of each recording with TSDF fusion~\citep{tsdf}. We treat pixels outside objects that move during the recording as static and integrate their depth from all frames with reliable poses into a single volume. We then render the fused surface into each frame and use the rendered depth for static pixels, leaving static pixels without a valid fused depth empty. Moving objects, people, and hands keep their per-frame depth. The fused depth reduces per-frame noise, fills small holes, and gives each static point a consistent 3D position across frames.

\subsubsection{Motion Segmentation}
\label{sec:supp-motion-segmentation}

We obtain object instance masks from GMOS~\citep{gmos} and label each instance as static or moving in every frame, based on multi-view depth consistency, hand contact, and whether the instance is a person. We detect hand contact from SAM~3.1 hand masks~\citep{sam3} and, where available, hand motion capture. We label people, and objects whose motion is uncertain, as moving. While an object is out of view, we assume that it remains in place if it was static when it left, unless later frames show that it has moved.

\subsubsection{Surface Trajectories}
\label{sec:supp-surface-trajectories}

We obtain trajectories for static surfaces from scene geometry, for moving objects from point tracking, and for hands from motion capture.

\paragraph{Static surfaces.}
For static surfaces, we project each 3D point of the fused scene into every frame with the camera poses. We mark a static point visible in a frame when its projected depth matches the depth map.

\paragraph{Moving objects.}
We track points on moving instances with AllTracker~\citep{alltracker} and sample new query points when the existing tracks no longer cover the instance. We sample no query points on hands. Each AllTracker run covers at most 5\,s before and 10\,s after its query frame. AllTracker also predicts a visibility and a confidence score for each point, and a point is visible when its visibility is at least 0.5. We keep a visible point only if its confidence is at least 0.5 and it lies on its own object mask, has valid depth, and is not on a hand, so that a track does not jump onto an occluding hand. Otherwise, we keep its 2D position but mark it invalid. Depth errors, for example on transparent surfaces, can make a static object appear to move. We therefore also triangulate the tracks of moving objects from the camera poses alone, without depth. If an object's tracks over frames labeled as moving triangulate to fixed 3D points with a reprojection error of at most 3 pixels, the object is in fact static in these frames, and we relabel them as static. If the depth in these frames differs from the triangulated depth by more than 10\%, we use the triangulated points as the 3D positions, and frames without a triangulated point have no 3D position.

\paragraph{Hands.}
We fit MANO~\citep{mano} meshes to the hand motion capture, whose 778 vertices per hand give exact correspondences over time. We project the vertices into each frame and compare them with the depth map to label each vertex as visible, occluded, self-occluded, or inconsistent with the depth. Frames without motion capture have no hand annotations.

\subsubsection{Reconnecting Reappearing Points}
\label{sec:supp-reconnection}

{\makeatletter\let\@xfloat\supp@xfloat\makeatother
\begin{figure*}[t]
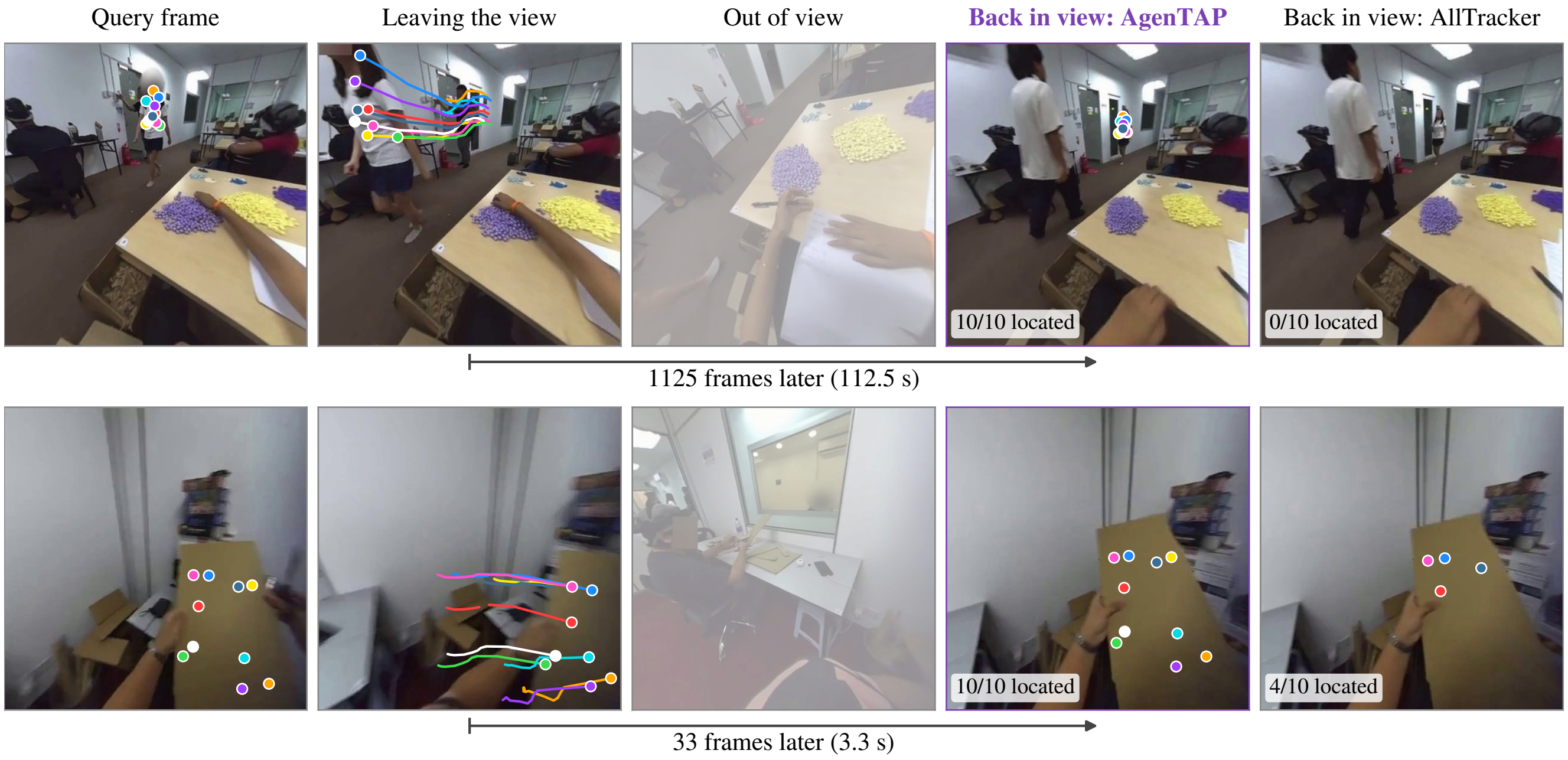

    \centering
    \includegraphics[width=\textwidth,trim=0 463.2bp 0 0,clip]{figures/graphics/annotation_reappearance_blurred.pdf}\par\nointerlineskip
    \includegraphics[width=\textwidth,trim=0 0 0 241.3bp,clip]{figures/graphics/annotation_reappearance_blurred.pdf}
    \caption{\textbf{More point reconnections after reappearance.} Two further Xperience-10M sequences in the layout of Fig.~\ref{fig:egotap4d-reconnection}. From left to right, the columns show the query points, their tracks shortly before the exit, a frame in which they are out of view (dimmed), and the predictions of \themodel and AllTracker after the return. Colors denote point identity. Points predicted invisible are omitted, and each prediction panel reports how many of the ten query points are predicted visible. The arrow gives the time between exit and re-entry.}
    \label{fig:annotation-reappearance}
\end{figure*}
}

After an object leaves the field of view, GMOS sometimes assigns its instance identity to a different object that appears later. Before reconnecting tracks, we show GPT-6.1 Sol the object before and after the gap and ask whether it is the same object. If it is not, we give the later mask a new instance identity.

When an object or a person leaves the field of view and returns, we try to continue its existing tracks after the gap. We bridge gaps of up to one second with AllTracker and keep a bridge when its forward--backward error is at most 3 pixels. By design, we leave gaps between one and three seconds unconnected and use the VLM for longer ones.

For gaps of at least three seconds, we ask \themodel with GPT-6 Astra to locate 10 query points from before the gap in several frames after the object returns (Fig.~\ref{fig:egotap4d-reconnection} and Fig.~\ref{fig:annotation-reappearance}). The query points are the object's existing tracks or textured points that we sample in the last frame before the gap in which the object is large enough. Each image pair shows the frame of the query points and one frame after the return. Sharing one session among several points keeps accuracy and lowers token usage (Sec.~\ref{sec:supp-scaling}). We do not query the VLM when the object is static on both sides of the gap, stays in view for less than one second after returning, or never appears large enough to sample query points. Of the 14,091 image pairs with 10 points each, 7,057 pass filtering, with 70,570 query points. Of these points, 23,072 pass the checks below and enter the dataset.

We accept a VLM prediction only if the VLM finds the queried point and the prediction lies on the same object, has valid depth, and is not on a hand. We then track each accepted prediction with AllTracker to a second frame after the return and compare it with the VLM prediction in that frame. Matching predictions verify the reconnection, and conflicting ones are rejected. Predictions that cannot be cross-checked are kept but marked unverified. We do not fill in positions during the gap.

\subsubsection{3D Lifting, Filtering, and Annotation Format}
\label{sec:supp-lifting-records}

We back-project every valid track point to 3D with Eq.~\ref{eq:lift}. Stereo noise can make a back-projected point jump along the viewing ray. We invalidate the 3D position of a track in frames where it spikes or deviates from its neighboring frames along the ray by more than 2\,cm or 4\% of its distance, whichever is larger, while keeping its 2D position. When at least half of an object's tracks jump in the same frame, we discard the depth of the entire object in that frame.

Each track stores its 2D and 3D positions, object instance, temporal segments, per-frame visibility state, and annotation source. The source indicates whether a position comes from AllTracker, a short-gap bridge, a verified or unverified VLM reconnection, or triangulation. The visibility state is one of visible, masked, occluded, out of frame, object absent, or position missing. A point may have a valid 2D position but no valid 3D position.

\subsection{Annotation Quality}
\label{sec:supp-annotation-quality}

Fig.~\ref{fig:egotap4d-reconnection} of the main paper and Fig.~\ref{fig:annotation-reappearance} compare how \themodel and AllTracker locate ten query points after an object or a person returns to the field of view. Fig.~\ref{fig:annotation-examples} of the main paper shows 3D trajectories from six recordings in world coordinates, together with the corresponding 2D trajectories in the video frames.

\paragraph{Human audit.}
For the audit in Tab.~\ref{tab:reappearance-human-agreement}, we select reappearance events from Xperience-10M in which a rigid object or a person is out of view for at least 3\,s and visible for at least 1\,s before and after. We leave out deformable and transparent objects. Two annotators label the query points in the first second after the object returns. They work independently and see neither the model predictions nor each other's labels. Points on people and on moved objects count as dynamic, and the rest as static. The audit has 582 point--frame pairs from 49 events, 428 dynamic from 41 events and 154 static from 8 events. We score each method against both annotators with the benchmark metrics (Sec.~\ref{sec:supp-evaluation-protocol}) and report the average. For the human row, we score one annotator against the other.

\subsection{D4RT Training and Evaluation}
\label{sec:supp-d4rt}

\paragraph{Training comparison.}
\label{sec:supp-d4rt-training}

Tab.~\ref{tab:d4rt-data} compares our reproduction of D4RT~\citep{d4rt} with and without \thedataset. Both models use the D4RT architecture and train for 200k steps on 8 nodes, whereas the published D4RT trains for 500k steps on a mixture of public and internal datasets. The baseline uses eight public datasets: Dynamic Replica~\citep{dynamicstereo}, PointOdyssey~\citep{pointodyssey}, FallingThings~\citep{fallingthings}, Kubric~\citep{kubric}, CO3Dv2~\citep{co3d}, TartanAir~\citep{tartanair}, TartanGround~\citep{tartanground}, and Waymo Open~\citep{waymo}. The second model also uses \thedataset, and all datasets are sampled with equal weights. The added-data run uses all annotation sources, including VLM reconnections, with supervision restricted to valid image, visibility, and 3D targets.

\paragraph{Evaluation protocol.}
\label{sec:supp-d4rt-evaluation}

Evaluation covers Aria Digital Twin (ADT) and PStudio from TAPVid-3D~\citep{tapvid3d}, DexYCB~\citep{dexycb}, and DROID~\citep{droid}. For DexYCB, we sample points on the annotated object meshes. For DROID, we sample points on the robot, whose 3D pose follows from its joint angles. We report AJ, $\mathrm{APD}_{\mathrm{3D}}$, and OA as defined in TAPVid-3D, separately for each dataset. Overall is the unweighted arithmetic mean of the four dataset scores, so larger evaluation sets do not dominate the aggregate. Adding \thedataset improves all three aggregate metrics. Localization improves on ADT, PStudio, and DROID, while DexYCB has lower AJ and $\mathrm{APD}_{\mathrm{3D}}$ despite improved visibility.

\end{document}